\documentclass[11pt,fullpage,english]{article}

\usepackage{lgrind}
\usepackage{amsmath}
\usepackage{amsthm}
\usepackage[frak=mma]{mathalfa}
\usepackage[pdftex]{graphicx}
\usepackage{amsfonts}
\usepackage{amssymb}
\usepackage[usenames]{color}
\usepackage{mathrsfs}
\usepackage{natbib}
\usepackage[shortlabels]{enumitem}
\usepackage{setspace}
\usepackage[normalem]{ulem}

\usepackage{hhline}
\usepackage{latexsym}
\usepackage{ifthen}
\usepackage{lscape}
\usepackage{chngpage}
\usepackage{bibunits}
\usepackage{enumitem}
\usepackage{relsize}

 \usepackage{chngcntr}
\counterwithout{table}{section}

\DeclareMathOperator*{\argmax}{argmax}

\usepackage{amsthm}
\newtheorem{prop}{Proposition}

\newtheorem{Cor}{Corollary}
\newtheorem{definition}{Definition}
\newtheorem{lem}{Lemma}
\newtheorem{theorem}{Theorem}
\newtheorem{rem}{Remark}
\newtheorem{Assum}{Assumption}
\newtheorem{exm}{Example}

\def\N{\mathbb N}

\usepackage[english,activeacute]{babel}
\usepackage{graphicx}
\usepackage[usenames]{color}
\usepackage{amssymb}
\usepackage[pagebackref=true, colorlinks=true, citecolor=blue, anchorcolor=blue, urlcolor=blue]{hyperref}

\usepackage{hyperref}

\newcommand{\Indi}{1\!\!1}
\newcommand{\e}{\mathbb{E}}
\newcommand{\ed}{\mathbb{E}_{\delta}}
\newcommand{\eL}{\mathbb{E}_{\rm \tiny 0}}

\newcommand{\D}{\mbox{\rm d}}
\newcommand{\var}{\mathbb{V}\mbox{\rm ar}}
\newcommand{\Pro}{\mathbb{P}}
\newcommand{\Fbv}{\bar{F}_{\mbox{\tiny V}}}
\newcommand{\pv}{p_{\mbox{\tiny V}}}

\newcommand{\bFv}{\bar{F}_{\mbox{\tiny V}}}

\newcommand{\thetaH}{\theta_{\mbox{\tiny \rm 1}}}
\newcommand{\thetaL}{\theta_{\mbox{\tiny \rm 0}}}

\newcommand{\lamH}{\lambda_{\mbox{\tiny \rm H}}}
\newcommand{\lamL}{\lambda_{\mbox{\tiny\rm L}}}

\newcommand{\QLi}{Q_{\mbox{\tiny\rm 0}i}}

\newcommand{\QLZ}{Q_{\mbox{\tiny\rm 0}0}}

\newcommand{\RoLi}{\rho_{\mbox{\tiny\rm L},i}}
\newcommand{\RoHi}{\rho_{\mbox{\tiny\rm H},i}}

\newcommand{\deltau}{\underline{\delta}}
\newcommand{\deltab}{\bar{\delta}}

\newcommand{\lam}{\lambda}

\newcommand{\LL}{\mbox{\tiny \rm 0}}
\newcommand{\HH}{\mbox{\tiny \rm 1}}

\newcommand{\Fixed}{{\mbox{\tiny \rm F}}}
\newcommand{\Asym}{{\mbox{\tiny \rm A}}}
\newcommand{\Heu}{{\mbox{\tiny \rm H}}}

\newcommand{\F}{{\mbox{\tiny \rm F}}}

\newcommand{\DD}{{\mbox{\tiny \rm D}}}
\newcommand{\TS}{{\mbox{\tiny \rm TS}}}

\newcommand{\cD}{{\mathcal{D}}}
\newcommand{\cG}{{\mathcal{G}}}
\newcommand{\cS}{{\mathcal{S}}}
\newcommand{\cR}{{\mathcal{R}}}

\newcommand{\cJ}{{\mathcal{J}}}
\newcommand{\cK}{{\mathcal{K}}}

\newcommand{\MV}{{\mbox{\tiny \rm MV}}}
\newcommand{\MR}{{\mbox{\tiny \rm MR}}}

\newcommand{\NP}{{\mbox{\tiny \rm NP}}}
\newcommand{\IH}{{\mbox{\tiny \rm IH}}}

\DefineNamedColor{named}{OliveGreen}    {cmyk}{0.64,0,0.95,0.40}
\DefineNamedColor{named}{BrickRed}      {cmyk}{0,0.89,0.94,0.28}
\DefineNamedColor{named}{SeaGreen}      {cmyk}{0.69,0,0.50,0}
\DefineNamedColor{named}{BlueGreen}     {cmyk}{0.85,0,0.33,0}
\DefineNamedColor{named}{BlueViolet}    {cmyk}{0.86,0.91,0,0.04}
\DefineNamedColor{named}{NavyBlue}      {cmyk}{0.94,0.54,0,0}

\newcommand{\Victor}[1]{\textsf{\textcolor{OliveGreen}{[V: #1]}}}

\title{Diffusion Approximations for a Class of Sequential Testing  Problems}

\author{{\large Victor Araman\thanks{Olayan School of Business, American University of Beirut, Beirut, Lebanon.} \hspace{3cm} Ren\'e Caldentey}\thanks{Booth School of Business, The University of Chicago.} }
\date{}

\begin{document}
\parindent 0em
\maketitle \thispagestyle{empty}
\date{}
\begin{abstract}
{\setstretch{1}
We consider a decision maker who must choose an action in order to maximize a reward function that depends on the action that she selects as well as on an unknown parameter $\Theta$. The decision maker can delay taking the action in order to experiment and gather additional information on $\Theta$. We model the decision maker's problem using a Bayesian sequential experimentation framework and use dynamic programming and diffusion-asymptotic analysis to solve it. For that, we scale our problem in a way that both the average number of experiments that is conducted per unit of time is large and the informativeness of each individual experiment is low. 
Under such regime, we  derive a diffusion approximation for the sequential experimentation problem, which provides a number of important insights about the nature of the problem and its solution. First, it reveals that the problems of (i) selecting the optimal sequence of experiments to use and (ii) deciding the optimal time when to stop experimenting decouple and can be solved independently. Second, it shows that an optimal experimentation policy is one that chooses the experiment that maximizes the instantaneous volatility of the belief process. Third, the diffusion approximation provides a more mathematically malleable formulation that we can solve in closed form and suggests efficient heuristics for the non-asympototic regime. Our solution method also shows that the complexity of the problem grows only quadratically with the cardinality of the set of actions from which the decision maker can choose.

We illustrate our methodology and results using a concrete application in the context of assortment selection and new product introduction.  Specifically, we study the problem of a seller who wants to select an optimal assortment of products to launch into the marketplace and is uncertain about consumers' preferences.  Motivated by emerging practices in e-commerce, we assume that the seller is able to use a {\em crowdvoting} system to learn these preferences before a final assortment decision is made. In this context, we undertake an extensive numerical analysis to assess the value of learning and demonstrate the effectiveness and robustness of the heuristics derived from the diffusion approximation. 

}

\vspace{0.1cm}
\noindent{\em Keywords:} Sequential experimentation, sequential testing, Bayesian demand learning, experiment design, optimal stopping, dynamic programming, crowdvoting

\end{abstract}


\setcounter{footnote}{1}

\section{Introduction}
This paper is concerned with the problem faced by a decision maker (or DM for short) who must choose an action $a$ from a finite set of available actions $\mathscr{A}$ in order to maximize a reward function ${\cal R}(a,\Theta)$ that depends on the action $a$ taken, as well as on a parameter $\Theta$. The DM does not know the true value of $\Theta$ but has only incomplete information about it and hence about the reward function. Instead of selecting  an action immediately, the DM has the option of postponing this decision in order to {\em experiment} and gather additional information about the true value of $\Theta$. In this context, the decision maker needs to select the most effective sequence of experiments to implement through time as well as the time when to stop these experiments and select a final action $a \in \mathscr{A}$.\vspace{0.2cm}

A wide range of applications can be modeled using the above general framework. For example, the DM can be a factory manager who needs to decide if a batch of production meets specific quality standards. For that she can sample items sequentially to measure their individual condition and accordingly, extrapolate the quality assessment on the entire batch (\citealp{Qiu2014}). Alternatively, the DM can be a pharmaceutical company conducting a sequence of clinical trials to evaluate the efficacy of some new drug or vaccine (\citealp{Armitage02}). In yet  another example, the decision maker can be an educational institution designing computerized adaptive testing systems to assess the level of proficiency of a cohort of examinees in a particular subject area (\citealp{Bartroff2008}, \citealp{Finkelman08}). \vspace{0.2cm}


One particular application, which has served as our initial motivation for this paper, relates to the problem of assortment selection in the context of new product introduction. Launching new products into the marketplace offers great opportunities for companies to generate new revenue streams and increase sales. However, such endeavors represent risky bets as consumers' preferences are typically unknown and unsuccessful products are a major liability generating possibly great capital expenditure, early markdowns, serious goodwill cost, and loss of market share. It is not infrequent to witness major brands preferring to discontinue a product, shortly after its introduction, rather than taking more risks and incurring higher draining costs\footnote{Making the wrong selection has even driven many major brands to discontinue some of their products, shortly after introduction (see,  Sell Big or Die Fast, \textit{New York Times}, J. Wortham and V.G. Kopytoff, August 23, 2011).}.  To mitigate these risks, companies seek to test the market's reaction (e.g., value for the price) to new products before launching decisions are made. \vspace{0.2cm}

In general experimentation can be expensive and difficult to conduct effectively and probably worth doing only seldomly. However, in many situations, this reality is now changing as companies are beginning to recognize the potential to {\em crowdsource} such market testing activities. Online experimentation has been indeed growing exponentially in the last decade or so. Companies such as Uber, Netflix, Amazon, Microsoft and many more\footnote{We refer the reader to the \textit{spot light} articles of the March-April 2020 issue of the Harvard Business Review.} have been aggressively implementing market experimentation, through dedicated platforms, with the objective of continuously improving the online experiences of their customers and infer customers preferences. In the context of new product introduction, some companies have created {\em crowdvoting} platforms (e.g., Threadless.com\footnote{A site where anyone can design a T-shirt and submit it to a weekly contest. Viewers vote for their favorite T-shirts and the winning designs are selected for production and their designers get rewarded.}) where customers can vote for their favorite products among a menu of available options. By doing so, companies generate continuous feedbacks from the ``crowd" at almost no cost, except often for the lack of accuracy and veracity of the data gathered.  In view of these challenges, an effective execution of a crowdvoting system is required which involves deciding what is the best assortment of products to display to each individual voter in order to maximize the speed of learning as well as when to stop the experimentation process and decide which new products should be commercialized (\citet{Kohavi17}). Section~\ref{sec:Crowdvoting} is devoted to this particular crowdvoting example, which we use to illustrate the methodology and results that we develop first in Section \ref{sec:AsymptoticForm} for the general case. \vspace{0.2cm}

Motivated by the operating conditions of many online experimentation platforms, our general formulation of the decision maker's problem and its analysis are based on two important and distinctive features:\linebreak (i) We assume that the time epochs at which experimentation is possible are driven by an exogenous point process that the DM does not control.  (ii) We consider environments in which the average number of experiments that can be conducted per unit of time is large but the amount of information generated by each individual experiment is low. In the context of the crowdvoting example, the first assumption accounts for a stochastic arrival of viewers/voters to the platform website. As for the second feature, it depicts, as mentioned above, the high velocity at which data can be collected online but also captures the fact that such data is inherently more noisy and less reliable than when experiments are more targeted and carefully designed (e.g., focus groups or surveying experts).  Under these conditions, we are able to use asymptotic analysis to derive a diffusion approximation for both the sequential experimentation problem and the underlying optimal stopping problem that the decision maker must solve. As we will see, the diffusion model provides a number of important insights about the nature of the problem and its solution. First, it reveals that the problems of (i) selecting the optimal sequence of experiments and (ii) deciding the optimal time to stop experimenting decouple and can be solved independently. Second, it shows that an optimal experimentation policy is one that chooses the experiment that maximizes the instantaneous volatility of the belief process, a proxy of the learning process. This {\em maximum volatility principle} reduces dramatically the complexity of the dynamic experimentation selection problem and its solution.  Third, the diffusion approximation also provides a more mathematically malleable formulation of the optimal stopping problem that we can solve in closed form. Interestingly, the computational complexity of the latter grows only quadratically with the cardinality of the set of actions $\mathscr{A}$; in fact we show that solving a problem with $|\mathscr{A}|$ actions is equivalent to solving a collection of $|\mathscr{A}|\,(|\mathscr{A}|-1)$ problems each with only two actions. Fourth, by reinterpreting the maximum volatility principle, we can reformulate the problem of selecting an optimal experimentation policy  as a Tchebycheff moment problem that sheds some light on how one could tackle the problem of \textit{experiment design}, i.e., which experiments to make available in the first place to the decision maker. In addition, we obtain from our diffusion approximations, heuristics-policies for the moderate, non-asymptotic regime. These heuristics turn out to be extremely effective and robust as shown in our numerical analysis. Finally, as a by-product of our analysis of the crowdvoting example in Section~\ref{sec:Crowdvoting}, we derive diffusion approximations for a setting in which experimentation and learning are driven by the choices that voters make under a multinomial choice model (MNL).  Given the popularity of the MNL model to represent consumer preferences, we believe that our approach to obtaining  diffusion approximations can possibly be applied to a number of other applications beyond those discussed in this paper.

\vspace{0.2cm}



{}


\section{Related Literature}\label{sec:Literature}
Our paper is related  to two streams of literature.  Methodologically, we contribute to the literature on hypothesis testing and sequential design of experiments initiated by \cite{Wald47} in the early 40's. In terms of applications, we contribute to the operations literature on assortment planning and demand learning (e.g., \citealp{CaroGallien07} and \citealp{Kok2009}).
\vspace{0.2cm}


Sequential analysis is concerned with the problem of effectively detecting the validity of a hypothesis through sequential sampling or tests. After each (possibly costly) test and on the basis of the observed history of outcomes, the decision maker needs to either accept one of the hypotheses being tested or continue the experimentation.  The sequential probability  ratio test (SPRT) developed by \cite{Wald45} (see also  \citealp{WaldWolfowitz48}) establishes that under certain conditions an optimal policy  is determined by the first exit time of an appropriately defined likelihood ratio process from a bounded interval; the end points of this interval are determined by pre-specified  type I and II error targets. The initial formulation and ideas of Wald's SPRT test have been applied to a wide range of applications and extended in many different directions (e.g., \citealp{Siegmund85} and \citealp{Lai01}).
One important extension relevant to our work relates to the problem of sequential design of experiments, where the DM chooses dynamically the experiments to undertake from a set of available options (e.g., \citealp{Robbins52}, \citealp{Chernoff59,Chernoff72}), and do that until she decides to stop and selects what she believes is the true hypothesis. For brevity we denote thereafter this type of problem, sequential hypothesis testing. 

\vspace{0.2cm}

In terms of solution techniques large sample analysis has been commonly used to study sequential hypothesis testing problems and evaluate the asymptotic optimality of concrete (often simple) policies. The asymptotic regime in many of these studies is obtained by assuming that the cost of experimentation goes to zero (e.g., \citealp{Chernoff59}, and \citealp{Keener84}). 
Chernoff mentions that ``it may pay to continue sampling even though we are almost convinced about which is the true state of nature." 
The alluded ``inefficiency" in Chernoff's regime is required to guarantee a probability of error that is proportional to the cost of experimentation that is becoming increasingly small.  Our work also relies on a type of asymptotic analysis in which the number of experiments grow large, however, our approach differs significantly from  large sample methods as we not only scale the number of experiments but simultaneously decrease the informativeness of each experiment. As a result, in such asymptotic regime the `rate' of information that the DM collects remains comparable to those in small sample problems, and therefore when the DM is experimenting it does so only because she is still unsure of the true hypothesis.  This interplay between larger sample sizes and less informative experiments  was also recently explored by \cite{Naghshvar13}. They also rely on large sample analysis, but introduce a multiple hypothesis setting, and represent the limited informativeness by scaling the number of hypotheses. Their results are a generalization of \cite{Chernoff59} where they suggest adjusted policies and find tight bounds to prove their asymptotic optimality.  
In the context of multi-armed bandit problems, \cite{WagerXu21} and \cite{FanGlynn21} are two  recent arXiv preprints that study a similar type of asymptotic regime and diffusion limits as the ones considered in this paper. In particular, they consider a regime in which the mean rewards of the arms scale as $1/\sqrt{n}$, where $n$ is the number of arm pulls. \cite{WagerXu21} suggest a framework governed by a well behaved sampling function to implement such approach in the context of sequential experimentation.  \cite{FanGlynn21} develop the theory from first principles in the specific context of Thompson sampling.
In our two hypothesis setting, we  introduce a general framework  to  model  lack of informativeness. This framework includes for instance the case of asymptotically indistinguishable hypothesis as well as settings where the experiments generate increasingly noisy outcomes. We show that under our asymptotic regime, the sequential experimentation problem reduces to a diffusion free boundary problem which we are able to solve and develop approximations for the non-asymptotic regime. Other papers have studied diffusion models in the context of sequential testing (e.g.,  \cite{Chernoff61}, \cite{Breakwell64},  \cite{Peskirbook} or \cite{HarrisonSunar15}), although in our case we make no \textit{Gaussian} assumption regarding the initial process that is being observed. Other examples of sequential analysis papers that have relied on diffusion approximations include the work on Bayesian multi-armed bandits by \cite{ChangLai87} and \cite{BrezziLai02}, on ranking and selection problems by \cite{ChickGans09} and \cite{ChickFrazier12}, and also in the context of strategic experimentation, with \cite{Bolton} who consider a many-agent  two-armed Bernoulli bandit problem in which agents can learn from the experimentation of other agents (i.e., information as a public good).


\vspace{0.2cm}

Our work also contributes to a growing stream of  sequential  hypothesis  testing problems in the context of best  arm  identification (BAI)   (see, \citealp{russo2016simple}, \citealp{garivier2016optimal}, and \citealp{kaufmann2016complexity}). 
In our sequential hypothesis testing setup, we interpret each available experiment as an `arm' that when pulled generates information on the true hypothesis.  A key difference between our model and this literature is that we allow for the possibility that the set of arms available for learning to be different from the sets of arms from which the DM chooses a final action.
One feature of our model is that the DM learns about the true hypothesis from any pulled arm. This behavior is similar to some  BAI settings where the unknown parameters can affect the reward of multiple correlated arms, (see, \citealp{Soare14}). Moreover, in the illustrative example of Section~\ref{sec:Crowdvoting}, we assume that an experiment is an assortment of products offered to a customer and the outcome is the product selected by that customer. We assume in this example that this selection happens following an MNL model making this setup similar to an  MNL-bandit like exploration (see the recent work of, \citealp{Agrawal2017} and \citealp{OhIyengar19}). Despite some structural difference with BAI and more generally, MAB literature, we compare in the numerical section the performance of some MNL-bandit algorithms - introduced in the literature - with the ones we suggest here. 
\vspace{0.2cm}

Finally, we recall that this work naturally belongs to the broad area of reinforcement learning. Our suggested heuristics can be viewed as approximate DP techniques for solving a dynamic learning problem. Such techniques have been shown to be effective in managing the curse of dimensionality (see, \citealp{Powell12}). In this recent review, Powell divides ADP policies in four categories: myopic cost function approximations, lookahead policies, policy function approximations and  policies based on value function approximations. The latter two are often based on the specific structure of the problem. Indeed, most of the heuristics we suggest (see, Section~\ref{sec:heuristic}) belong to these two categories and are obtained either by reducing carefully the set of policies we are optimizing on, or by approximating the value function itself. These approximations are primarily inspired and obtained based on our asymptotic analysis. In our numerical analysis (see Section~\ref{sec:Numerics}) we also include a lookahead type policy. Some recent works have highlighted the effectiveness of simple policies in the context of dynamic learning such as greedy algorithms (e.g., \citealp{Bastani20}) and Certainty-Equivalence (e.g., \citealp{Keskin18}). The greedy algorithm behaves well when exploration is expensive while in our case it is free. As for the certainty equivalence (CE), it is not appropriate in our setting. Indeed, in a Bayesian setting, CE would assume that the current belief is constant moving forward and hence would always recommend to stop and never to explore. Having said that, we do show that in our case a simple (static) experimentation policy behaves well and is asymptotically optimal.\vspace{0.2cm}

Our paper also contributes to the operations literature on sequential testing and demand learning. There is a growing stream of papers in revenue management that have focused on the problem of characterizing optimal dynamic pricing strategies when there is incomplete information about consumers' price sensitivity (see, \citealp{AramanCaldentey11} and \citealp{denBoer2015}). In this context, pricing strategies play a dual role. On one hand, they have a direct impact on sales and revenues. On the other, they act as tools for experimentation  used by sellers  to learn demand characteristics. Optimal pricing strategies are those that balance the so-called exploration-exploitation tradeoff between these two roles, e.g., \cite{AramanCaldentey09}, \cite{BesbesZeevi09}, \cite{Bora2012}, \cite{BoerZwart14}, \cite{BroderRusme12}, \cite{GallegoTaleb12} and \cite{KeskinZeevi14}.   Another stream of papers, which is closer to the crowdvoting example that we consider in Section~\ref{sec:Crowdvoting}, focuses on optimal assortment planning under unknown demand characteristics.  In this literature, the decision maker wants to identify a revenue maximizing assortment of products from a (possibly very large) set of available options. Consumers' preferences over assortments are typically described in the form of a Luce-type choice model --with the MNL being by far the most popular choice--with unknown parameters. In this setting, the DM experiments by displaying different assortments to different consumers over time. Some representative papers in this area include \cite{CaroGallien07}, \cite{Uluetal2012}, \cite{SaureZeevi13}, \cite{Agrawal2017} and \cite{Yifan18}. A variant of this line of research is the recent paper of \cite{Keskin19}, where the seller faces unknown cost functions that increase with the quality of the products. At each period, the firm selects  vertically differentiated products and self-selection pricing mechanisms to learn and maximize its profit over a finite horizon.\vspace{0.2cm}

Finally, our research also contributes to the recent and growing literature on crowdsourcing and specifically crowdvoting. 
We mention the work of \cite{Krager14} that looks at adaptively  allocating small tasks to workers through crowdsourcing while meeting some reliability target. The  recent work of \cite{Papanastasiouetal18} tackles the provision of information dissemination in an online setting where customers' selection of products/services is affected by historical outcomes. 
On the crowdfunding end, \cite{Alaei16} suggest a dynamic model of crowdfunding and assess the probability of success of a campaign by introducing the notion of anticipating random walks. On crowdvoting, the paper by \cite{MarinesiGirotra13} focuses on measuring  the information that is acquired from a customer voting system. Using a two-period game-theoretical model, they prove among other results that by offering a sufficiently high discount during the voting phase, crowdvoting systems - used to decide whether to develop the product or not -   represent an effective way to elicit information on customers willingness-to-pay. Finally, motivated by recent applications in blockchain-based platforms,  \cite{Tsoulkalas19} consider the problem of information aggregation from a collection of  partially informed agents having private information about some unknown state of the world (e.g., the quality of product). Agents submit a vote --in the form of an estimate of the true value of state of the world-- and the platform aggregates these votes to produce a final estimate. The value of the platform and the payoffs collected by the agents depend on the accuracy of this final estimate. The paper studies the impact of using different weighting mechanisms to aggregate votes on agents voting strategies and the informativeness of the resulting equilibrium outcomes. Our focus however through our crowdvoting example, is on how to operationally manage a voting platform that faces a stream of myopic (non-strategic) consumers. In that regard our work is close to \cite{Yifan18}.

\vspace{0.2cm}



The rest of the paper is organized as follows. In the next section, we introduce the different components of the general model together with the main assumptions and formulate the problem as a two-stage dynamic programming problem where the first stage is concerned with the experiment design while the second stage tackles the duration of the experimentation. We also prove the convexity of the value function and discuss how one can leverage this property to simplify the optimization and generate a simple heuristic for the experimentation. Section~\ref{sec:AsymptoticForm} is fully devoted to the asymptotic analysis. We start by describing the scaling and the corresponding regime and obtain a diffusion formulation of the original problem. We move next to solving for the corresponding optimal experimentation policy as well as the optimal stopping of the experimentation phase. These two decisions are shown to decouple. We first show that a static experimentation policy is optimal at the limit and the optimal experiment is the one that maximizes the volatility of the belief process. As for the optimal duration of experimentation,  it is formulated as an optimal stopping problem very much in the spirit of Wald's SPRT test. The solution is fully characterized by a partition of the belief space into a collection of intervals that determine those regions where experimentation is needed or not.  Both the value function and the expected value of the stopping time are obtained in closed form. Motivated by the simple optimality principle that characterizes the diffusion problem, we suggest in Section~\ref{sec:heuristic} heuristics for the experimentation policy under a non-asymptotic regime, and discuss ways to identify the various components of the heuristics while relying only on the initial primitives of the problem.  In Section~\ref{sec:Crowdvoting}, we discuss in detail an illustrative example of the assortment selection in the context of new product introduction. The results of the previous sections are adapted to this setting followed, in Section~\ref{sec:Numerics}, by an extensive numerical analysis. For that, various heuristics of the display set policy are introduced and compared numerically to the diffusion-derived heuristics, confirming the high performance and robustness of the latter. Finally, given the connection of our setting in Section~\ref{sec:Crowdvoting} with MNL-bandit and best arm identification literature, we also show numerically that our suggested heuristics outperforms off-the shelf algorithms from this literature.   
We conclude in Section~\ref{sec:Conclusion} and offer some possible directions for future work. Most proofs have been relegated to the appendix.

\setcounter{footnote}{1}

\section{Model Description}\label{sec:model}


 We consider a decision maker (DM) who must choose an action $a$ in order to maximize a reward function ${\cal R}(a,\Theta)$ that depends on both the action $a$ as well as a parameter $\Theta$.
The DM selects the action $a$ from a finite set of available actions $\mathscr{A}$ and does not know the value of  $\Theta$ that can take one of two possible values $\{\thetaL,\thetaH\}$. Specifically, the DM has incomplete information on $\Theta$, and hence on the reward function, having a prior that $\Theta=\thetaL$ with probability $\delta \in (0,1)$.\vspace{0.2cm}

We assume that the decision maker is risk neutral. If  she were to make a decision at time $t=0$, she would then select an action that maximizes her expected reward conditional on her prior belief $\delta$. That is, she would select an action $a^*\in\mathscr{A}$ that maximizes  $\e_{\delta}[{\cal R}(a,\Theta)] $, where $\e_\delta[\cdot]$ is the expectation operator conditional on the prior belief that $\Theta=\thetaL$ with probability $\delta$\footnote{To be precise,  the probabilistic framework that we consider is defined by a probability space $(\Omega, {\cal F}; \Pro_0 ,\Pro_1)$ equipped with two probability measures $\Pro_0$ and $ \Pro_1$. For each $\delta \in [0,1]$, we associate a probability measure $\Pro_\delta= \delta\, \Pro_0+(1-\delta)\, \Pro_1$ and let $\ed[\cdot]$ denote its expectation operator. Finally, $\Theta$ is a Bernoulli random variable that satisfies $\Pro_\delta(\Theta=0)=\delta$.}. We define the optimal expected reward function 
\begin{equation}\label{eq:payoff}G(\delta):=\max_{a \in \mathscr{A}} \e_\delta[{\cal R}(a,\Theta)] = \max_{a \in \mathscr{A}}\Big\{\delta\,{\cal R}(a,\thetaL)+(1-\delta)\,{\cal R}(a,\thetaH)\Big\} \end{equation}
and let ${\cal A}^*(\delta) \subseteq \mathscr{A}$ be the set of actions at which the maximum reward is achieved. Without loss of optimality, we assume that for every $a \in \mathscr{A}$, there exists a $\delta \in (0,1)$ such that $a \in {\cal A}^*(\delta)$ (otherwise, some actions are uniformly dominated and can be removed from the set $\mathscr{A}$ of available actions). It is worth  noticing that, since $\mathscr{A}$ is a finite set, the function $G(\delta)$ is piece-wise linear in $\delta$.\vspace{0.2cm}
\subsection{The Experimentation Process}
Instead of selecting immediately an action from the set ${\cal A}^*(\delta)$, the DM has the option of postponing this decision in order to {\em experiment} and gather additional information about the true value of $\Theta$. The type of experimentation process that we consider is characterized by two key features:
\begin{enumerate}
\item The decision maker has at her disposal a finite set $\mathscr{E}$ of experiments. Each experiment ${\cal E} \in \mathscr{E}$ has associated a finite set
${\cal X}_{\cal E}$ of possible outcomes and a  likelihood function
$${\cal L}(x,{\cal E}):={Q(x,{\cal E},\thetaH) \over Q(x,{\cal E},\thetaL)}, \quad x \in {\cal X}_{\cal E},$$
where $Q(x,{\cal E},\theta):=\Pro_\theta(x|{\cal E})$ is the conditional probability of observing outcome $x \in {\cal X}_{\cal E}$ when the experiment ${\cal E}$ is used and $\Theta=\theta$. We assume that every experiment ${\cal E} \in \mathscr{E}$ is informative in the sense that there exists $x \in {\cal E}$ such that ${\cal L}(x,{\cal E}) \not=1$.


\item There exists an exogenous  Poisson  process $N_t$, with rate $\Lambda$,  that determines the time epochs $\{t_i\}_{i \geq 1}$ at which experiments are conducted, where
$t_i=\inf\{t \geq 0: N_t\geq i\}$. As a result, while the decision maker selects the experiment at each experimentation epoch, she does not have control over the exact times when these experiments are conducted\footnote{ This description assumes that the outcomes of the experiments are instantly observed.  Alternatively, we can think that each experiment takes an exponential random time (with rate $\Lambda$) to generate an outcome and only at this point in time the next experiment can be set. As a result, the outcomes of the experiments will follow again a Poisson process with rate $\Lambda$. \\ For instance, in the crowdvoting example mentioned in the introduction, experimentation occurs when a customer arrives to the online platform and votes, which we model as a Poisson process. See Section~\ref{sec:Crowdvoting} for more details.}.
\end{enumerate}

In this setting, a policy is a triplet $(\pi,\tau,a_\tau)$, where $\pi$ is an experimentation policy that adaptively determines the sequence of experiments $\{{\cal E}_{t_1},{\cal E}_{t_2}, \dots,\}$ to conduct at the experimentation epochs $\{t_i\}_{i \geq 1}$, $\tau$ is a stopping time that defines the duration of the experimentation process, and $a_\tau \in \mathscr{A}$ is the action taken at time $\tau$. Since at optimality we have $a^*_\tau \in {\cal A}^*(\delta_\tau)$, we will simply denote by $(\pi,\tau)$ a generic policy. We also denote by $\{x_{t_1},x_{t_2}, \dots, x_{t_{N_\tau}}\}$ the sequence of outcomes of the experiments and by  ${\cal F}_t$ the history (filtration) generated by the experimentation process up to time $t$. We denote by $\mathbb{T}$ the set of stopping times with respect to $\mathbb{F}=({\cal F}_t)_{t \geq 0}$. Also, and using a slight abuse of notation, we denote by ${\cal E}_t$ the experiment that is used at time $t$ and by $x_t$  the corresponding outcome. Naturally, we must have $x_t \in {\cal X}_{{\cal E}_t}$.

\vspace{0.2cm}
By judiciously selecting an experimentation policy $\pi$ and observing the outcomes of each experiment, the decision maker can gradually learn the true value of $\Theta$ over time. In particular, we define the belief process $\delta_t:=\Pro_\delta(\Theta=\thetaL|{\cal F}_t)$ whose evolution is governed by Bayes rule.

\begin{lem}[Belief Process]\label{lem:dynamic-q_t} Let $\{{\cal E}_{t_1},{\cal E}_{t_2}, \dots\}$ be a sequence of experiments and $\{x_{t_1},x_{t_2}, \dots \}$ be the corresponding sequence of observed outcomes. If the decision maker has a prior belief $\delta=\Pro_\delta(\Theta=\thetaL)$, then the belief process $\delta_t$ evolves as an ${\cal F}_t$-martingale given by:
\begin{equation}\label{eq:beliefdynamics}
\delta_t = {\delta \over \delta+(1-\delta)\, L_t},\; \mbox{where $L_t$ is the likelihood-ratio function } L_t := \prod_{i=1}^{N_t} {\cal L}(x_{t_i},{\cal E}_{t_i}).\end{equation}
\end{lem}
{\sc Proof:} This and other proofs are relegated to the Appendix. $\Box$ \vspace{0.3cm}
\subsection{The Optimization Problem}
Under some mild assumptions on the likelihood ratios ${\cal L}(x,{\cal E})$ the belief process converges to 0 or 1 depending on whether $\Theta=\thetaL$ or $\Theta=\thetaH$, respectively. Hence, an infinitely patient decision maker will eventually learn the true value of $\Theta$. However,  by running a long experimentation process the decision maker is also delaying the time when the final decision is made. If we assume  that, {\em ceteris paribus}, the decision maker prefers to collect these rewards as early as possible then she faces a trade-off between learning the true value of $\Theta$ (exploration) and collecting the reward ${\cal R}(a,\Theta)$ (exploitation). To model this trade-off we assume that the decision maker's objective is to maximize the expected discounted reward that she will collect at the time a final decision is made. That is, she is interested in solving the following optimal stopping time problem:
\begin{equation}\label{eq:probformobfnew}\Pi(\delta):=\sup_{(\pi,\tau)} \e_\delta\left[e^{-r\,\tau} G(\delta_\tau)\right],\end{equation}
where  $r$ is the decision maker's discount factor.  We will tackle the solution of \eqref{eq:probformobfnew} using dynamic programming. To this end,  we find it convenient to express the dynamic evolution of the belief process $\delta_t$ in equation \eqref{eq:beliefdynamics} using the following SDE representation.

\begin{lem}\label{lem:SDEdelta} The belief process in \eqref{eq:beliefdynamics} admits the  SDE representation:
$$\D \delta_t = \eta(\delta_{t-},x_{t},{\cal E}_{t-})\, \D N_{t}, \quad \mbox{where}\quad \eta(\delta,x,{\cal E}):=(1-\delta)\,\delta\, \left({1-{\cal L}(x,{\cal E}) \over \delta+(1-\delta)\,{\cal L}(x,{\cal E}) }\right).$$
\end{lem}
In the statement of the previous lemma, the left-limit notation  ${\cal E}_{t-}$ ($\delta_{t-}$) stands for the experiment (belief) that is chosen (observed) right before a jump of   $N_t$ at time $t$. The factor $\eta(\delta,x,{\cal E})$ is the size of the jump of the belief process ({\em i.e.}, the ``amount'' of learning) if an experiment ${\cal E}$ is chosen that produces an outcome $x$ when the belief process (just before the experiment) is equal to $\delta$.\vspace{0.2cm}

Equipped with Lemma~\ref{lem:SDEdelta}, we  formulate the decision maker's problem as a Markov Decision Problem (MDP) and without loss of optimality, restrict our attention to the class of deterministic Markovian   policies (e.g., \citealp{Blackwell65} and Section 4.4 in \citealp{Puterman05}). In particular, the experimentation policy $\pi$ maps each value of the belief $\delta$ to
an experiment $\pi(\delta) \in \mathscr{E}$ and the stopping time $\tau$ is  a hitting time of the belief process on some {\em intervention} set ${\cal I}$. We will interchangeably use $\tau$ and $\cal I$ depending on the context. In the following definition, ${\cal M}(\mathscr{E})$ is the set of measurable functions from $[0,1]$ to ${\mathscr E}$ and ${\cal B}$ is the set of Borel sets in $[0,1]$.

\begin{definition}\label{def:Markovpolicy} {\rm (Deterministic Markovian Policy)} A deterministic Markovian policy  corresponds to a pair $(\pi, {\cal I}) \in {\cal M}(\mathscr{E}) \times {\cal B}$.  For all $\delta \not\in {\cal I}$, the DM displays experiment $\pi(\delta) \in \mathscr{E}$. On the other hand, for $\delta \in {\cal I}$ the decision maker chooses to stop the experimentation process and implements an optimal action $a \in {\cal A}^*(\delta) $. 
\end{definition}

Putting all the pieces together, the decision maker's optimization in \eqref{eq:probformobfnew} can be rewritten as the following optimal control problem:
\begin{align}\label{eq:probformobfnew2}
\Pi(\delta):=&\sup_{(\pi, {\cal I})} \ed\left[e^{-r\,\tau} G(\delta_\tau)\right] \\
 \mbox{subject to:}   \quad \D \delta_t = &\,\eta(\delta_{t-},x_{t},\pi(\delta_{t-}))\, \D N_{t}, \quad \delta_0=\delta, \quad \mbox{and}\quad \tau=\inf\big\{t>0 \colon \delta_t \in {\cal I}\big\}.\nonumber
 \end{align}

Finally, we can express the optimality conditions of  the control problem in \eqref{eq:probformobfnew2} in the form of the following Hamilton Jacobi Bellman (HJB) equation:
\begin{equation}\label{eq:HJBpure}0=\max\left\{G(\delta)-\Pi(\delta) \;,\;
\Lambda\,\max_{{\cal E} \in \mathscr{E}} \Big\{\ed\Big[\Pi\big(\delta+\eta(\delta,x,{\cal E})\big)-\Pi(\delta)\Big]\Big\}-r\,\Pi(\delta) \right\},\end{equation}
with border conditions $\Pi(0)=G(0)$ and $\Pi(1)=G(1)$ since both $\delta=0$ and $\delta=1$ are absorbing belief states (see Lemma~\ref{lem:dynamic-q_t}). By solving the inner maximization, we can compute an optimal experimentation policy ${\cal E}^*(\delta)$, that is,
\begin{equation}\label{eq:Opt_Exp}
{\cal E}^*(\delta) \in \argmax_{{\cal E} \in \mathscr{E}} \Big\{\ed \Big[\Pi\big(\delta+\eta(\delta,x,{\cal E})\big)\Big]\Big\}.\end{equation}

\vspace{0.2cm}

The  HJB equation in \eqref{eq:HJBpure}  leads to a tractable computational approach to solve the decision maker's problem. For instance, we can implement the {\em value iteration} algorithm
\begin{equation}\label{eq:ValueIteration}\Pi_0(\delta)=G(\delta) \qquad \mbox{and}\qquad \Pi_{l+1}(\delta) = \max\left\{G(\delta), {\Lambda \over \Lambda+r} \,\max_{{\cal E} \in \mathscr{E}} \Big\{\ed\Big[\Pi_l\big(\delta+\eta(\delta,x,{\cal E})\big)\Big]\Big\}\right\},\end{equation}
which defines a sequence of continuous functions $\big\{\Pi_l(\delta):l\geq 0\big\}$ that are monotonically increasing in $l$ and converge uniformly to a limit $\Pi(\delta)=\lim_{l \to \infty} \Pi_l(\delta)$ that satisfies the HJB equation in \eqref{eq:HJBpure}, 
(see the proof of Proposition~\ref{prop:valuefunctionconvex} for details).\vspace{0.2cm}

Despite its computational simplicity, the HJB equation \eqref{eq:HJBpure} is not particularly malleable for the purpose of analysis and to derive structural results about an optimal solution and its properties. For this reason, in the next sections, we tackle the decision maker's optimization problem using a diffusion approximation that preserves the same trade-offs as in the original formulation but provides a more transparent representation of the problem and its optimal solution. \vspace{0.2cm}


We end this section with a numerical example that illustrates the value iteration method used above and highlights some feature of an optimal solution.

\begin{exm}\label{ex1} Suppose the decision maker has four alternative actions to choose from (i.e., $|\mathscr{A}|=4$) with corresponding payoffs ${\cal R}_1(\delta)=6-30\,\delta$, ${\cal R}_2(\delta)=4-5\,\delta$, ${\cal R}_3(\delta)=3\,\delta$ and ${\cal R}_4(\delta)=-20+25\,\delta$. There are nine possible experiments that the DM can use (i.e., $|\mathscr{E}|=9$) and each experiment produces a binary outcome, that is, ${\cal X}_{\cal E}=\{0,1\}$ for all ${\cal E} \in \mathscr{E}$. The table below specifies the probability $Q(0,{\cal E},\Theta)$ for each of the nine experiments for $\Theta=\thetaL$ and $\Theta=\thetaH$. Finally, we let $\Lambda=8$ and $r=0.5$.

\begin{table}[h]
\begin{center}
$\hspace{2.5cm}Q(0,{\cal E},\Theta)$\\
\begin{tabular}{||l||c|c|c|c|c|c|c|c|c|c|c|c|c||}  \hline
Experiment & 1 & 2 & 3 & 4 & 5 & 6 & 7 & 8 & 9  \\ \hline \hline
$\Theta=\thetaL$ & 0.1 & 0.2	& 0.3 &	0.4 &	 0.5 & 0.6 & 	0.7 & 	0.8 & 	0.9  \\  \hline
$\Theta=\thetaH$ & 0.03& 0.04 &	0.09 &	0.16 &	0.25 &	0.36 & 	0.49 &	0.68 &	0.86  \\  \hline
\end{tabular}
\caption{\footnotesize \sf  Probabilities of observing outcome `0' for each of the nine experiments as a function of the value of $\Theta$.}   \label{table:datainstance}
\end{center}\label{Table:ex1}
\end{table}

 Figure~\ref{fig:ExValueItera} depicts the numerically computed solution using the value iteration in \eqref{eq:ValueIteration} after 200 iterations. The left panel shows the value function $\Pi(\delta)$ while the right panel shows the optimal experiment ${\cal E}^*(\delta)$. We use the convention ${\cal E}^*(\delta)=0$ for those values of $\delta$ at which $\Pi(\delta)=G(\delta)$ and no experimentation is used.

\begin{figure}[htb]
    \begin{center}
    \includegraphics[width=16cm]{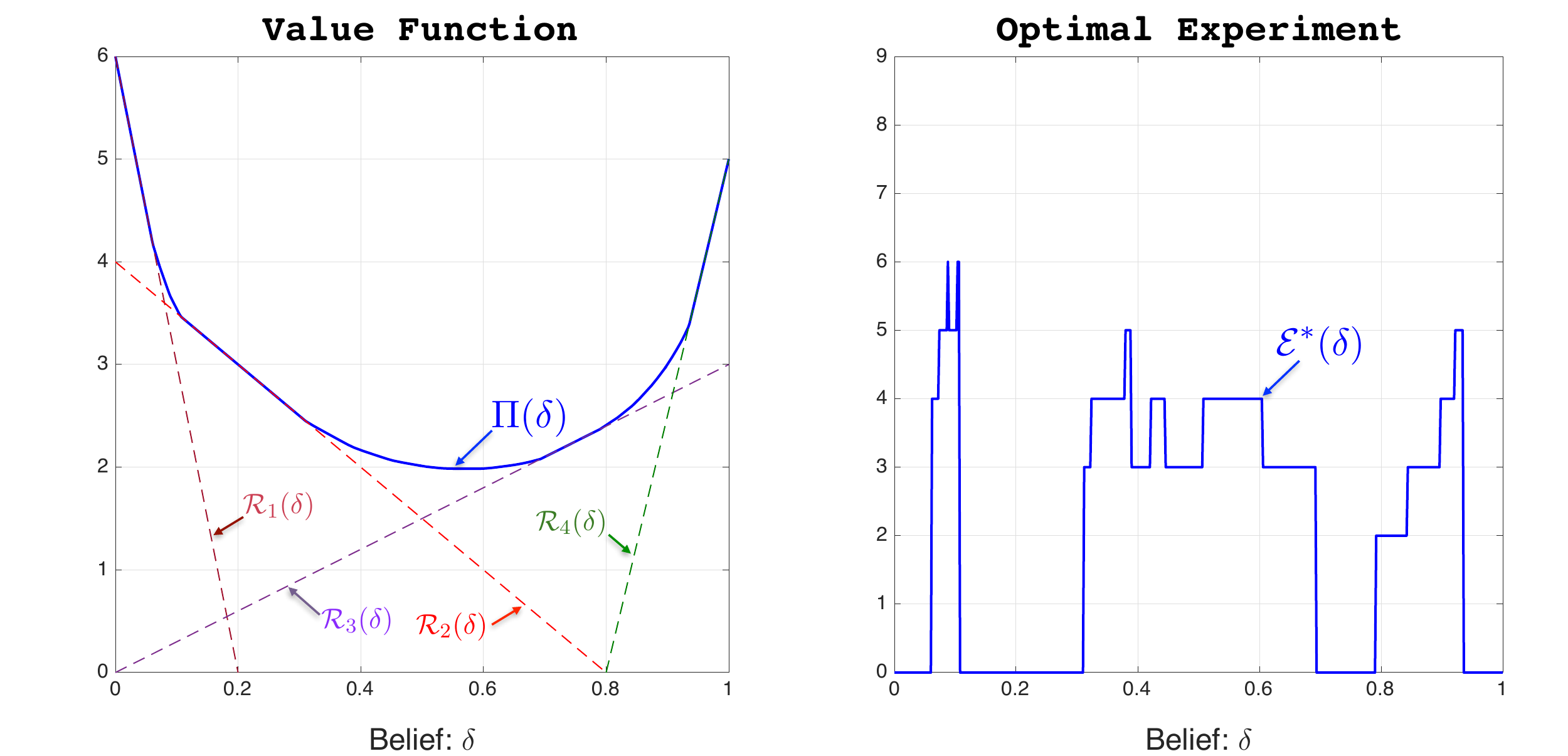}\vspace{-0.5cm}
    \end{center}\caption{\footnotesize \sf Numerically computed solution. The left panel depicts the value functions $\Pi({\delta})$. The right panel depicts the optimal experiment ${\cal E}^*(\delta)$. {\sc Data}: ${\cal R}_1(\delta)=6-30\,\delta$, ${\cal R}_2(\delta)=4-5\,\delta$, ${\cal R}_3(\delta)=3\,\delta$, ${\cal R}_4(\delta)=-20+25\,\delta$, $r=0.5$ and $\Lambda=8$. }    \label{fig:ExValueItera}
\end{figure}

As we can see, in an optimal solution, the belief space is partitioned into a collection of intervals that define the regions where experimentation is used or not used.  For example, in the interval $\delta \in [0.1,0.31]$ the decision maker does not use any experimentation and selects immediately (at time $t=0$) an action in ${\cal A}^*(\delta)$ that maximizes her expected reward (in this case ${\cal A}^*(\delta)=\{2\}$). On the other hand, in the interval $\delta \in (0.31,0.69)$ the DM wants to experiment. In this case the interval $(0.31,0.69)$ is further partitioned into a collection of subintervals in which a specific experiment is selected. For instance, for $\delta \in (0.45,0.51)$ the decision maker uses experiment 3 while for $\delta \in (0.51,0.61)$ she uses experiment 4.

We note that experimentation occurs around those values of $\delta$ where the payoff function $G(\delta)=\max_i\{{\cal R}_i(\delta)\}$ has a kink, i.e., where two payoff functions intersect.  Intuitively, in these regions a small change in the value of $\delta$ can lead to a discrete change in the optimal action to select and so the DM has locally more incentive to experiment and learn in these regions.  

\end{exm}

\subsection{On the Convexity of the Value Function}\label{rem:stochdominance}

The following proposition will prove useful in various places in the analysis that follows.

\begin{prop}\label{prop:valuefunctionconvex} The functions $G(\delta)$  and $\Pi(\delta)$ are both convex in $\delta \in[0,1]$. \end{prop}

One way in which we can take advantage of this property is to simplify the optimization problem. In some applications, the cardinality of the set of possible experiments $\mathscr{E}$ can be rather large adding an extra layer of complexity to the problem of solving the HJB equation in \eqref{eq:HJBpure}\footnote{For example, in the context of an optimal assortment selection problem with $n$ products, there are $2^n-1$ possible display sets that could be offered. We will discuss this example in detail  Section~\ref{sec:Crowdvoting}.}. 
One possible step to mitigate this issue is to reduce the number of potential experiments  to consider. Specifically, we can use the fact that the value function $\Pi$ is convex to eliminate those experiments that are dominated in a {\it convex order dominance} sense\footnote{The notion that an experiment {\it dominates} another one is similar to the notion that an experiment is {\it more informative} than another one as discussed in \cite{Blackwell51} (see also \citealp{Lindley56} and \citealp{Cam96}).}.

\vspace{0.2cm}

Indeed, for every $\delta \in (0,1)$ and ${\cal E} \in \mathscr{E}$, let $Z(\delta,{\cal E}):=\delta+\eta(\delta,x,{\cal E})$  be the random variable that defines the value of the posterior belief when the prior belief is $\delta$ and experiment ${\cal E}$ is selected. Note that $\ed[Z(\delta,{\cal E})]=\delta$ for all ${\cal E} \in \mathscr{E}$.
Suppose that for two experiments ${\cal E}_1,{\cal E}_2 \in \mathscr{E}$ we have that $Z(\delta,{\cal E}_1) \leq_{cx} Z(\delta,{\cal E}_2)$, that is,  the random variable $Z(\delta,{\cal E}_2)$ dominates $Z(\delta,{\cal E}_1)$ in the convex order sense (see  \citealp{ShakedShanthikumar94}). Then, by convexity of $\Pi$ in Proposition~\ref{prop:valuefunctionconvex}, we get that $\ed[\Pi(Z(\delta,{\cal E}_1)] \leq \ed[\Pi(Z(\delta,{\cal E}_2)]$. As a result, experiment ${\cal E}_1$  can be excluded from the set of possible experiments to be considered when the belief process equals $\delta$. It is worth highlighting that this elimination procedure does not rely on any specific knowledge of the value function beyond the fact that it is convex.\vspace{0.2cm}

One  class of problems for which this elimination scheme is particularly simple is the class of problems in which each experiment can only generate two possible outcomes, that is, $|{\cal X}_{\cal E}|=2$ for all ${\cal E} \in \mathscr{E}$. This is an important special case given the popularity of pairwise comparison methods (see \citealp{szorenyi2015online}, \citealp{Heckel2016} and references therein). In this case,  $Z(\delta,{\cal E}_1) \leq_{cx} Z(\delta,{\cal E}_2)$  if and only if
the range of $Z(\delta,{\cal E}_1)$ is contained in the range of $Z(\delta,{\cal E}_2)$. But this is the same as requiring that the range of the likelihood ratio ${\cal L}({\cal E}_1)$ is contained in the range of ${\cal L}({\cal E}_2)$, which is a condition that is independent of $\delta$ and one can check efficiently. For example, if we apply this elimination scheme to the special instance in Example~\ref{ex1}, we get that Experiments 1, 8 and 9 can be eliminated. To see this, note that the likelihood ratio of Experiment 1 takes values ${\cal L}({\cal E}_1)\in\{0.3, 1.078\}$ while the likelihood ratio  for Experiment 2 takes values ${\cal L}({\cal E}_2)\in\{0.2, 1.2\}$. It follows that Experiment 2 dominates Experiment 1. (Similar calculations reveal that Experiments 7 dominates Experiments 8 and 9.)
\vspace{0.2cm}

We can use stochastic dominance one step further to derive a simple experimentation policy. Since $Z(\delta,{\cal E}_2) \leq_{cx} Z(\delta,{\cal E}_1)$ implies that $\var[Z(\delta,{\cal E}_2)]\leq \var[Z(\delta,{\cal E}_1)]$, we can implement a  heuristic policy which for each value of $\delta$ selects the experiment ${\cal E}^{\Heu}(\delta)$ that maximizes  $\var[Z(\delta,{\cal E})]$. But  $\var[Z(\delta,{\cal E})]=\e[\eta^2(\delta,x,{\cal E})]$ and so this heuristic policy reduces to
$${\cal E}^{\Heu}(\delta)  =\argmax_{{\cal E} \in \mathscr{E}} \Big\{\e\Big[\eta^2(\delta,x,{\cal E})\Big]\Big\}.$$
In the following sections we will show that this simple experimentation policy is indeed optimal in an appropriate asymptotic regime in which  the magnitude of the jumps $\eta^2(\delta,x,{\cal E})$ converges uniformly to zero, i.e., in a regime in which each experiment becomes less and less informative.\vspace{0.2cm}

We can also use the convexity of the value function to gain some intuition about this asymptotic result.  
Indeed, if we assume that the value function is twice-continuously differentiable, then a second order expansion of $\Pi(\delta)$ leads to the following  equality:
$$\Pi\big(\delta+\eta(\delta,x,{\cal E})\big)-\Pi(\delta)=\dot{\Pi}(\delta)\, \eta(\delta,x,{\cal E}) + {1 \over 2}\,\ddot{\Pi}(\delta)\,(\eta(x,\delta,{\cal E}))^2+O((\eta(\delta,x,{\cal E}))^3).$$ The optimal experiment is characterized in equation $(\ref{eq:Opt_Exp})$, which we write here as follows, $${\cal E}^*(\delta) =  \argmax_{{\cal E} \in \mathscr{E}} \Big\{\ed\Big[\Pi\big(\delta+\eta(\delta,x,{\cal E})\big)-\Pi(\delta)\Big]\Big\}.$$  By Lemma~\ref{lem:SDEdelta}, $\delta_t$ is a martingale and so we have that $\ed[\eta(\delta,x,{\cal E})]=0$. Also, by convexity of the  value function, $\ddot{\Pi}(\delta) \geq 0$. Combining these two observations together with the assumption that the magnitude of $\eta(\delta,x,{\cal E})$ is uniformly small, we 
conclude that
$${\cal E}^*(\delta) =  \argmax_{{\cal E} \in \mathscr{E}} \ed\Big[\eta^2(x,\delta,{\cal E})\Big].$$


\section{Asymptotic Approximation}\label{sec:AsymptoticForm}
\setcounter{footnote}{1}

In this section we specialize the problem described in the previous section to a particular class of instances in which  (i) experiments are conducted at ``high frequency" 
while (ii) the ``informativeness'' of each experiment is low.  There are many natural and practical situations in which the decision maker has  access to a large number of experiments, but where the informativeness of each individual one is low. For instance, online experiments are becoming quite common in the business world where each experiment is often linked to one visitor who is offered a set of choices to select from. Such common setup generates a large volume of experiments in a relatively short time period. However,  one of the major issues faced by the experimenter is the relevance and veracity of the data generated (we refer the reader to the section  ``Beware of Low-Quality data" in \citealp{Kohavi17}). 
In some cases, the heterogeneity of the experementees in online experimentation can generate very noisy data. Moreover, the hypotheses being tested can be marginally different making the task of distinguishing them harder. Both settings are typical and are examples of how little informative online experiments can be. Our illustrative example in Section~\ref{sec:Crowdvoting} build on these ideas. The notion of limited informativeness of experiments is also present in other settings. In their recent work,  \citealp{LewisRao15} show how difficult it is to prove the return on investment of advertising campaigns. The paper notes specifically that informative advertising experiments can require more than 10-million person-weeks, which reflects exactly the tension of our regime between large sample size and little informativeness. Clinical trials is another major area that suffers from serious data error and lack of accuracy, see, \citealp{Nahm12} and as a result would require  large sample sizes.

\subsection{Diffusion Formulation} 
To formalize the notion of a ``{\em high frequency  vs. low informativeness}''  regime of experimentation, we consider a sequence of instances of the problem indexed by a non-negative integer $k$ in such a way that as $k$ grows large both the number of experiments conducted per unit of time becomes high and the  `amount' of information generated from each individual experiment goes to zero. Under our proposed scaling the magnitude of the jumps of $\delta_t$ as well as the time between two experiments converge to zero resulting in a belief process that  converges weakly to a diffusion process.\vspace{0.2cm}

We let $Q^k(x,{\cal E},\theta)$ be the conditional probability of observing outcome $x \in {\cal X}_{\cal E}$ when experiment ${\cal E} \in \mathscr{E}$ is conducted conditional on  $\Theta=\theta$ for the $k^{\mbox{\tiny th}}$ instance of the problem. To capture the notion of low informativeness of an experiment, we impose the following requirement on the sequence  $\{Q^k(x,{\cal E},\theta)\}$.


\begin{Assum}\label{assm:asymregime}{\rm(Low Informativeness Regime)} For each ${\cal E}\in \mathscr{E}$, there exists a probability distribution ${\cal Q}(\cdot,\cal E)$ such that for $\theta \in \{\thetaL,\thetaH\}$

\begin{equation}\label{eq:asymptQ}
\sqrt{k}\,\left(\frac{Q^k(x,{\cal E},\theta)}{{\cal Q}(x,{\cal E})}-1\right)\longrightarrow \alpha (x,{\cal E},\theta),
\end{equation}
where $\alpha(x,{\cal E},\theta)$ satisfies
$$\sum_{x \in {\cal X}_{\cal E}} \alpha(x,{\cal E},\theta)\,{\cal Q}(x,{\cal E}) =0.$$
\end{Assum}

Intuitively, the asymptotic scaling in Assumption~\ref{assm:asymregime} has the following property:  as $k \to \infty$, the likelihood function ${\cal L}^k(x,{\cal E})=Q^k(x,{\cal E},\thetaH)/Q^k(x,{\cal E},\thetaL)$ converges to one for every $x \in {\cal X}_{\cal E}$ and as a result the jumps $\eta^k(\delta,x,{\cal E})$ of $\delta_t$  (see, Lemma~\ref{lem:SDEdelta})  converge to zero.  In other words, in this asymptotic regime, the outcomes of an experiment become  less and less informative as $k$ grows large. \vspace{0.2cm}


On its own,  the scaling in equation \eqref{eq:asymptQ} would lead to a trivial limit in which $\delta_t$ remains constant over time. To counterbalance the fact that individual experiments become less informative under \eqref{eq:asymptQ}, we also scale up the arrival rate of $N_t$ in a way that the `amount' of information collected by the experimentation process per unit of time remains comparable to the one in the original unscaled system. Specifically,  let $N^k_t$ denote the Poisson process that determines the experimentation epochs for the $k^{\mbox{\tiny th}}$ instance.

\begin{Assum}\label{assum:HighFrequency} {\rm (High  Frequency Regime)}  Let  $\Lambda^k$ be the intensity of $N^k_t$. Then, $\Lambda^k$ satisfies:\begin{equation}\label{eq:Lamdaasymptotic}
\Lambda^k=k\,\Lambda,
\end{equation}
for some fixed constant $\Lambda>0$.
\end{Assum}

Our objective at this point is to suggest a diffusion approximation of the general formulation problem (\ref{eq:probformobfnew2}). For that, we combine the parameter scalings in \eqref{eq:asymptQ} and \eqref{eq:Lamdaasymptotic} to obtain a well-defined diffusion limit for the belief process, $\delta_t$\footnote{In Sections \ref{sec:heuristic} and \ref{subsec:AsympApproxMNL} we show how to interpret and operationalize our asymptotic scaling in practical settings.}. We derive this limit over the class of continuous randomized Markovian policies defined below and show that this class contains an $\varepsilon$-optimal policy for any $\varepsilon>0$ (see, Proposition~\ref{prop:L1convergence}).


\vspace{0.5cm}



 In the following definition, 
$\Delta\big({\mathscr E}\big)$ is the set of probability distributions on the collection of possible  experiments in $\mathscr{E}$ and ${\cal M}_c(\Delta\mathscr{E})$ is the set of continuous measurable functions from $[0,1]$ to $\Delta\big({\mathscr E}\big)$. For a randomized experimentation policy  $\pi \in {\cal M}_{c}(\Delta\mathscr{E})$,  we let $\pi(\delta,{\cal E})$ denote the probability of selecting experiment ${\cal E}$, which is continuous in $\delta$ for all ${\cal E} \in \mathscr{E}$.

\begin{definition}\label{def:RandomMarkovpolicy} {\rm (Continuous Randomized Markovian Policy)} A continuous randomized Markovian  policy  is a pair $(\pi, {\cal I}) \in {\cal M}_c(\Delta\mathscr{E}) \times {\cal B}$.  For all $\delta \not\in {\cal I}$, the DM displays experiment ${\cal E} \in \mathscr{E}$ with  probability $\pi(\delta,{\cal E})$. On the other hand, for $\delta \in {\cal I}$ the decision maker chooses to stop the experimentation process and implements an optimal action $a \in {\cal A}^*(\delta) $.  
\end{definition}

Next, we move to state our limiting result for the belief process under the scalings given in \eqref{eq:asymptQ} and \eqref{eq:Lamdaasymptotic}.
\begin{prop}\label{prop:weakconv}  Consider a fixed experimentation policy $\pi \in {\cal M}_c(\Delta\mathscr{E})$ and  let $\delta_t^k$ be the belief process induced by $\pi$ for instance $k$ under the scaling in equations \eqref{eq:asymptQ} and \eqref{eq:Lamdaasymptotic}.
Then,  we have that $\delta_t^k \Rightarrow \tilde{\delta_t}$ as $k\rightarrow\infty$, where $\tilde{\delta_t}$ is a diffusion process solution of the  SDE
  $$\D\, \tilde{\delta_t} = \tilde{\sigma}(\tilde{\delta}_t,\pi)\,\tilde{\delta}_t\,(1-\tilde{\delta}_t)\,\D W_t,$$
where $W_t$ is a  Wiener process and,
\begin{equation}\label{eq: volatility} \tilde{\sigma}^2(\delta,\pi):=\Lambda\,\sum_{{\cal E} \in \mathscr{E}} \sum_{x \in {\cal X}_{\cal E}} \pi(\delta,{\cal E})\, \big(\alpha(x,{\cal E},\thetaH)-\alpha(x,{\cal E},\thetaL)\big)^2\,{\cal Q}(x,{\cal E}).\end{equation}
\end{prop}

\begin{rem} {\sf Throughout the paper, we use tildes (`$\sim$') to denote quantities that are related to the asymptotic approximation.} $\Diamond$
\end{rem}

The next result shows that restricting our attention to continuous randomized policies  is without a significance loss of optimality in the sense of the $L^1$ norm.  

\begin{prop}\label{prop:L1convergence}
Let $(\pi, {\cal I}) \in {\cal M}(\mathscr{E}) \times {\cal B}$ be an optimal Markovian policy with a corresponding value function $\Pi(\delta)$. For any $\varepsilon>0$, there exists a continuous randomized policy $(\pi_c,{\cal I}_c)$ in  ${\cal M}_{c}(\Delta\mathscr{E}) \times {\cal B}$ with expected payoff function $\widehat{\Pi}(\delta)$ such that,
$$\|\Pi - \widehat{\Pi}\|_1<\varepsilon,$$
where $\|\cdot\|_1$ is the $L^1$ norm in $[0,1]$.
\end{prop}

In sum, and in light of  Proposition~\ref{prop:weakconv} and \ref{prop:L1convergence},  we suggest the following diffusion-asymptotic approximation of the decision maker's  problem given in (\ref{eq:probformobfnew2}):



 
\begin{equation}\label{eq:diffusionprobformobf1}
 \widetilde{\Pi}(\delta)=\sup_{(\pi,{\cal I}) \in {\cal M}_{c}(\mathscr{E}) \times {\cal B}}\ed\Big[e^{-r\,\tau}\,G(\tilde{\delta}_\tau)\Big]\quad
\mbox{s.t.} \quad \D \tilde{\delta_t} = \tilde{\sigma}(\delta,\pi)\,\tilde{\delta}_t\,(1-\tilde{\delta}_t)\,\D W_t \quad \mbox{and}\quad \tau=\inf\big\{t>0 \colon \tilde{\delta}_t \in {\cal I}\big\}.
\end{equation}

We can view problem  \eqref{eq:diffusionprobformobf1} as  having two decision variables, namely, the experimentation policy $\pi$ and the intervention region ${\cal I}$. Interestingly, it turns out that we can decouple the optimization of these two decisions, in particular  we can  solve for the optimal experimentation  $\pi^*$ without computing  explicitly ${\cal I}^*$. Surprisingly, this implies that the choice of an optimal experiment is independent of the intervention region, and thus of how long the decision maker decides to run the experimentation process. We formalize this observation in the following section.\vspace{0.2cm}


\subsection{Asymptotically Optimal Experimentation Policy}\label{sec:displayset}
From the diffusion approximation in equation  \eqref{eq:diffusionprobformobf1}, one can easily see that the impact of an experimentation policy $\pi$ has on the decision maker's optimization problem is channelled only through the volatility of the belief process $\tilde{\sigma}(\delta,\pi)$. We use this fact to derive a rather simple solution to the problem of selecting an asymptotically optimal policy $\pi^{\Asym}$. (The superscript `A' is mnemonic of Asymptotic). To this end, let us define the mapping
$$T_t^\pi: = \int_0^t {1 \over \tilde{\sigma}^2(\delta_s,\pi)}\, \D s,$$
which acts as a random time change in the following proposition.

\begin{prop}\label{lem:timechange} The optimization problem in \eqref{eq:diffusionprobformobf1} is equivalent to 
$$
 \widetilde{\Pi}(\delta)=\sup_{(\pi,{\cal I}) \in {\cal M}_{c}(\mathscr{E}) \times {\cal B}}\ed\Big[e^{-r\,T_\tau^\pi}\,G(\tilde{\delta}_\tau)\Big]\quad
\mbox{s.t.} \quad \D \tilde{\delta_t} =\tilde{\delta}_t\,(1-\tilde{\delta}_t)\,\D W_t \quad \mbox{and}\quad \tau=\inf\big\{t>0 \colon \tilde{\delta}_t \in {\cal I}\big\}.
$$

\end{prop}

The previous result provides an alternative interpretation of the effect an experimentation policy $\pi$ has on the decision maker's performance. According to Proposition~\ref{lem:timechange}, a policy $\pi$ impacts only the discount factor, $r\,T_\tau^\pi$ that the decision maker uses to penalize the time value of money. The following corollary follows directly from this observation.

\begin{Cor}\label{cor:maxsigma}  {\rm (Maximum Volatility)}  For any stopping set ${\cal I} \in {\cal B}$, an optimal asymptotic experimentation policy  $\pi^{\Asym}$ minimizes the modified discount factor $r\,T_\tau^\pi$ pathwise, or equivalently, maximizes pointwise the  belief process's volatility $\tilde{\sigma}^2(\delta,\pi)$. Thus, from \eqref{eq: volatility}, we conclude that we can select $\pi^{\Asym}$ to be a static experimentation policy, namely, $\pi^{\Asym}(\delta, {\cal E})=\Indi\big({\cal E}=\widetilde{\cal E}^{\Asym}\big)$ for all $\delta \in {\cal I}^c$ where  $\widetilde{\cal E}^{\Asym}$ is given by
$$\widetilde{\cal E}^{\Asym}= \argmax_{{\cal E} \in \mathscr{E}}\left\{\sum_{x \in {\cal X}_{\cal E}} \big(\alpha(x,{\cal E},\thetaH)-\alpha(x,{\cal E},\thetaL)\big)^2\,{\cal Q}(x,{\cal E})\right\}.$$

(If there are multiple experiments that maximize the expression inside the brackets then we can select any static experimentation policy that uses an experiment  $\widetilde{\cal E}^{\Asym}$ from the argmax set.)
\end{Cor}

A few remarks about this result are in order. First, Corollary~\ref{cor:maxsigma} confirms our previous claim that an optimal experimentation policy is independent of the choice of the stopping set ${\cal I}$ and so we can effectively decouple the problem of determining an optimal experimentation strategy and that of when to stop experimenting. We also note that an optimal static experimentation strategy is continuous in $\delta$ and so we can invoke the weak convergence in Proposition~\ref{prop:weakconv}  directly to $\pi^{\Asym}(\delta, \widetilde{\cal E}^{\Asym})$.

{\setstretch{1}
\setcounter{exm}{1}
\begin{exm}{\rm (Example~\ref{ex1} Revisited)} {\it To illustrate the result in Corollary~\ref{cor:maxsigma}, let us revisit the  instance in Example~\ref{ex1} in the context of the asymptotic regime. To this end, suppose the probabilities $Q^k(0,{\cal E},\theta)$ for the $k^{\mbox{\tiny th}}$ instance of the problem are equal to

\begin{table}[h]{\small
\begin{center}
$Q^k(0,{\cal E},\theta)={\cal Q}(0,{\cal E})\,\left(1+{\alpha(0,{\cal E},\theta) \over \sqrt{k}}\right),\quad \mbox{where}\quad$
 \begin{tabular}{||l||c|c|c|c|c|c|c|c|c|c||}  \hline
Experiment &  2 & 3 & 4 & 5 & 6 & 7   \\ \hline \hline
${\cal Q}(0,{\cal E})$ &  0.2	& 0.3 &	0.4 &	 0.5 & 0.6 & 	0.7   \\  \hline
$\alpha(0,{\cal E}, \thetaL)$ &  0.0 &	0.0 &	0.0 &	0.0 &	0.0 & 	0.0   \\  \hline
$\alpha(0,{\cal E}, \thetaH)$ &  -0.8 & -0.7 &	-0.6 &	-0.5 &	-0.4 & 	-0.3   \\  \hline
\end{tabular}.
\end{center}\label{Table:ex2}}
\end{table}

Note that we are not including Experiments 1, 8 and 9 since they are dominated (see Section~\ref{rem:stochdominance}).
Also, the original probabilities in Table~\ref{ex1} correspond to the case $k=1$.  Figure~\ref{fig:ExValueItera2} mimics Figure~\ref{fig:ExValueItera} but for $k=10,000$.

\begin{figure}[htb]
    \begin{center}
    \includegraphics[width=14cm]{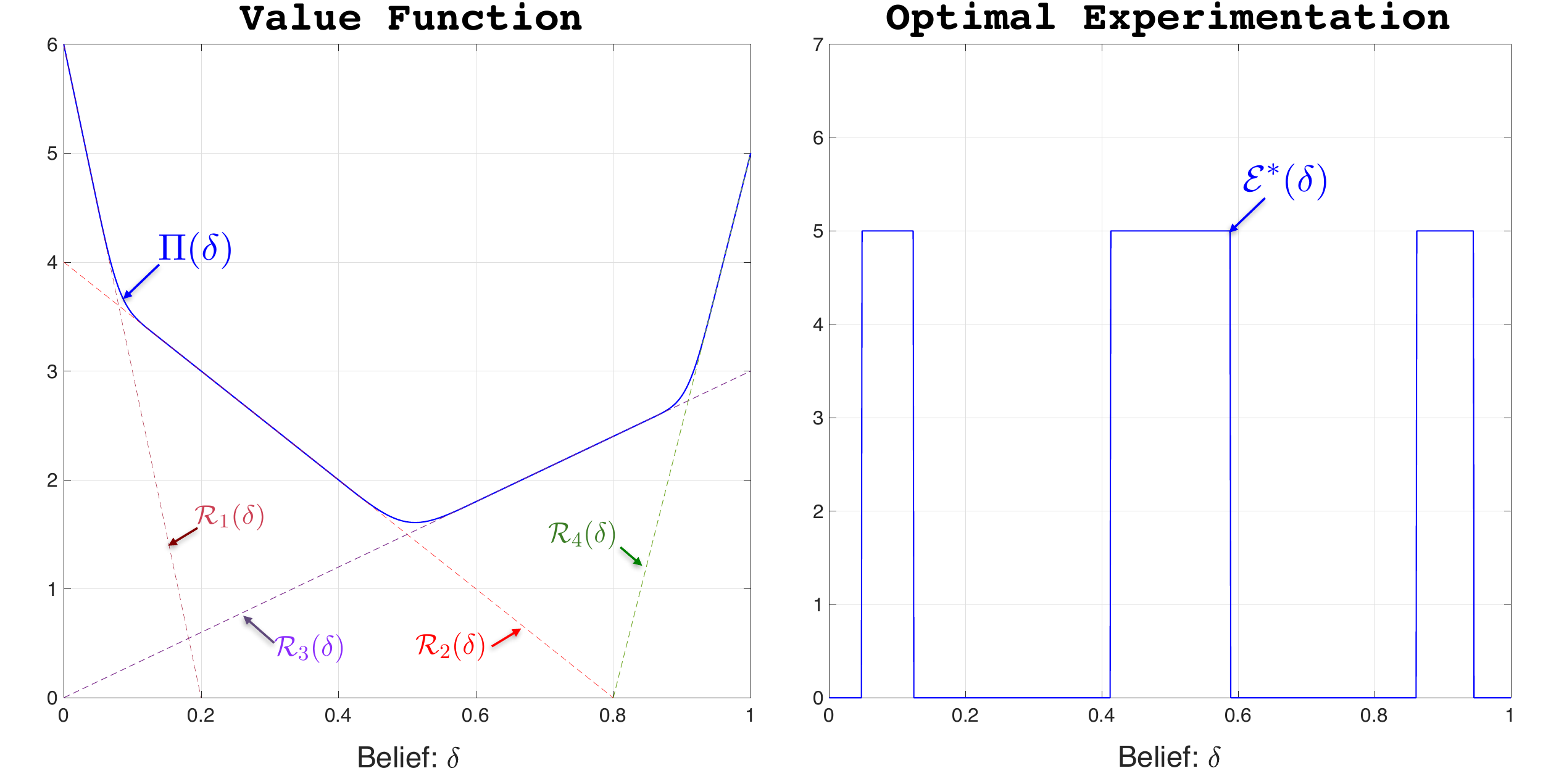}\vspace{-0.5cm}
    \end{center}\caption{\footnotesize \sf {\sc Data}: ${\cal R}_1(\delta)=6-30\,\delta$, ${\cal R}_2(\delta)=4-5\,\delta$, ${\cal R}_3(\delta)=3\,\delta$, ${\cal R}_4(\delta)=-20+25\,\delta$, $r=0.5$,  $\Lambda=8\,k$ and $k=10,000$. }    \label{fig:ExValueItera2}
\end{figure}
Consistent with the result in Corollary~\ref{cor:maxsigma}, for $k$ sufficiently large, the optimal experimentation strategy ${\cal E}^{\Asym}(\delta)$ consists of a single experiment independent of $\delta$, which in this case corresponds to Experiment 5. One can check that Experiment 5 maximizes the instantaneous volatility of the belief process. $\Diamond$

}
\end{exm}}


\subsection{Optimal Stopping of Experimentation}\label{sec:Duration}
\setcounter{footnote}{1}

Let us now turn to the problem of determining the optimal intervention region in the asymptotic regime under consideration. In what follows, we assume that an optimal experimentation policy has been selected based on Corollary~\ref{cor:maxsigma}. That is, we focus on solving problem \eqref{eq:diffusionprobformobf1}  given the optimal experimentation policy $\pi^{\Asym}(\delta, {\cal E})=\Indi\big({\cal E}=\widetilde{\cal E}^{\Asym}\big)$. We find convenient to rewrite this problem using the following optimal stopping time formulation:
\begin{equation}\label{eq:diffusionprobformobf2} \widetilde{\cal G}(\delta):=\sup_{\tau \in \mathbb{T}}
\ed\left[e^{-r\,\tau}\,G(\tilde{\delta}_\tau)\right] \qquad
\mbox{subject to} \qquad\D \,\tilde{\delta_t} = \tilde{\sigma}\,\tilde{\delta}_t\,(1-\tilde{\delta}_t)\,\D W_t, \qquad \tilde{\delta}_0=\delta.
\end{equation}
For notational convenience, throughout this section we suppress the dependence of $\widetilde{\cal G}$ and $\tilde{\sigma}$ on the display set $\widetilde{\cal E}^{\Asym}$ since it remains fixed.\vspace{0.2cm}

We approach the problem in two steps. First, we derive optimality conditions in the form of a set of partial differential inequalities that characterize the optimal stopping time. Then, we use these inequalities to characterize an optimal solution and the corresponding payoff.

\subsubsection{Quasi-Variational Inequalities}
Let ${\cal C}^k[0,1]$ denote the set of real-valued continuous functions on $[0,1]$ having  derivatives of order $k\geq 0$. We define also the set
\begin{equation}\label{def:C2}
\widehat{\cal C}^2:= \Big\{f \in {\cal C}^1[0,1]\;:\;\mbox{there exists a finite set $N_f \subseteq [0,1]$ such that  $f''(\delta)$ exists  $\forall \delta \in [0,1] \setminus N_f$} \Big\}.
\end{equation}
(Note that the set $N_f$ depends on the specific function $f$.) We also define the operator $\mathcal{H}$ on $\widehat{\cal C}^2$ as follows
\begin{equation}\label{eq:defH}{\cal H}f(\delta):={1 \over 2}\,\tilde{\sigma}\,\delta\,(1-\delta)\,f''(\delta)-r\,f(\delta), \qquad \mbox{for all } \delta \in [0,1] \setminus N_f.
\end{equation}

\begin{definition}{\sc (QVI)} \label{def:QVI}The function $f\in \widehat{\cal C}^2$ satisfies the  quasi-variational inequalities for the optimization problem \eqref{eq:diffusionprobformobf2}, if for all $\delta \in [0,1] \setminus N_f$
\begin{eqnarray}\nonumber
& & f(\delta)-G(\delta) \geq 0  \\
& & \mathcal{H}f(\delta) \leq 0   \label{eq:QVI}\\
& & \big(f(\delta)-G(\delta)\big)\,\mathcal{H}f(\delta)=0.\qquad \Box\nonumber
\end{eqnarray}
\end{definition}
As one might expect, a solution to these QVI conditions partition the interval $[0,1]$ into two regions: a {\em continuation} region in which the firm's optimal strategy is to keep experimenting and an {\em intervention} region in which stopping the experimentation process is optimal.
\begin{eqnarray*}
\mbox{\sf Continuation:}& & \mathcal{C}:=\big\{\delta \in [0,1]: f(\delta) > G(\delta) \quad \mbox{and}\quad \mathcal{H}f(\delta) = 0\big\} \\
\mbox{\sf Intervention:} & & \mathcal{I}:=\big\{\delta \in [0,1]: f(\delta) = G(\delta) \quad \mbox{and}\quad \mathcal{H}f(\delta) \leq  0\big\}
\end{eqnarray*}
For every solution of the QVI  conditions we can associate a control $\tau \in \mathbb{T}$.
\begin{definition} \label{def:control}Let $f \in \widehat{\cal C}^2$ be a solution of the QVI conditions in \eqref{eq:QVI}. We define the control $\tau$ as follows $$\tau =\inf\big\{t >0 : f(\tilde{\delta}_t)=G(\tilde{\delta}_t)\big\}$$ and refer to it as the QVI-control associated to $f$.
\end{definition}

We are now ready to formalize the \textit{verification} theorem that provides the connection between the QVI  conditions and the original optimization problem in  \eqref{eq:diffusionprobformobf2}.

\begin{theorem}{\sc (Verification)}\label{thm:verification} Let $f\in\widehat{\cal C}^2$ be a  solution of the QVI in \eqref{eq:QVI}. Then,
$$f(\delta) \geq \widetilde{\cal G}(\delta)\quad \mbox{for every $\delta \in [0,1]$}.$$
In addition, if there exists a QVI-control $\tau$ associated with $f$ such that $\e[\tau] <\infty$,  then $\tau$  is optimal and \mbox{$f(\delta) = \widetilde{\cal G}(\delta)$}.
\end{theorem}
This verification theorem reduces the problem of determining the value function $\widetilde{\cal G}(\delta)$ to that of solving the QVI equations defined above. In order to find a solution, we take full advantage of the fact that the payoff function $G(\delta)$ is a piecewise linear continuous function of $\delta \in [0,1]$ (see  equation \eqref{eq:payoff}). Moreover, an important  building block in our methodology is the solution to a special case in which $G(\delta)$ has only two linear pieces, that is, the set $\mathscr{A}$ includes only two actions. We will focus on this simpler case first and then show how to leverage this solution and extend it to the general case in which $\mathscr{A}$ includes an arbitrary number of actions.

\subsubsection{Special Case: $|\mathscr{A}|=2$}\label{subsec:O=2}

Suppose the set of actions is given by $\mathscr{A}=\{a_i,a_j\}$ for two distinctive actions $a_i$ and $a_j$.  Let us denote by $\widetilde{G}_{ij}(\delta)=\max\big\{\widetilde{\cal R}_i(\delta),\widetilde{\cal R}_j(\delta)\big\}$, where $\widetilde{\cal R}_n(\delta)=\e_\delta[{\cal R}(a_n,\Theta)]$ for $n=i,j$ (see equation~\eqref{eq:payoff}).  Without loss of generality, we will assume that $\widetilde{G}_{ij}(\delta) \geq 0$ for all $\delta \in [0,1]$. Let us denote by $\hat{\delta}_{ij}$ the value of the belief at which $\widetilde{\cal R}_i(\hat{\delta}_{ij})=\widetilde{\cal R}_j(\hat{\delta}_{ij})$. (Recall that we have assumed that there is no action in the set $\mathscr{A}$ that is uniformly dominated and this assumption guarantees the existence of $\hat{\delta}_{ij} \in [0,1]$.)\vspace{0.2cm}

\begin{figure}[htb]
    \begin{center}
    \includegraphics[width=9cm]{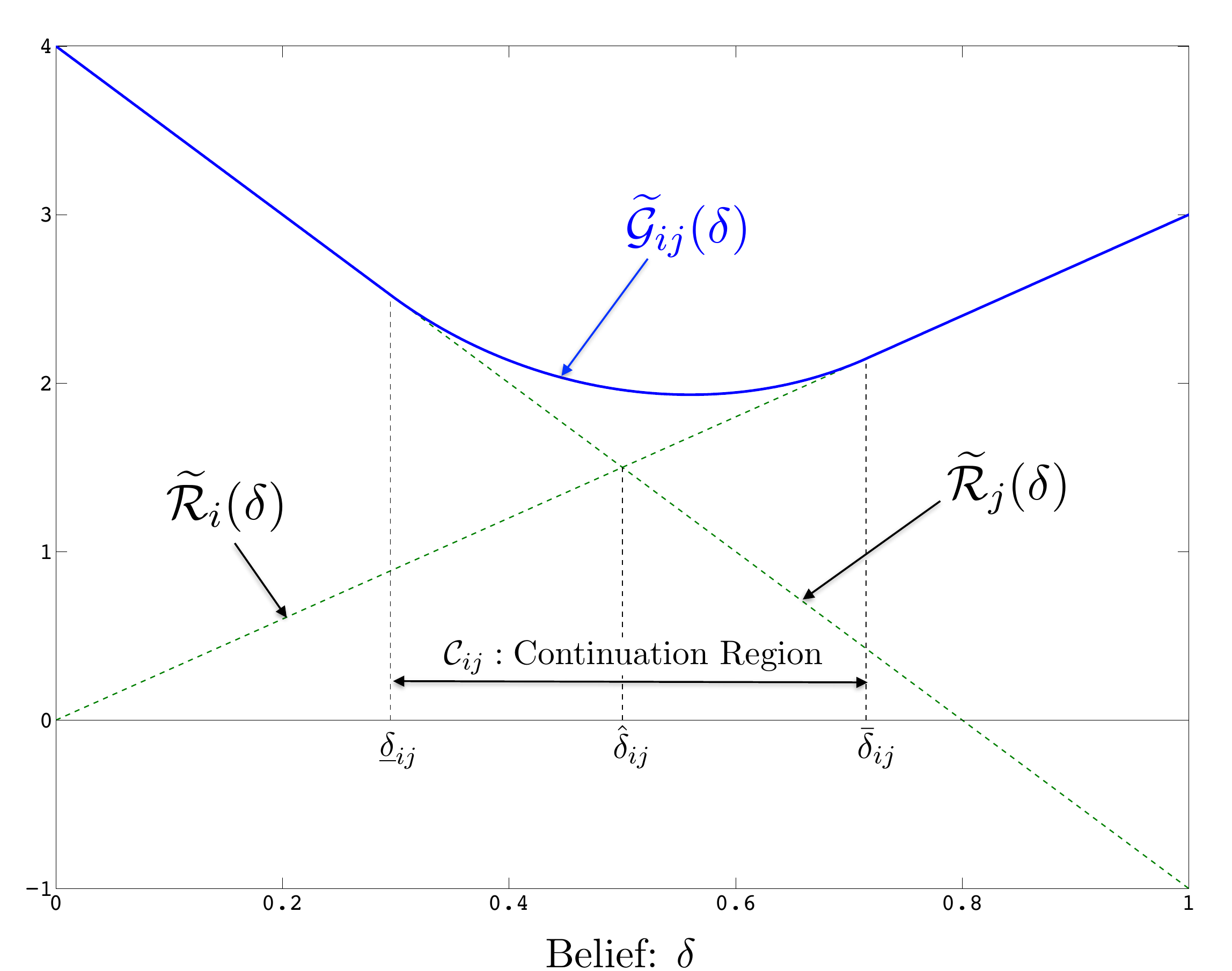}\vspace{-0.5cm}
    \end{center}\caption{\footnotesize \sf Example of the  value function $\widetilde{\cal G}_{ij}(\delta)$ and continuation region ${\cal C}_{ij}$ for the case in which $\mathscr{A}$ includes two actions. \newline {\sc Data}: $\widetilde{\cal R}_i(\delta)=3\,\delta$, $\widetilde{\cal R}_j(\delta)=4-5\,\delta$, $r=1$ and $\tilde{\sigma}=2$. }    \label{fig:ExTwoPieces}
\end{figure}

To solve the QVI conditions in this special case we take ``an educated guess'' approach and assume that the continuation region ${\cal C}_{ij}$ is given by an interval $[\deltau_{ij},\deltab_{ij}]$, for two thresholds $0 \leq \deltau_{ij} \leq \deltab_{ij} \leq 1$. Furthermore, we assume the intuitive fact that $\hat{\delta}_{ij} \in [\deltau_{ij},\deltab_{ij}]$. To illustrate, consider the example in Figure~\ref{fig:ExTwoPieces} that depicts the value function  $\widetilde{\cal G}_{ij}(\delta)$ as well as the payoff functions $\widetilde{\cal R}_i(\delta)$ and $\widetilde{\cal R}_j(\delta)$ for products $i$ and $j$.

By definition, in the interior of the continuation region we have that $\widetilde{G}_{ij}(\delta) <\widetilde{\cal G}_{ij}(\delta)$.  Hence, according to the third QVI condition, in this region the value function must satisfy ${\cal H}\widetilde{\cal G}_{ij}(\delta)=0$. This is a second-order differential equation
$${(\tilde{\sigma}\,\delta\,(1-\delta))^2 \over 2} \, \widetilde\cG_{ij}''(\delta) -r\,\widetilde\cG_{ij}(\delta) =0,$$

whose general solution is given by
\begin{equation}\label{eq: Two linear formulation}\widetilde\cG_{ij}(\delta) = C_{ij}^0\,{(1-\delta)^{\gamma} \over {\delta}^{\gamma-1}} + C_{ij}^1 {{\delta}^{\gamma} \over (1-{\delta})^{{\gamma}-1}}, \qquad \mbox{where } \gamma:={1+\sqrt{1+8r/\tilde{\sigma}^2} \over 2},
\end{equation}
and  $C_{ij}^0$ and $C_{ij}^1$ are two constants of integration.

To complete our proposed characterization of the value function, we need to determine the constants of integration as well as the two thresholds $\deltau_{ij}$ and $\deltab_{ij}$ that define the continuation region. To do that we impose the so-called  value-matching and smooth-pasting conditions that regulate  the behavior of the value function at the boundaries between the intervention  and continuation regions. Specifically, we impose the conditions
\begin{equation}\label{eq:value&smooth}
\widetilde\cG_{ij}(\delta)=\widetilde{G}_{ij}(\delta) \qquad \mbox{and}\qquad \widetilde\cG_{ij}'(\delta)=\widetilde{G}_{ij}'(\delta)\qquad \mbox{for }\delta=\deltau_{ij},\deltab_{ij}.\end{equation}

We formalize our previous discussion in the next proposition.

\begin{prop}\label{prop:diffusion_2}
Let $\gamma=(1+\sqrt{1+8r/\tilde{\sigma}^2})/2$. If $\mathscr{A}=\{a_i,a_j\}$ then the QVI conditions admit a solution that we denote by $\tilde\cG_{ij}(\cdot)$ given as follows
\begin{equation}\label{eq: Close_form_general}\widetilde\cG_{ij}({\delta})=\left\{\begin{array}{cl}
\widetilde{G}_{ij}(\delta) & \mbox{if}\qquad 0 \leq {\delta} \leq \deltau_{ij} \\ & \\
C_{ij}^0\, (1-{\delta})^\gamma\, {\delta}^{1-\gamma}+C_{ij}^1\,(1-{\delta})^{1-\gamma}\,{\delta}^\gamma  & \mbox{if}\qquad   \deltau_{ij} \leq {\delta} \leq \deltab_{ij}  \\ & \\
\widetilde{G}_{ij}(\delta) & \mbox{if}\qquad  \deltab_{ij} \leq \delta\leq 1.\end{array} \right. \end{equation}\vspace{0.1cm}

where $\deltau_{ij},\deltab_{ij} \in (0,1)$ and $C_{ij}^0$ and $C_{ij}^1$ are positive constants all determined by imposing the value matching and smooth pasting conditions in \eqref{eq:value&smooth}. The function $\widetilde\cG_{ij}({\delta})$ is convex and in $\widehat{\cal C}^2$.
\end{prop}

The verification theorem guarantees that the solution expressed in Proposition~\ref{prop:diffusion_2} is such that $\tilde\cG_{ij}=\tilde\cG$ when $\mathscr{A}=\{i,j\}$. In terms of implementation, this solution corresponds  to the following policy. \vspace{0.3cm}
{\setstretch{1}
\begin{quotation}
\noindent {\sc Asymptotically Optimal Intervention Policy:} {\sf Suppose the initial belief lies in the interior of the continuation region ${\cal C}_{ij}$ that is, $\delta\in (\deltau_{ij},\deltab_{ij})$. In this case, the decision maker runs an experimentation process and keeps it running  as long as $\delta_t\in (\deltau_{ij},\deltab_{j})$. As soon as the belief process $\delta_t$ hits one of the two thresholds $\deltau_{ij}$ or $\deltab_{ij}$ then the experimentation process stops and the decision maker  selects the action that maximizes $\widetilde{G}_{ij}$ at that time. On the other hand, if the initial belief $\delta$ is not in the interior of the continuation region then no experimentation is needed and the decision maker  selects the action that maximizes $\widetilde{G}_{ij}(\delta)$ at time 0. $\Diamond$}
\end{quotation}}
\vspace{0.2cm}

The simple representation of the value function in Proposition~\ref{prop:diffusion_2} is due to the diffusion approximation obtained in this asymptotic regime. Moreover, this same diffusion approximation allows one to use some standard results for one-dimensional diffusion processes (e.g., Section 5.5 in  \citealp{KaratzasSchreve}) to analyze its optimal solution. For instance,  in those cases where $\delta\in (\deltau_{ij},\deltab_{ij})$ experimentation should be conducted, and its duration would correspond to the first exit time of $\delta_t$ from the interval $(\deltau_{ij},\deltab_{ij})$. The following corollary characterizes the expected duration of this experimentation phase as well as the likelihood that action $a_i$ or $a_j$ will be eventually selected.

\begin{Cor}\label{cor:diffusion} Suppose $\delta\in (\deltau_{ij},\deltab_{ij})$ and let  $\tau^* = \inf\{t>0 \colon \tilde{\delta}_t \not\in (\deltau_{ij},\deltab_{ij})\}$ and $\bar{p}(\delta)=\Pro(\tilde{\delta}_{\tau^*}=\deltab_{ij}\,|\,\delta_0=\delta)$. Then,
$$\bar{p}(\delta)={\delta-\deltau_{ij} \over \deltab_{ij}-\deltau_{ij}} \qquad \mbox{and}\qquad \e[\tilde{\tau}^*]=\bar{p}(\delta)\,\mathcal{T}(\deltab_{ij})+(1-\bar{p}(\delta))\,\mathcal{T}(\deltau_{ij})-\mathcal{T}(\delta),$$
where $\mathcal{T}(\delta)$ is the function
$$\mathcal{T}(\delta):={2 \over \tilde{\sigma}^2}\,(2\delta-1)\,\ln\left(\delta \over 1-\delta\right).$$
\end{Cor}

We conclude our discussion of this special case with $\|\mathscr{A}\|=2$ by exploiting the result in Proposition~\ref{lem:timechange} to derive upper and lower bounds for the value function.

\begin{prop} The value function  $\widetilde{\cal G}_{ij}$  satisfies

$$\widetilde{G}_{ij}(\delta) \leq \widetilde{\cal G}_{ij}(\delta) \leq  \widetilde{G}_{ij}(0)\,(1-\delta) +\widetilde{G}_{ij}(1)\,\delta\qquad \mbox{for all } \delta \in (0,1).$$
Furthermore,
$$\max_{\delta \in (0,1)} \Big\{\widetilde{\cal G}_{ij}(\delta) - \widetilde{G}_{ij}(\delta)\Big\} =
\widetilde{\cal G}_{ij}(\hat{\delta}_{ij}) - \widetilde{G}_{ij}(\hat{\delta}_{ij}).$$

\end{prop}

\subsubsection{General Case: $|\mathscr{A}|\geq 2$}\label{subsec:arbitraryO}

Let us now turn to the general case in which the set $\mathscr{A}$ includes an arbitrary but finite number of actions.
Our derivation of the value function $\widetilde\cG({\delta})$ in \eqref{eq:diffusionprobformobf2} will be obtained based on the solution derived in the previous section.  For each pair of actions $\{a_i,a_j\} \in \mathscr{A}$, the function $\widetilde\cG_{ij}({\delta})$ in Proposition~\ref{prop:diffusion_2}  is the value function of a problem in which only actions $a_i$ and $a_j$ are available. It follows that $\widetilde\cG({\delta}) \geq \widetilde\cG_{ij}({\delta})$ for all $\delta \in [0,1]$ and so
$$ \widetilde\cG({\delta}) \geq  \widetilde{V}(\delta):=\max_{\{a_i,a_j\} \in \mathscr{A}} \Big\{\widetilde\cG_{ij}({\delta}) \Big\},$$
where $\widetilde{V}(\delta)$ is the point-wise maximum of the functions $\widetilde\cG_{ij}({\delta})$. Our main result in this section establishes that the inequality is in fact an equality, that is, $\widetilde\cG({\delta}) =  \widetilde{V}(\delta)$.  To prove this, we will show that the function $\widetilde{V}(\delta)$ satisfies the QVI conditions so that we can invoke the verification Theorem~\ref{thm:verification}.
To this end, we first show that $\widetilde{V}(\delta)$ satisfies all three QVI conditions in \eqref{eq:QVI}.
\begin{prop}\label{prop:QVI-1} For all $\delta \in [0,1]$, we have that $\widetilde{V}(\delta) \geq \widetilde{G}(\delta)$.
Also, there exists a finite set $N_{\mbox{\tiny \rm $\widetilde{V}$}} \subseteq [0,1]$ such that  ${\cal H}\widetilde{V}(\delta) \leq 0$  and  $\big(\widetilde{V}(\delta)-\widetilde{G}(\delta)\big)\,\mathcal{H}\widetilde{V}(\delta)=0$ for all
$\delta \in [0,1]\setminus N_{\mbox{\tiny \rm $\widetilde{V}$}}$.
\end{prop}

The attentive reader might have noticed that the result in Proposition~\ref{prop:QVI-1} is not enough to invoke the verification Theorem~\ref{thm:verification}. The reason is that, besides verifying the QVI conditions, we also need to show that the function $\widetilde{V}(\delta)$ is  sufficiently
smooth and belongs to the set $\widehat{\cal C}^2$ (see equation~\eqref{def:C2}). We formalize this condition in the following result.

\begin{theorem}\label{thm:Vsmooth} The function $\widetilde{V}(\delta)=\max_{\{a_i,a_j\} \in \mathscr{A}} \big\{\widetilde\cG_{ij}({\delta}) \big\}$ is in
$\widehat{\cal C}^2$. As a result, $\widetilde\cG({\delta}) =  \widetilde{V}(\delta)$.
\end{theorem}

It is worth noticing that the previous theorem shows that  the complexity of the diffusion optimal stopping problem grows only quadratically with the cardinality of the action set $\mathscr{A}$. In fact, Theorem~\ref{thm:Vsmooth} reveals that solving a problem with $|\mathscr{A}|$ actions is equivalent to solving a collection of $|\mathscr{A}|\,(|\mathscr{A}|-1)$ problems each with only two actions.

\vspace{0.2cm}

\begin{figure}[htb]
    \begin{center}
    \includegraphics[width=16cm]{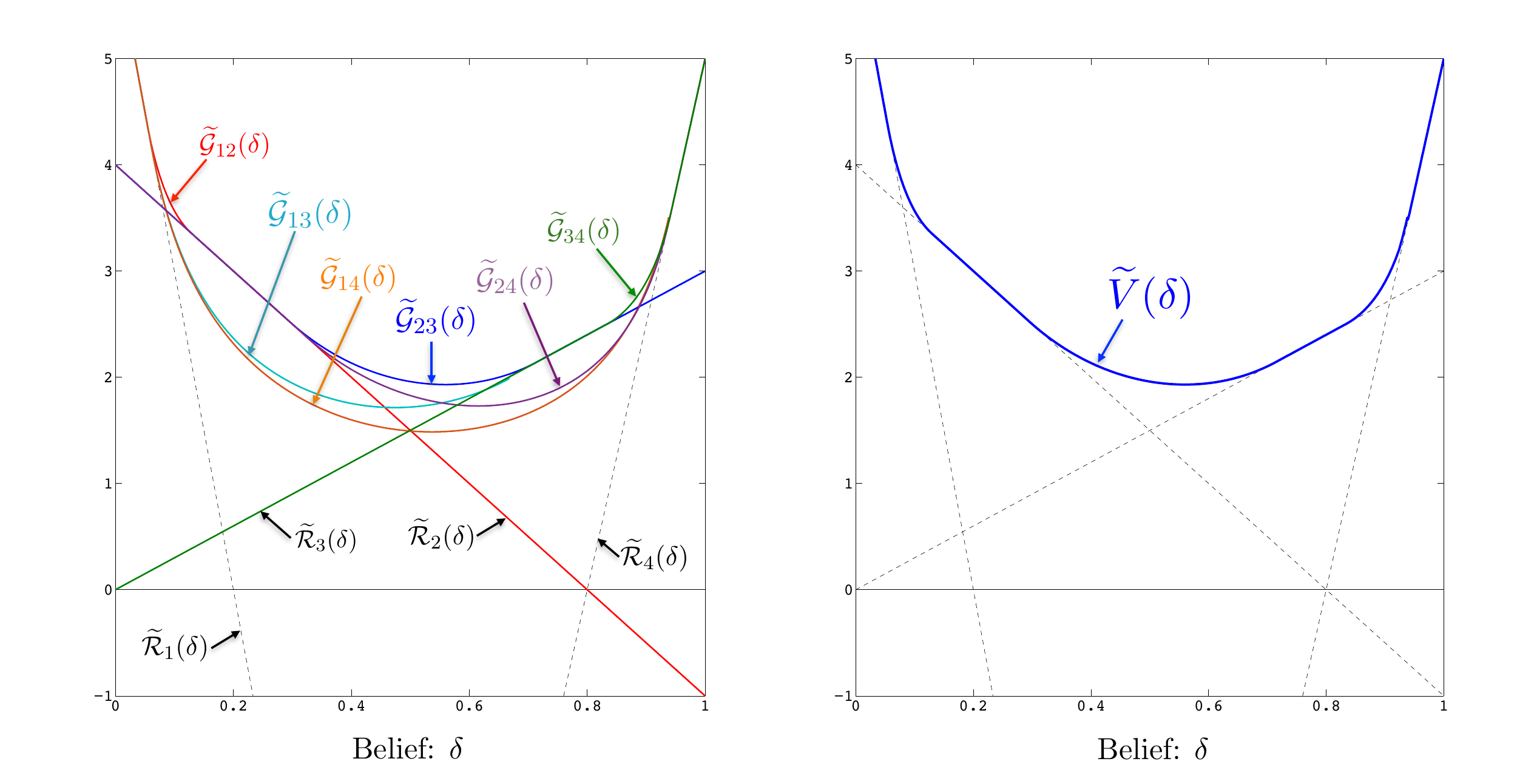}\vspace{-0.5cm}
    \end{center}\caption{\footnotesize \sf Example in which the offer set ${\cal O}$ includes four products. The left panel depicts the value functions $\widetilde\cG_{ij}({\delta})$ derived in Proposition~\ref{prop:diffusion_2}. The right panel depicts the function $V(\delta):=\max_{\{i,j\} \in {\cal O}} \big\{\widetilde\cG_{ij}({\delta}) \big\}$. {\sc Data}: ${\cal R}_1(\delta)=6-30\,\delta$, ${\cal R}_2(\delta)=4-5\,\delta$, ${\cal R}_3(\delta)=3\,\delta$, ${\cal R}_4(\delta)=-20+25\,\delta$, $r=1$ and $\tilde{\sigma}=2$. }     \label{fig:ExFourPieces}
\end{figure}

To illustrate the result in Theorem~\ref{thm:Vsmooth} and resulting optimal policy, let us consider the example in Figure~\ref{fig:ExFourPieces}  in which the  set $\mathscr{A}$ has four actions. The left panel depicts all six functions $\big\{\widetilde\cG_{ij}({\delta})\colon \{a_i,a_j\} \in \mathscr{A}\big\}$ while the right panel depicts the function  $\widetilde{V}(\delta):=\max_{\{a_i,a_j\} \in \mathscr{A}} \big\{\widetilde\cG_{ij}({\delta}) \big\}$.

After a quick inspection, we can check that in this example there exist two (non-unique) thresholds $\deltau$ and $\deltab$ such that
$$\widetilde{V}(\delta)= \left\{\begin{array}{cl} \widetilde\cG_{12}({\delta}) &\mbox{if } 0 \leq \delta \leq \deltau \\
\widetilde\cG_{23}({\delta}) &\mbox{if } \deltau \leq \delta \leq \deltab \\
\widetilde\cG_{34}({\delta}) &\mbox{if } \deltab \leq \delta \leq 1.
\end{array} \right.$$
Furthermore, at $\delta=\deltau$ the functions $\widetilde\cG_{12}(\delta)$ and $\widetilde\cG_{23}(\delta)$ meet smoothly since $\widetilde\cG_{12}(\deltau)=\widetilde\cG_{23}(\deltau)=\widetilde{\cal R}_2(\deltau)$. A similar smooth pasting occurs at $\delta=\deltab$ since $\widetilde\cG_{23}(\deltab)=\widetilde\cG_{34}(\deltab)=\widetilde{\cal R}_3(\deltab)$. As a result, since each of the functions $\widetilde\cG_{ij}({\delta})$ is in $\widehat{\cal C}^2$ by Proposition~\ref{prop:diffusion_2}, it follows that $\widetilde{V}(\delta)$ is also in $\widehat{\cal C}^2$. Note also that an optimal policy is given by a sequence of thresholds that define the   continuation and intervention regions.  In this example, we have that
\begin{eqnarray*}
\mbox{\sf Continuation:}& & \mathcal{C}^{\Asym}= (\deltau_{12},\deltab_{12}) \cup (\deltau_{23},\deltab_{23}) \cup (\deltau_{34},\deltab_{34})\\
\mbox{\sf Intervention:} & & \mathcal{I}^{\Asym}= [0, \deltau_{12}] \cup [ \deltab_{12},\deltau_{23}] \cup [\deltab_{23},\deltau_{34}] \cup [\deltab_{34},1],
\end{eqnarray*}
where the thresholds  $\deltau_{ij}$ and $\deltab_{ij}$ are defined in Proposition~\ref{prop:diffusion_2}.

\section{Non-Asymptotic Experimentation Policies}\label{sec:heuristic}

In this section we discuss how to interpret the asymptotic analysis developed in the previous section to construct experimentation policies that can be used in an arbitrary instance. Recall from Definition~\ref{def:Markovpolicy} that a policy consists of two components: (a) an intervention region ${\cal I}$ that defines the set of beliefs $\delta$ at which the decision maker stops the experimentation process and selects an optimal action $a^* \in {\cal A}^*(\delta)$, and (b) an experimentation policy $\pi(\delta) \in \mathscr{E}$ that identifies the experiment that the DM should conduct at each $\delta$ in the continuation region ${\cal C}={\cal I}^c$. \vspace{0.2cm} 

The asymptotic analysis of the previous section produces an experimentation strategy $\pi(\delta)=\widetilde{\cal E}^{\Asym}$ and an intervention region ${\cal I}^{\Asym}$ defined in Corollary~\ref{cor:maxsigma} and Proposition~\ref{prop:diffusion_2}, respectively. However, we cannot implement these policies directly since they are computed in terms of the non-primitive quantities ${\cal Q}(x,{\cal E})$,  $\alpha(x,{\cal E},\thetaL)$ and $\alpha(x,{\cal E},\thetaH)$ appearing in Assumption~\ref{assm:asymregime}. Therefore, in order to recover a solution to an arbitrary instance of the problem from the asymptotic analysis of the previous section, we need to derive the values of ${\cal Q}(x,{\cal E})$,  $\alpha(x,{\cal E},\thetaL)$ and $\alpha(x,{\cal E},\thetaH)$ from the primitives of the model, namely, from the values of $\Lambda$, $Q(x,{\cal E},\thetaL)$ and $Q(x,{\cal E},\thetaH)$.  \vspace{0.2cm} 

In some settings this derivation can be done directly by imposing a specific parametric structure in the definitions of $Q(x,{\cal E},\thetaL)$ and $Q(x,{\cal E},\thetaH)$. 
The idea in these settings is that the parametric structure is used to capture a distinctive feature of the problem at hand which in turn would determine the asymptotic regime of interest. Let us illustrate this point with a concrete example.

\begin{exm} Consider a setting where the primitives $Q(x,{\cal E},\theta)$ are known and assumed to be continuously differentiable in $\theta$ for $\theta$ in some open neighborhood that contains the two hypothesis $\Theta=\thetaL$ and $\Theta=\thetaH$. Suppose that we are interested in a setup where a distinctive characteristic of the problem is that the two hypotheses are hard to distinguish. We can model this feature by setting $\theta_1=\theta_0+\xi/\sqrt{k}$ for some fixed scalar $\xi$. 
Using a first order Taylor expansion, it follows that
$Q(x,{\cal E},\thetaH)=Q(x,{\cal E},\thetaL)+Q_\theta(x,{\cal E},\thetaL)\,\xi/\sqrt{k}+o(k^{\-1/2})$, where $Q_\theta(x,{\cal E},\theta)$ is the partial derivative of $Q(x,{\cal E},\theta)$ with respect to $\theta$. Under this specific parametrization of the problem, we can now apply the asymptotic analysis of the previous section (by letting $k$ go to infinity) to derive the corresponding values of ${\cal Q}(x,{\cal E})$,  $\alpha(x,{\cal E},\thetaL)$ and $\alpha(x,{\cal E},\thetaH)$. In this case, it is not hard to see that
$${\cal Q}(x,{\cal E})=Q(x,{\cal E},\thetaL), \qquad \alpha(x,{\cal E},\thetaL)=0 \qquad \mbox{and}\quad \alpha(x,{\cal E},\thetaH)=Q_\theta(x,{\cal E},\thetaL)\,\xi.~~\Diamond $$

\end{exm}
In Section~\ref{sec:Crowdvoting} we consider at length a concrete application related to crowdvoting where $Q(x,{\cal E},\theta)$ are viewed as choice probabilities governed by an MNL model. In this context, similarly to Example~3 we also derive the quantities ${\cal Q}(x,{\cal E})$ and $\alpha(x,{\cal E},\theta)$ not only for the case of indistinguishable  hypotheses but also for the case where the experiments outcomes are very noisy.\vspace{0.2cm}

The previous example provides some insights on how one can leverage some concrete knowledge about the structure of the problem to identify the proper asymptotic regime to use. However, this approach does not generalize in an obvious way to an arbitrary setting for which such knowledge is not available. In what follows we propose a methodology that does not rely on any additional information beyond the values of $\Lambda$, $Q(x,{\cal E},\thetaL)$ and $Q(x,{\cal E},\thetaH)$.\vspace{0.2cm}


Combining the asymptotic scalings in equations \eqref{eq:asymptQ} and \eqref{eq:Lamdaasymptotic}, we have that the input parameters $\Lambda^k$, $Q^k(x,{\cal E},\thetaL)$ and $Q^k(x,{\cal E},\thetaH)$ satisfy the following relationship for $k$ large
\begin{equation}\label{eq:asymrregimecombined}\sqrt{\Lambda^k}\,\left(\frac{Q^k(x,{\cal E},\theta)}{{\cal Q}(x,{\cal E})}-1\right)\approx \alpha (x,{\cal E},\theta).\end{equation}
Furthermore, going back to the conditions that define the asymptotic regime in Corollary~\ref{assm:asymregime}, we see that the value of ${\cal Q}(x,{\cal E})$ is such that the quantity 
$$\frac{Q^k(x,{\cal E},\theta)}{{\cal Q}(x,{\cal E})}-1$$
converges to zero at a rate of $1/\sqrt{k}$ uniformly in ${\cal E} \in \mathscr{E}$, $x \in {\cal E}$ and $\theta=\thetaL,\thetaH$. Thus, in a non-asymptotic regime, we can reinterpret this condition as one that requires ${\cal Q}(x,{\cal E})$ to be as close as $Q(x,{\cal E},\theta)$ as possible for all ${\cal E} \in \mathscr{E}$, $x \in {\cal E}$ and $\theta=\thetaL,\thetaH$.  In other words, we can represent this problem as an optimization problem that minimizes the `distance' between ${\cal Q}(x,{\cal E})$ and $Q(x,{\cal E},\theta)$. We propose the following min-max formulation to compute ${\cal Q}(x,{\cal E})$
\begin{equation}\label{eq:Q(x,E)}
\mbox{For every ${\cal E} \in \mathscr{E}$ solve:}\qquad 
\min_{{\cal Q} \geq 0}\; \max_{\theta \in \{\thetaL,\thetaH\}}\; \max_{x \in {\cal E}}\; \left|\frac{Q(x,{\cal E},\theta)}{{\cal Q}(x,{\cal E})}-1\right|\quad \mbox{subject to}\quad \sum_{x \in {\cal E}} {\cal Q}(x,{\cal E})=1.
\end{equation}

After computing the value of ${\cal Q}(x,{\cal E})$, we can obtain the values of $\alpha(x,{\cal E},\thetaL)$ and $\alpha(x,{\cal E},\thetaH)$ using \eqref{eq:asymrregimecombined}, that is,

\begin{equation}\label{eq:alpha(x,E,theta)}
\alpha(x,{\cal E},\theta)\approx\sqrt{\Lambda}\,\left(\frac{Q(x,{\cal E},\theta)}{{\cal Q}(x,{\cal E})}-1\right), \quad \mbox{for } \theta=\thetaL,\thetaH.
\end{equation}

The following proposition establishes the consistency between the value of the probability kernel ${\cal Q}(x,{\cal E})$ computed in \eqref{eq:Q(x,E)} and the corresponding asymptotic limit.

\begin{prop}\label{prop:consistency} Consider a sequence of probability distributions $\{Q^k(x,{\cal E},\theta);\;{\cal E} \in \mathscr{E}, x \in {\cal E}, \theta=\thetaL,\thetaH\}$ satisfying the condition in Assumption~\ref{assm:asymregime}. In particular, $Q^k(x,{\cal E},\theta)\rightarrow{\cal Q}(x,{\cal E})$ for some probability kernel ${\cal Q}(x,{\cal E})$, for $\theta=\theta_0,\theta_1$.
Moreover, for each $k$ and $Q^k(x,{\cal E},\theta)$, let ${\cal Q}^k(x,{\cal E})$ be the corresponding solution to \eqref{eq:Q(x,E)}.
Then, ${\cal Q}^k(x,{\cal E})$ converges to ${\cal Q}(x,{\cal E})$ as $k \uparrow \infty$ for all ${\cal E} \in \mathscr{E}$ and $x \in {\cal E}$.
\end{prop}

Let us now turn to the issue of how to adapt the asymptotic solutions to derive implementable policies. Equations~\eqref{eq:Q(x,E)} and \eqref{eq:alpha(x,E,theta)} allow us to compute the values of ${\cal Q}(x,{\cal E})$,  $\alpha(x,{\cal E},\thetaL)$ and $\alpha(x,{\cal E},\thetaH)$ that are needed to derive the asymptotic strategy $\pi(\delta)=\widetilde{\cal E}^{\Asym}$ and ${\cal I}^{\Asym}$. From Corollary~\ref{cor:maxsigma}, we have that 
$$\widetilde{\cal E}^{\Asym}= \argmax_{{\cal E} \in \mathscr{E}}\left\{\sum_{x \in {\cal X}_{\cal E}} \big(\alpha(x,{\cal E},\thetaH)-\alpha(x,{\cal E},\thetaL)\big)^2\,{\cal Q}(x,{\cal E})\right\}.$$
On the other hand, the value of ${\cal I}^{\Asym}$ is obtained from our diffusion analysis of the optimal stopping problem combining the results in Proposition~\ref{prop:diffusion_2} and Theorem~\ref{thm:Vsmooth}. The volatility $\tilde{\sigma}$ of the underlying diffusion process is the  one identified in Proposition~\ref{prop:weakconv}, that is,
$$ \tilde{\sigma}^2=\sum_{x \in {\cal X}_{\widetilde{\cal E}^{\Asym}}}  \big(\alpha(x,\widetilde{\cal E}^{\Asym},\thetaH)-\alpha(x,\widetilde{\cal E}^{\Asym},\thetaL)\big)^2\,{\cal Q}(x,{\cal E}).$$

We use this asymptotic solution $(\widetilde{\cal E}^{\Asym},{\cal I}^{\Asym}$) to propose two concrete approximation policies.

\begin{itemize}
    \item {\sf Asymptotic Policy (A):} This policy implements directly the strategy $(\widetilde{\cal E}^{\Asym},{\cal I}^{\Asym}$).

    \item {\sf Maximum Volatility Policy (MV):} This policy uses the same intervention region ${\cal I}^{\Asym}$ as the Asymptotic policy. On the other hand, in terms of experimentation, the MV policy reinterprets the solution in Corollary~\ref{cor:maxsigma} and for each $\delta$ in the continuation region selects the experiment that maximizes the instantaneous volatility, that is, $\pi^{\MV}(\delta)={\cal E}^{\MV}(\delta)$, where
    \begin{equation}\label{eq:maxvolatility}
{\cal E}^{\MV}(\delta) =\argmax_{{\cal E} \in \mathscr{E}} \left\{ \eL\left[{\big(1-{\cal L}({\cal E})\big)^2 \over \delta  +(1-\delta)\,{\cal L}({\cal E})}\right] \right\}.\end{equation}
(The subscript  `MV' is mnemonic for `Maximum Volatility'.)
    
\end{itemize}

Note that the Asymptotic policy suggests a static experimentation while the Maximum volatility offers a dynamic experimentation, function of the current belief. However, it should be clear from our previous discussion that both of these 
policies are asymptotically equal and optimal in the limiting regime defined by equations \eqref{eq:asymptQ} and \eqref{eq:Lamdaasymptotic}.  It is also worth noticing that in contrast to the derivation of an optimal experimentation policy in equation \eqref{eq:Opt_Exp} that requires full knowledge of the value function, the MV policy can be computed directly using only the knowledge of the likelihood function ${\cal L}(x,{\cal E})$. This, of course, simplifies significantly its computational complexity. 
\vspace{0.2cm}

In Section~\ref{sec:Numerics}, we conduct a set of numerical experiments to test the performance of our proposed policies using a concrete application in the context of new product introduction that we present in the next section.
We conclude this section with a remark on how to extend some of the insights that have developed to the problem of designing the type of experiments that the DM can use.

\subsection{A Remark on the Optimal Design of Experiments}\label{subsec:ExpDesign}

In some applications (such as the assortment selection problem that will be discussed in the next section),  the decision maker has some degree of control over the design of the set $\mathscr{E}$ of available experiments. In such cases,  observe that the optimization in \eqref{eq:maxvolatility}  that defines ${\cal E}^{\MV}(\delta)$ can be reformulated over a more  abstract set  $\mathbb{L}$ of likelihood ratios, where each $\cal L\in\mathbb L$ corresponds to an experiment $\cal E$.  As a result, the optimization problem that defines the Maximum Volatility policy  is given by,
$$\max_{{\cal L} \in \mathbb{L}} \ed\left[  \left({1-{\cal L} \over \delta+(1-\delta)\,{\cal L} }\right)^2   \right], \quad \mbox{or equivalently,}\qquad \max_{{\cal L} \in \mathbb{L}} \eL\left[  { (1-{\cal L})^2 \over \delta+(1-\delta)\,{\cal L} }   \right] $$
where, $\eL[\cdot]$ denotes the expectation under the  probability measure $\Pro_0(\cdot)$.\vspace{0.2cm}

Depending on the nature of the set $\mathbb{L}$, the optimization problem above can be casted as a Tchebycheff moment problem. Consider the following setting where the DM can design experiments that correspond to any possible likelihood ratio $\cal L$, as long as $\cal L$ is bounded by two given quantities $\underline{L}$ and $\overline{L}$. In this case the following result holds:

\begin{prop}\label{prop:MaxRange} Suppose that $\mathbb{L}=\big\{{\cal L}\colon \eL[{\cal L}]=1 \;\mbox{and}\; \underline{L} \leq {\cal L} \leq \overline{L}\big\}$ for two non-negative scalars $\underline{L}$ and $\overline{L}$, and let
$${\cal L}^* = \argmax_{{\cal L} \in \mathbb{L}} \eL\left[  { (1-{\cal L})^2 \over \delta+(1-\delta)\,{\cal L} }   \right].$$
Then, ${\cal L}^*$ is a random variable with a two-point distribution with mass at $\underline{L}$ and $\overline{L}$.
\end{prop}
{\sc Proof:} {\sf The result follows from noticing  that the function $(1-\ell)^2/(\delta+(1-\delta)\ell)$ is convex and so an optimal solution is a two-point distribution with mass at $\underline{ L}$ and $\overline{ L}$. $\Box$}\vspace{0.2cm}

 The solution in Proposition~\ref{prop:MaxRange} suggests that the decision maker should select an experimentation policy that maximizes the range of the likelihood function. In Section~\ref{sec:Numerics}, we explore this idea and propose a variation of the Maximum Volatility policy that incorporates this {\rm `maximum range'} condition and show very good numerical performance.

{}



\section{Illustrative Example:  New Product Introduction }\label{sec:Crowdvoting}

We discuss in this section a concrete application of the methodology and results presented in the previous sections in the context of a new product introduction problem. The literature on the topic is quite broad (see, the recent work of \cite{Sunaretal19} and references their). In particular,  we consider an environment in which the experimentation outcomes are the result of a consumers' voting process driven by a Multinomial choice model (MNL). Our objective in developing this example is twofold. First, we use it to provide some specific details on how to formulate and derive our proposed asymptotic approximation policies discussed in the previous section. As a by-product of this discussion, we also show how to obtain diffusion approximations for a belief process that is governed by an MNL model using two different types of asymptotic regimes. Given the popularity of the MNL model to represent consumer preferences, we believe that our  diffusion approximation has applications beyond the one discussed in this section. Our second objective is to use this concrete example in Section~\ref{sec:Numerics} to conduct a set of numerical experiments to test the quality of our proposed methodology. For instance, we are interested in testing the accuracy of the {\em maximum volatility} principle derived in Corollary~\ref{cor:maxsigma} and the two heuristic policies introduced in Section~\ref{sec:heuristic}, which provide remarkably simple rules for conducting dynamic experimentation.  

\subsection{Model Setup}

The specific setting that we consider is as follows. Consider a seller (or firm)  who is contemplating the possibility of introducing a new product (or products) into the \mbox{marketplace}. In the process of developing these new products, the seller has prototyped $n$ different versions and  would like to decide which is the right subset to commercialize, if any.  These prototypes differ in terms of some specific set of attributes which might include their price and quality as well as launching and manufacturing costs, to name a few.  We assume that the intrinsic utility that a consumer assigns to version $i \in [n]$ is equal to $u_i(\Theta)$, where $\Theta>0$ is some unknown real parameter.

\begin{exm}\label{ex:linearutilities}{\rm (Linear Utilities)} A popular modeling approach is to assume that the utilities $u_i(\Theta)$ are linear in the unknown parameter $\Theta$. For instance, we can have $u_i(\Theta)=q_i-p_i\,\Theta$, where $q_i$ and $p_i$ are product $i$'s quality  and price, respectively. In this case $\Theta$ measures consumers' price sensitivity. $\Diamond$
\end{exm}
 \vspace{0.2cm}

The seller is uncertain about market conditions and does not know the value of the  parameter $\Theta$. In an attempt to reduce the risk of launching the wrong version(s), the seller sets up an online voting system in which potential customers (those visiting the seller's website) can vote for the different prototypes.  For simplicity, we assume that each voter votes for at most one version and the seller only tracks the cumulative number of votes for each one. (In practice, we could imagine a more sophisticated interface using a more detailed scoring system, {\em e.g.}, a 0 to 10 scale, or even allowing for consumer reviews.)  This voting phase occurs before the seller decides to launch a product and has the potential of offering a win-win situation whereby a consumer who votes hopes to influence the seller to commercialize the right version; and on the other hand, these votes and their pace provide valuable information that the seller can use to better forecast the value of $\Theta$. As we show later, it is not necessarily optimal for the seller to display the entire set $[n]$ during the voting phase. Hence, we assume that the seller selects a subset ${\cal E} $ of prototypes to show during the voting phase. We call ${\cal E}$ the {\em display} set and let $|{\cal E}|$ be its cardinality. To keep some consistency between the notation in this and the previous sections, we note that, in the most general case,  both the set of  experiments $\mathscr{E}$ and available actions $\mathscr{A}$ coincide with the power set of $[n]$, that is,  $\mathscr{E}=\mathscr{A}=2^{[n]}$.  In some cases, however, one might need to restrict the set of experiments and actions. For instance, if the number of prototypes is large then it might be impractical to display the entire menu and experimentation should be restricted to display sets of a given cardinality. Similarly, it is also possible that the seller is constrained in the number of versions that she can  launch.

\vspace{0.2cm}

Voters arrive according to a Poisson process with rate  $\Lambda$ and vote for one alternative from the display set according to a multinomial choice model.  Specifically, a voter who observes a display set ${\cal E}$ assigns to each version $i \in {\cal E}$ a utility ${\cal U}_i(\Theta)=u_i(\Theta)+\varepsilon_i$, where $\{\varepsilon_i \colon i \in {\cal E}\}$ are idiosyncratic  utility shocks that are independent and identically distributed according to a Gumbel distribution with mean zero and variance $\var[\varepsilon]=\pi^2/(6\,\mu^2)$, for some  fixed constant $\mu>0$. It follows that a utility-maximizing voter votes for version $i \in {\cal E}$ with probability
\begin{equation}\label{eq:MNLChoice}Q(i,{\cal E},\Theta) := \Pro\big({\cal U}_i(\Theta) \geq {\cal U}_j(\Theta), \; \forall j \in {\cal E}\big) = {\exp(\mu\,u_i(\Theta)) \over \sum_{j \in {\cal E}} \exp(\mu\,u_j(\Theta))}.\end{equation}
 Note that our formulation allows for the possibility that a voter might end-up not selecting any of the available options. To model this no-vote option we simply include  version `0' with quality, price and intrinsic utility equal to zero, $u_0(\Theta)=0$. In what follows we assume that every display set ${\cal E}$ includes the non-purchase option.\vspace{0.2cm}

We assume that the seller has a prior belief about  the value of $\Theta$ that can take one of two possible values $\{\thetaL,\thetaH\}$, and her prior is that $\Theta=\thetaL$ with probability $\delta \in (0,1)$. We let $u_i(\thetaL)$ and $u_i(\thetaH)$ denote voters' intrinsic utilities under these two hypotheses for $i \in [n]$ and define the likelihood ratio function  by
\begin{equation}\label{eq:likelihoodratio}{\cal L}(i,{\cal E}) := {Q(i,{\cal E},\thetaH) \over Q(i,{\cal E},\thetaL)}\qquad \forall i \in {\cal E}.\end{equation}

We complete the description of the  model by specifying the seller's objective function. As in the general case, we assume that there exists a piecewise linear function $G(\delta)$ (see equation \eqref{eq:payoff}) that represents the seller's expected payoff  as function of her belief $\delta$. The seller's optimization problem is given by
\begin{equation}\label{eq:probformobfMNL}
\Pi(\delta)=\sup_{(\pi,{\cal I})} \; \e_\delta\left[e^{-r\,\delta_\tau}\, G(\delta_{\tau})\right], \qquad \mbox{subject to}\quad \tau=\inf\big\{t>0 \colon \delta_t \in {\cal I}\big\}.\end{equation}
Recall that a policy is defined by an experimentation policy $\pi$ that determines the collection of display sets $\{{\cal E}_t \in \mathscr{E} \colon 0 \leq t \leq \tau\}$ to use throughout the voting process and an intervention region ${\cal I}$ that defines the duration of the voting campaign.
\vspace{0.2cm}

{\setstretch{1}
\begin{rem}\label{rem:payoffsale}{\rm (Payoffs from Sales)} {\sf To illustrate a concrete example of a piecewise  linear payoff function $G(\delta)$ in the context of new product introduction, consider the case in which the seller is interested in maximizing the expected discounted value of the cash-flows generated by the sales that occur after time $\tau$.  Specifically, at time $\tau$, the seller stops the voting process and selects a subset ${\cal A} \in \mathscr{A}$ of products to launch based on the available information at this time. Suppose consumers arrive according to a Poisson process of rate $\Lambda_s$ and make buying decision according to the same MNL model that governs the voting process. Under this assumption, the seller expected discounted payoff is given by
\begin{align*}
{\cal R}(\delta_\tau,{\cal A})&:=\e\left[\sum_{i \in {\cal A}}  \int_\tau^\infty e^{-r\,(t-\tau)} (p_i-c_i)\, \D S_{it} - K_i \Big|{\cal F}_\tau\right]  = \sum_{i \in {\cal A}} \left[{(p_i-c_i) \over r}\,\Lambda_s\: \e\left[Q(i,{\cal A},\Theta) \Big| {\cal F}_\tau\right]-K_i \right] \\
& = \phi({\cal A})+\beta({\cal A})\,\delta_\tau,
\end{align*}
where $p_i$, $c_i$ and $K_i$ are the per-unit price, manufacturing cost and fixed launching cost of product $i \in \mathscr{S}$, respectively, and
$$\phi({\cal A}):= \sum_{i \in {\cal A}} \left[{(p_i-c_i) \over r}\,\Lambda_s\,Q(i,{\cal A},\theta_1)-K_i\right] \quad \mbox{and}\quad
\beta({\cal A}):= \sum_{i \in {\cal A}} \left[{(p_i-c_i) \over r}\,\Lambda_s\,(Q(i,{\cal A},\theta_0)-Q(i,{\cal A},\theta_1))\right].$$

In the case that all products are discarded, one can assume the seller receives a fixed payoff ${\cal R}_0$ (possibly zero) which captures the opportunity cost of her business.  Finally, the seller's payoff function in this case is given by $G(\delta)=\max \Big\{{\cal R}(\delta,{\cal A})\colon {\cal A} \in \mathscr{A}\Big\}$. $\Diamond$}

\end{rem}}

\subsection{Asymptotic Approximation}\label{subsec:AsympApproxMNL}

We move now to  apply the results in Section~\ref{sec:AsymptoticForm} to approximate the optimization in \eqref{eq:probformobfMNL} by a diffusion control problem. In order to invoke the weak convergence result in Proposition~\ref{prop:weakconv}, we need to specify an asymptotic regime under which the MNL choice probabilities satisfy the condition in equation~\eqref{eq:asymptQ}. In what follows we propose two concrete alternatives, each capturing a different type of uninformativeness associated with the voting process.

\subsubsection{Noisy Preferences}\label{sec:NoisyPreferences} 

Motivated by the issue of low-quality data that has been reported in the context of online learning applications and advertising (\citealp{Kohavi17}, \citealp{LewisRao15}),   we consider a regime in which the variance of the MNL idiosyncratic shocks in the $k^{\mbox{\tiny th}}$ instance of the problem grows proportionally with $k$, namely,  $\var[\varepsilon^k]=k\,\pi^2/(6\,\mu^2)$. In other words, this asymptotic regime is one in which votes --and the information they contain-- become more and more noisy as $k$ grows large.  

Under this scaling, one can show that the choice probability $Q^k(i,{\cal E},\theta)$ in \eqref{eq:MNLChoice}  can be written as
\begin{equation}\label{eq:Muasymptotic} Q^k(i,{\cal E},\theta)={1 \over |{\cal E}|}\,\left[1+{\mu \over |{\cal E}|\,\sqrt{k}}\,\sum_{j \in {\cal E}} (u_i(\theta)-u_j(\theta))+o(k^{-1/2})\right],\end{equation}
which satisfies the requirements in Assumption~\ref{assm:asymregime} with
$${\cal Q}(i,{\cal E})={1 \over |{\cal E}|} \qquad \mbox{and}\qquad \alpha(i,{\cal E},\theta)= \mu\,\big(u_i(\theta)-\bar{u}({\cal E},\theta)\big)\quad \mbox{where}\quad \bar{u}({\cal E},\theta):= {1 \over |{\cal E}|}\,\sum_{j \in {\cal E}} u_j(\theta).$$ 

{}

Recall that under our asymptotic scaling,  voters arrive according to a Poisson process $N^k_t$ with intensity $\Lambda^k=k\,\Lambda$ in the $k^{\mbox{\tiny th}}$ instance of the problem. Given this scaling of $\var[\varepsilon^k]$ and $\Lambda^k$, we can use the result in Proposition~\ref{prop:weakconv} to obtain the following corollary.

\begin{Cor}\label{cor:weakconvMNL} Let $\Delta u_i:=u_i(\thetaH)-u_i(\thetaL)$ and $\Delta \bar{u}({\cal E}):=\bar{u}({\cal E},\thetaH)-\bar{u}({\cal E},\thetaL)$. Suppose the seller uses a static display policy ${\cal E}_t={\cal E}$  during the voting process. Then, the belief process $\delta_t^k$ converges weakly to the solution of the SDE
  $$\D\, \tilde{\delta_t} = \tilde{\sigma}({\cal E})\,\tilde{\delta}_t\,(1-\tilde{\delta}_t)\,\D W_t, \qquad \mbox{where}\quad \tilde{\sigma}^2({\cal E})={\Lambda \,\mu^2\over |{\cal E}|}\,\sum_{i \in {\cal E}} \big(\Delta\,u_i-\Delta\bar{u}({\cal E})\big)^2,$$
and  $W_t$ is a  Wiener process.
\end{Cor}
Combining this result together with the {\em maximum volatility} principle in  Corollary~\ref{cor:maxsigma} we can now identify an optimal display set in this asymptotic regime under consideration, namely

\begin{equation}\label{eq:optdisplayMNLasympt}\widetilde{\cal E}^{\Asym}_{\NP} = \argmax_{{\cal E} \in \mathscr{E}} \left\{ {1 \over |{\cal E}|}\,\sum_{i \in {\cal E}} \big(\Delta\,u_i-\Delta\bar{u}({\cal E})\big)^2\right\}.\end{equation}
(The subscript `NP' stands for Noisy Preferences regime.)\vspace{0.2cm}

Without loss of generality, let us index the prototypes in ascending order of $\Delta u$ so that $\Delta u_1 \leq \Delta u_2 \leq \cdots \leq \Delta u_{n}$. Also, 
for $0 \leq i \leq j \leq n$, let us define the display set
\begin{equation}\label{eq:IntervalSets}{\cal E}[i,j]:=\{0\}\cup\{1,\dots,i\} \cup\{j,\dots,n\},\end{equation}
 which includes the non-purchase option together with the first $i$ prototypes with the lowest values of $\Delta u$ and the $n-j+1$ prototypes with the highest values of $\Delta u$.

\begin{prop}\label{prop:E*MNL} Let $\widetilde{\cal E}^{\Asym}_{\NP}$ be a solution to \eqref{eq:optdisplayMNLasympt}, then there exist integers $n_1$ and $n_2$ with $0 \leq n_1 < n_2 \leq n$ such that $ \widetilde{\cal E}^{\Asym}_{\NP}={\cal E}[n_1,n_2]$. Furthermore, in the special case that all the $\{\Delta u_i\}$ have the same sign  (i.e., $\Delta u_1 \geq 0$ or $\Delta u_{n} \leq 0$) then $\widetilde{\cal E}^{\Asym}_{\NP}$ consists of a single prototype $i^* =\argmax\{|\Delta u_i|\colon i \in \mathscr{S}\}$ together with the non-purchase option `0', that is, $\widetilde{\cal E}^{\Asym}_{\NP}=\{0,i^*\}$.
\end{prop}
An important corollary of Proposition~\ref{prop:E*MNL} is that instead of solving \eqref{eq:optdisplayMNLasympt} over  the power set of $\mathscr{E}$ we can restrict ourselves to the much simpler problem of maximizing the volatility of the belief process over the significantly smaller class of display sets $\big\{{\cal E}[i,j]\colon 0 \leq i \leq j \leq n\big\}$ which has a cardinality of $O(n^2)$.
 \vspace{0.2cm}

\begin{exm} {\rm (Example~\ref{ex:linearutilities} Revisited)} Suppose the intrinsic utility of product $i$ is equal to $u_i(\Theta)=q_i-p_i\,\Theta$, then $\Delta u_i= p_i\,(\thetaL-\thetaH)$. If all the $\{p_i\}$ are of the same sign, for example, if they correspond to the prices of the products, then the $\{\Delta u_i\}$ are also of the same sign and the optimal display set $\widetilde{\cal E}^{\Asym}_{\NP}$ includes a single prototype, namely, the one with the highest price.~$\Diamond$
\end{exm}

\subsubsection{Asymptotically Indistinguishable Hypotheses}\label{sec:IH}

An alternative regime in which we can apply the asymptotic analysis of Section~\ref{sec:AsymptoticForm} corresponds to the case in which the values of $\thetaL$ and $\thetaH$ become indistinguishable as $k$ grows large. To be precise,  let us consider the case in which $u_i(\thetaH)=u_i(\thetaL)+\xi_i/\sqrt{k}$ for $i \in [n]$, where $\{\xi_1, \xi_2,\dots,\xi_n\}$ are fixed constants independent of $k$. Under this scaling, the  choice probability $Q^k(i,{\cal E},\theta)$ in \eqref{eq:MNLChoice}  admit the following representation:
\begin{equation}\label{eq:Qk_IH}Q^k(i,{\cal E},\thetaL)={\nu_i \over \sum_{j \in {\cal E}} \nu_j} \quad \mbox{and}\quad Q^k(i,{\cal E},\thetaH)={\nu_i \over \sum_{j \in {\cal E}} \nu_j}\,\left[1 + {1 \over \sqrt{k}}\, {\sum_{j \in {\cal E}} \nu_j\,(\xi_i-\xi_j) \over \sum_{j \in {\cal E}} \nu_j}+o(k^{-1/2})\right],\end{equation}
where $\nu_i:=\exp(\mu\,u_i(\thetaL))$. It follows that these choice probabilities satisfy the conditions in Assumption~\ref{assm:asymregime} with
$${\cal Q}(i,{\cal E})={\nu_i \over \sum_{j \in {\cal E}} \nu_j}, \qquad \alpha(i,{\cal E},\thetaL)=0 \qquad \mbox{and}\qquad  \alpha(i,{\cal E},\thetaH)= \sum_{j \in {\cal E}} (\xi_i-\xi_j)\,{\cal Q}(j,{\cal E}).$$ 

From Corollary~\ref{cor:maxsigma} the optimal display set in this asymptotic regime is given by

\begin{equation}\label{eq:optdisplayMNLasympt_IH}\widetilde{\cal E}^{\Asym}_{\IH} = \argmax_{{\cal E} \in \mathscr{E}} \left\{ \sum_{i \in {\cal E}} \Big(\alpha(i,{\cal E},\thetaH)\Big)^2\,{\cal Q}(i,{\cal E})
\right\}.\end{equation}
(The subscript `IH' stands for Indistinguishable Hypotheses regime.)\vspace{0.2cm}

To get some intuition about $\widetilde{\cal E}^{\Asym}_{\IH}$, consider an arbitrary display set ${\cal E} \in \mathscr{E}$  and let $\xi({\cal E})$ be a random variables taking values in $\{\xi_1, \xi_2,\dots,\xi_n\}$ with probability distribution  ${\cal Q}(i,{\cal E})$. (In this definition we assume that  ${\cal Q}(i,{\cal E})=0$ if $i \not\in {\cal E}$.) 
Then, \eqref{eq:optdisplayMNLasympt_IH} can be rewritten as
$$ \widetilde{\cal E}^{\Asym}_{\IH} = \argmax_{{\cal E} \in \mathscr{E}} \Big\{ \var[{\xi}({\cal E})] \Big\}.$$

\begin{rem}{\sf  It is worth noticing that the asymptotic regime in which the two alternative hypotheses $\Theta=\thetaL$ and $\Theta=\thetaH$ are asymptotically indistinguishable does not imply that the DM optimization problem becomes trivial in the limit. To see this, let us consider the payoff structure discussed in Remark~\ref{rem:payoffsale}, where ${\cal R}(\delta,{\cal A})$ is the discounted payoff that the DM expects to collect if she launches assortment ${\cal A}$ when her belief is $\delta$. Under the scaling in \eqref{eq:Qk_IH} it is not hard to show that for the $k^{\mbox{\tiny th}}$ instance
$${\cal R}^k(1,{\cal A}) - {\cal R}^k(0,{\cal A})= \sum_{i \in {\cal A}} \left[{(p_i-c_i) \over r}\,{\Lambda_s^k \over \sqrt{k}}\,\alpha(i,{\cal A},\thetaH)\, {\cal Q}(i,{\cal A})\right]+o(k^{-1/2}),$$
where $\Lambda_s^k$ is the selling rate after launching. Thus, depending on the rate of grow of  $\Lambda_s^k$ with $k$ there is a non-negligible difference in payoffs between the two hypotheses and so it is in the DM best interest to try to learn which one holds true.\footnote{During the voting phase we have assumed that the arrival rate of voters $\Lambda^k$ is $O(k)$ but during the selling phase the arrival rate $\Lambda_s^k$ does not need to be of the same order and could drop to $O(\sqrt{k})$. In this case, the different payoffs between the two hypotheses would still be significant.} ~~$\Diamond$}
\end{rem}

\section{Numerical Experiments}\label{sec:Numerics}

In this section, we conduct a set of numerical experiments to assess the quality of our methodology using the application discussed in the previous section. In particular, we are interested in investigating the performance of our proposed Asymptotic and Maximum Volatility policies introduced in Section~\ref{sec:heuristic}.
\vspace{0.5cm}

{\bf Optimality Gap:} In our first set of computational experiments, we numerically evaluate the optimality gap of the Asymptotic and Maximum Volatility policies with respect to an optimal policy using the {\em Noisy Preferences} model in Section~\ref{sec:NoisyPreferences}.  We let $\Pi^{\Asym}(\delta)$, $\Pi^{\MV}(\delta)$ and $\Pi(\delta)$ denote the value functions generated by the A, MV and optimal policy, respectively, and define the optimality gap of these policies by 
 
$$\Delta \Pi^{j}:=\max_{\delta \in (0,1)} \left\{\Pi(\delta)-\Pi^{j}(\delta) \over \Pi(\delta)\right\}, \qquad j=\mbox{A, MV}.$$

We measure  $\Delta \Pi^{A}$ and $\Delta \Pi^{\MV}$ using a set of 500 random instances of the problem. Specifically, we consider a problem with $n=5$ products, whose intrinsic utilities $u_i(\thetaL)$ and $u_i(\thetaH)$ are randomly generated uniformly in $[0,1]$ for all $i \in [n]$. For each random instance we run five different scenarios in which $\Lambda= k$ and $\var[\varepsilon]=k\,\pi^2/(6\,\mu^2)$, with $k=10^\kappa$ for $\kappa=0,1,2,3,4$. The rest of the parameters  are kept fixed with $\mu=1$, $r=0.05$ and the terminal payoff function $G(\delta)=\max\{6-30\,\delta, 4-5\,\delta, 3\,\delta,-20+25\,\delta\} $. This is the same terminal payoff function that we used in the examples in Figures \ref{fig:ExValueItera2} and \ref{fig:ExFourPieces}. Finally, in these and the rest of our numerical computations we evaluate the value function of a given policy using Gauss-Seidel value iteration (see section 6.3 in \citealp{Puterman05}) with an error tolerance of $10^{-3}$ over a mesh of size $10^{-3}$ for the $[0,1]$ interval that defines the domain of $\delta$.\vspace{0.2cm}

Table~\ref{table:optimalitygap} presents the mean optimality gap --as well as the maximum value and standard deviation-- computed over a run of 500 randomly generated instances.
\begin{table}[!h]
\begin{center}
Optimality Gap: $\Delta \Pi^{\Asym}$ \vspace{0.1cm}

\begin{tabular}{c||c|c|c|c|c||} \hhline{~=====}
  &~~~$k=1$~~~  &~~~$k=10$~~~ & ~~~$k=100$~~~ & ~~~$k=1,000$~~~ & ~~~$k=10,000$~~~ \\ \hline \hline
\multicolumn{1}{||c||}{Mean}   & 2.39\% & 1.77\% & 0.87\% & 0.27\% & 0.13\% \\ \hhline{------}
\multicolumn{1}{||c||}{Max}   & 26.19\% & 22.95\% & 13.67\% & 1.99\% & 0.87\% \\ \hhline{------}
\multicolumn{1}{||c||}{St. Dev.}   & 4.88\% & 4.18\% & 1.55\% & 0.39\% & 0.13\% \\ \hhline{======}
\end{tabular}
\end{center}\vspace{-0.3cm}
\begin{center}
Optimality Gap: $\Delta \Pi^{\MV}$ 
\begin{tabular}{c||c|c|c|c|c||} \hhline{~=====}
  &~~~$k=1$~~~  &~~~$k=10$~~~ & ~~~$k=100$~~~ & ~~~$k=1,000$~~~ & $k=10,000$ \\ \hline \hline
\multicolumn{1}{||c||}{Mean}   & 0.56\% & 0.12\% & 0.26\% & 0.15\% & 0.09\% \\ \hhline{------}
\multicolumn{1}{||c||}{Max}   & 4.09\% & 1.47\% & 3.56\% & 1.87\% & 0.43\% \\ \hhline{------}
\multicolumn{1}{||c||}{St. Dev.}   & 0.86\% & 0.22\% & 0.41\% & 0.22\% & 0.06\% \\ \hhline{======}
 \multicolumn{6}{l}{\footnotesize {\sc Data}: $\mu=1$, $r=0.05$,  $G(\delta)=\max\{6-30\,\delta, 4-5\,\delta, 3\,\delta,-20+25\,\delta\} $ and $\Lambda=2k$, $\var[\varepsilon]=k\,\pi^2/(6\,\mu^2)$.}
\end{tabular}
\caption{\footnotesize \sf Optimality gap of the Asymptotic and Maximum Volatility policies.}
\label{table:optimalitygap}
\end{center}
\end{table}
As we can see from the table, the two policies performs very well on average, although, the MV policy is substantially better than the A policy, especially for small value of $k$. As $k$ grow large both policies approach the optimal policy, which is  consistent with our asymptotic analysis in Section~\ref{sec:AsymptoticForm}. By comparing the `Max' rows that report the maximum optimality gap, we can also see that the MV policy is significantly more robust than the A policy for small values of $k$. 
\vspace{0.5cm}

{\bf Running Times:} Another dimension of performance is the computational time required to compute a policy and its corresponding value function. Table~\ref{table:runningtime} shows the average running time (in seconds) of the optimal, Asymptotic and Maximum Volatility policies, as a function of the number of products $n$ available in the menu of prototypes. As we can see, the time required to compute an optimal solution grows exponentially fast with the number of products while the time needed to compute the Asymptotic or Maximum Volatility solution remains low across the range of values of $n$ considered in Table~\ref{table:runningtime}. 

\begin{table}[!h]
\begin{center}
Average Running Time (in seconds) \vspace{0.1cm}

\begin{tabular}{c||c|c|c|c|c||} \hhline{~=====}
  &~~~$n=3$~~~  &~~~$n=6$~~~ & ~~~$n=9$~~~ & ~~~~~$n=12$~~~~~ & ~~~~~$n=15$~~~~~ \\ \hline \hline
\multicolumn{1}{||c||}{Optimal}   & 0.90  & 19.00  & 208.60 & $22.20 \times 10^{2}$  & $24.90\times 10^{3}$ \\ \hhline{------}
\multicolumn{1}{||c||}{A}   & 0.10 & 0.19   & 0.22 & 0.22 & 0.27 \\ \hhline{------}
\multicolumn{1}{||c||}{MV}   & 0.11  &  0.22 &  0.24 & 0.34 &  0.97 \\ \hhline{======}
\end{tabular}\vspace{0.1cm}

{\footnotesize{\sc Data}: $\mu=1$, $r=0.05$,  $G(\delta)=\max\{6-30\,\delta, 4-5\,\delta, 3\,\delta,-20+25\,\delta\} $ and $\Lambda=2$, $\var[\varepsilon]=\pi^2/(6\,\mu^2)$.}
\caption{\footnotesize \sf Running times of the optimal, Asymptotic and Maximum Volatility policies.}
\label{table:runningtime}
\end{center}
\end{table}

The results in Tables~\ref{table:optimalitygap} and \ref{table:runningtime} lead us to conclude that the MV heuristic dominates the A heuristic as it has consistently better optimality gap and comparable running times. For this reason, in the rest of our numerical experiments we will focus  exclusively on further exploring the performance the Maximum Volatility policy.
\vspace{0.5cm}

{\bf Benchmark Analysis:} We next conduct a benchmark analysis in which we compare the performance of the MV policy against the following three alternative policies:

\begin{itemize}
\item {\sf Full Display} (F): This policy always displays the entire set of prototypes, that is,  
\begin{equation}\label{eq:DisplaySetFull}\mbox{\rm Full Display Policy:}\qquad \qquad{\cal E}^{\F} (\delta):= \{0,1,2,\dots,n\}\end{equation}
This is a simple and popular benchmark that does not require any type of optimization.

\item {\sf One-Step-Look-Ahead Policy Approximation} (LA): This is a commonly used value function approximation, which in our setting corresponds to selecting an optimal experiment to display under the assumption that a final decision must be made after the outcome of this experiment is revealed. That is,
\begin{equation}\label{eq:DisplaySetOneStep}
\mbox{\rm One-Step-Look-Ahead Policy:}\qquad {\cal E}^{\mbox{\tiny LA}}(\delta) \in \argmax_{{\cal E} \in \mathscr{E}} \Big\{\e_\delta \Big[G\big(\delta+\eta(\delta,x,{\cal E})\big)\Big]\Big\}.\end{equation}

\item {\sf Maximum Range Policy} (MR): Motivated by the result in Proposition~\ref{prop:E*MNL}, we consider the policy 
\begin{equation}\label{eq:DisplaySetHeuristic}\mbox{\rm Maximum Range Policy:}\qquad \qquad{\cal E}^{\MR} (\delta):= \argmax_{0 \leq i \leq j \leq n}\; \eL\left[{\big(1-{\cal L}({\cal E}[i,j])\big)^2 \over \delta  +(1-\delta)\,{\cal L}({\cal E}[i,j])}\right].\end{equation}
A key advantage of the MR policy over the Maximum Violate (MV) policy is that MR maximizes the instantaneous volatility of the smaller set of experiments ${\cal E}[i,j]$ defined in \eqref{eq:IntervalSets}, which simplifies its computation.

{}

\end{itemize}

We assess the performance of these three policies relative to the Maximum Volatility policy using the following relative error measure:

$$\mbox{Relative Error:}\qquad \bar{\Delta}\Pi^j = \int_0^1 {\Pi^{\MV}(\delta)-\Pi^j(\delta) \over \Pi^{\MV}(\delta)}\, \D \delta, \qquad j=\mbox{F, LA, MR}.$$ 

\begin{figure}[htb]
    \begin{center}
    \includegraphics[width=15cm]{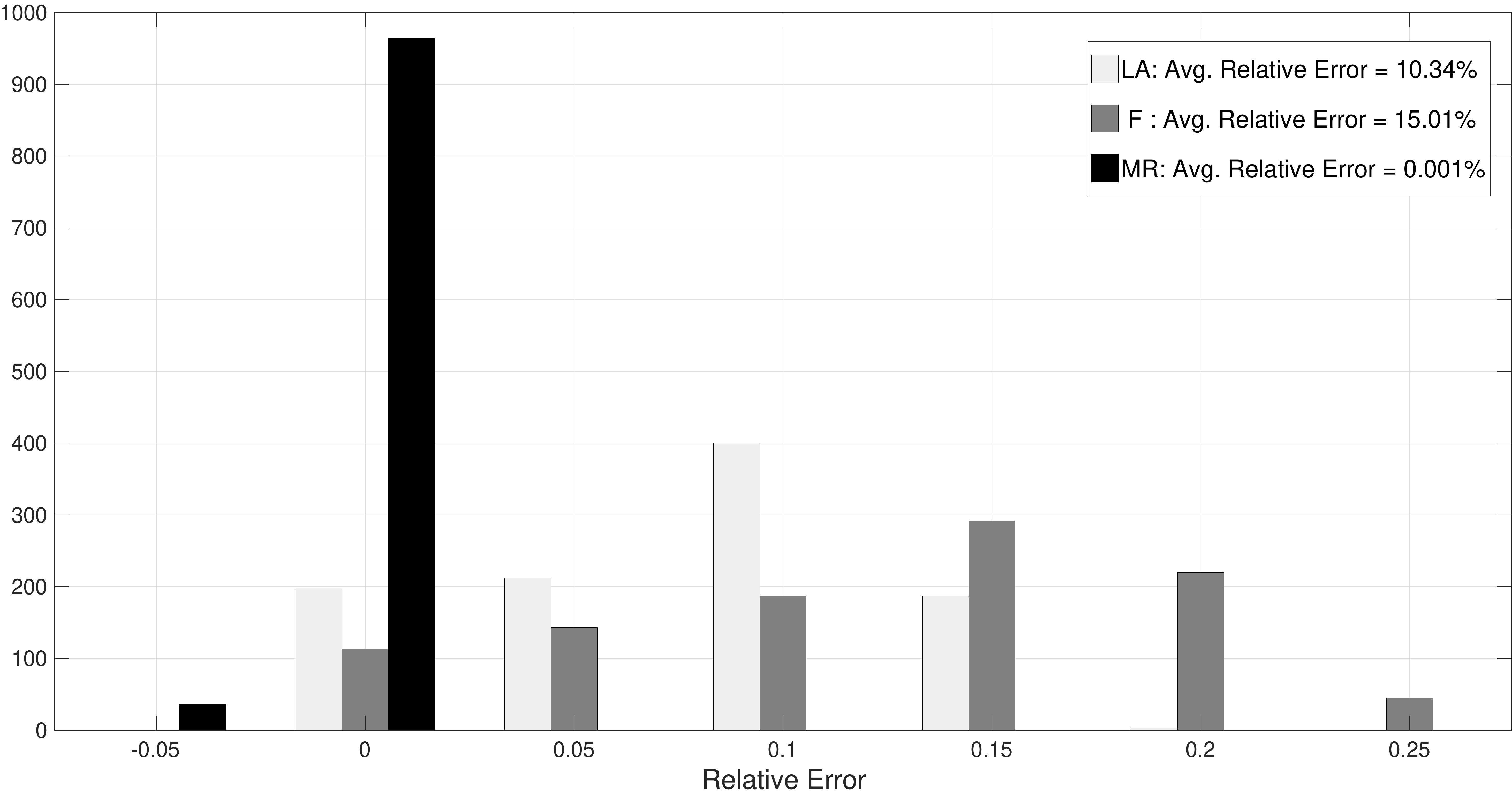}\vspace{-0.5cm}
    \end{center}\caption{\footnotesize \sf Distribution of relative error of the Full (F), One-Step-Look-Ahead (LA) and Maximum Range (MR) policies relative to the Maximum Volatility policy over 1,000 randomly generated instances. \linebreak
    {\sc Data}: $\mu=1$, $r=0.05$,  $G(\delta)=\max\{6-30\,\delta, 4-5\,\delta, 3\,\delta,-20+25\,\delta\} $ and $\Lambda=2$, $\var[\varepsilon]=\pi^2/(6\,\mu^2)$.}      \label{fig:benchmarks}
\end{figure}

Figure~\ref{fig:benchmarks} shows the distribution of this relative error measure for 1000 randomly generated instances of the problem with $n=10$ products each.

As we can see form the figure, the MV policy substantially outperforms the LA and F policies, which have an average relative error of 10.34\% and 15.01\%, respectively. On the other hand, the MR policy is essentially equivalent to the MV policy with an average relative error of 0.001\%. A similar conclusion holds when we compare the average running times of these policies.  Indeed, the average running time per instance are equal to 0.292, 0.224, 0.520 and 16.184 for the MV, MR, F and LA policies, respectively (all times in seconds).

\vspace{0.5cm}
{\bf Value of Optimal Stopping:} We continue our numerical experiments investigating the option value that the DM has by being able to stop the experimentation process at an arbitrary time. Our interest in measuring the value of optimal stopping is driven by the fact that most practical implementations of crowdvoting are executed with a fix, predetermined, time horizon and so we are interested in measuring the opportunity costs of these implementations. 
\vspace{0.2cm}

To this end, let us compare the expected payoffs that the DM collects if she uses the Maximum Volatility experimentation policy ${\cal E}^{\MV}(\delta)$ in equation \eqref{eq:maxvolatility} with and without optimal stopping. For the case with optimal stopping, this expected payoff is $\Pi^{\MV}$ as defined above.
For the case without optimal stopping, we assume that the DM has a fixed predetermined ``budget of experimentation''  of $T$ votes that she can collect. In practice, this budget might reflect external constraints on the amount of time or monetary resources available to experiment. Within this budget of experimentation we assume the DM implements the maximum volatility policy ${\cal E}^{\MV}(\delta)$. We denote by $\Pi^{\MV}_{\mbox{\tiny T}}$ the corresponding payoff.  We define the value of optimal stopping by 
$$\mbox{Value of Optimal Stopping:}\quad \max_{\delta \in [0,1]}\left\{ {\Pi^{\MV}-\Pi^{\MV}_{\mbox{\tiny T}} \over \Pi^{\MV}}\right\}.$$

\begin{figure}[!h]
    \begin{center}
    \includegraphics[width=15cm]{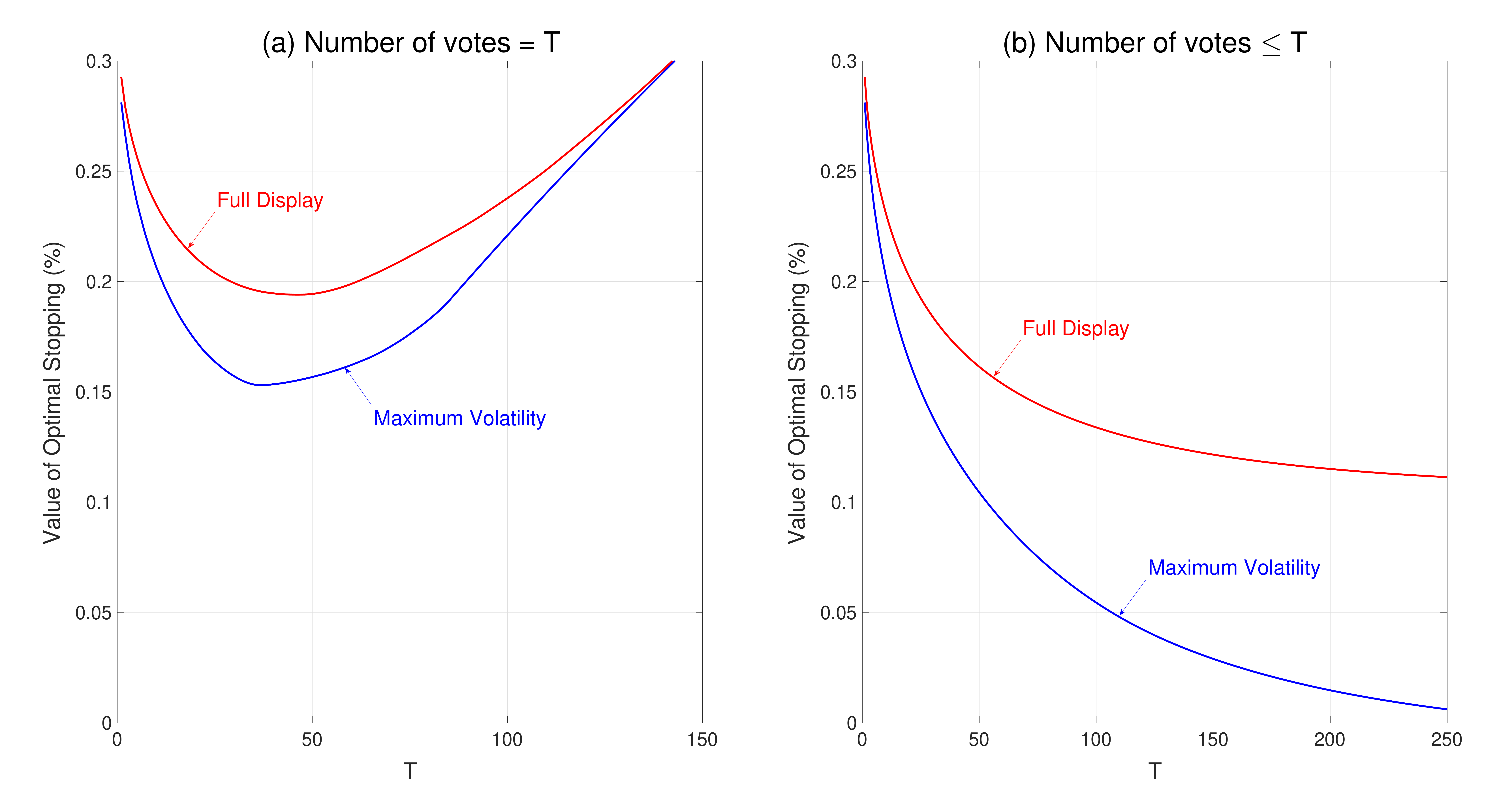}\vspace{-0.5cm}
    \end{center}\caption{\footnotesize \sf \mbox{Value of Optimal Stopping as a function of the number of votes $T$ for the maximum volatility and full display strategies.} \linebreak
    {\sc Data}: $\mu=1$, $r=0.05$,  $G(\delta)=\max\{6-30\,\delta, 4-5\,\delta, 3\,\delta,-20+25\,\delta\} $,  $\Lambda=2$, $\var[\varepsilon]=\pi^2/(6\,\mu^2)$.}      \label{fig:FiniteTime}
\end{figure}

Figure~\ref{fig:FiniteTime} illustrates the average value of optimal stopping in a concrete instance of the problem with $n=5$ products for 100 randomly generated instances in which consumers' utilities  $u_0(i)$ and $u_1(i)$ are uniformly distributed in [0,1] for $i\in [n]$. Panel (a) depicts the average value of optimal stopping when experimentation is constrained to last exactly $T$ rounds. On the other hand, panel (b) depicts the average value of optimal stopping when the experimentation is constrained to be at most $T$ rounds, that is, in this case the DM is able to stop experimenting before collecting $T$ votes. Also, for comparison purposes, Figure~\ref{fig:FiniteTime}  includes the average value of optimal stopping when the full display rule ${\cal E}^{\F}$ in \eqref{eq:DisplaySetFull} is used.\vspace{0.2cm}

As we can see from the figure, the value of optimal stopping can be quite significant depending on the value of $T$. This is specially clear on panel (a) in which the value of optimal stopping can be as large as 20\% or more   if the number of votes $T$ is too small or too large. Intuitively, when $T$ is too small the DM is not able to collect enough information and ends up making  wrong decisions. On the other hand, when $T$ is too large and the DM exhausts the experimentation budget, then she is guaranteed to collect a large amount of information but pays the price of delaying a final decision too much, which again has a negative effect on payoffs because of discounting, i.e., the DM collects more information that needed.  For panel (b), as expected, the value of optimal stopping decreases monotonically with $T$ as the DM in this case is not forced to exhaust all her experimentation budget.\vspace{0.2cm}

In this example, the average value of optimal stopping under a maximum volatility experimentation rule is minimized around $T=40$ when the DM operates under the constraint of collecting exactly $T$ votes (panel a) and is about 15\%. In contrast, if a Full display policy is used  under the same constraint the value of optimal stopping is significantly higher, achieving a minimum around 18\% when $T=45$ votes. By comparing panels (a) and (b) we can appreciate the option value of optimal stopping; not only the value of optimal stopping is monotonically decreasing in $T$ on panel (b) but is also significantly smaller compared to panel (a).  These results underscore the significance of giving the DM the option to stop at any time as well as the benefits --from a learning perspective-- of using maximum volatility ${\cal E}^{\MV}(\delta)$ instead of the popular full display strategy.

\vspace{0.5cm}

{\bf Comparison to MNL Bandit Algorithms:} We conclude our numerical experiment by comparing our proposed Maximum Volatility policy to a couple of policies from the growing literature on multi-armed bandit problems. We specifically selected one MNL-bandit algorithm and one best arm identification algorithm. The reason for selecting algorithms from this part of the broad literature on sequential testing is  that we can cast our assortment selection model in Section~\ref{sec:Crowdvoting} as a multi-armed bandit, where each assortment can be viewed as an arm and where at each arrival the DM has to pull one of them to experiment with. Moreover, the growing literature on bandit problems and specifically MNL-bandit setups have considered assortment planning as one of their primary and most natural application (see, e.g. \cite{CaroGallien07} and \cite{Agrawal2017}).  
Many of the algorithms in this literature have been developed with the objective of minimizing the DM regret over a finite time horizon. This setting is different than ours in the sense that our objective is to identify, as quick as possible, the best possible ``arm'' (action) to choose. However, we can still adapt our proposed Maximum Volatility methodology to this minimum regret setting. This shouldn't be of great concern given that our approach is obtained for a general reward function and as discussed our experimentation policy is independent of the duration.\vspace{0.2cm}

To this end, we assume that the DM has a non-informative uniform prior (i.e., $\delta=0.5$) and uses the MV policy for a fixed number of votes $T$. Using a slight abuse of notation, let us denote by $\delta_{\mbox{\tiny $T$}}^{\MV}$ the DM's posterior belief after this voting period has ended and let $a^{\MV}_{\mbox{\tiny $T$}} \in {\cal A}^*(\delta_{\mbox{\tiny $T$}}^{\MV})$  be the optimal action she chooses. For simplicity, we will consider the case in which the optimal action sets ${\cal A}^*(\delta)$ are restricted to include a single product, in other words, the DM wants to launch a single product into the marketplace. We let ${\cal R}(a^{\MV}_{\mbox{\tiny $T$}} ,\Theta)$ be the reward associated with this policy as a function of $\Theta$. On the other hand, we define ${\cal R}^*(\Theta)=\max_{a \in \mathscr{A}}\{ {\cal R}(a,\Theta)\} $ to be the optimal reward of a clairvoyant who knows the true value of $\Theta$. The terminal regret under this modified MV policy is given by 
$\Delta {\cal R}^{\MV}:={\cal R}^*(\Theta)-{\cal R}(a^{\MV}_{\mbox{\tiny $T$}} ,\Theta)$. \vspace{0.2cm}

The following are the two alternative algorithms that we use for comparison:

\begin{itemize}
\item {\sc MNL-Bandit.} The first algorithm that we consider is the one proposed by \cite{Agrawal2017} (Algorithm 1). This is a `general purpose' algorithm that makes no prior assumption on the MNL model, except for requiring that the no purchase option is the most frequent choice. The algorithm is also designed with the objective of minimizing the rate at which  cumulative regret grows as a function of the number of votes $T$ rather than the terminal regret at $T$, so the comparison is not ideal. The MNL-Bandit is a UCB-type algorithm that periodically during the voting process estimates upper bounds on the attraction scores $v_i=\exp(\mu\,u_i)$ of each product $i \in [n]$ and uses these upper bounds to display the assortment that maximizes rewards.  In the implementation of the MNL-Bandit algorithm, we initialize the value of the attraction scores to one. We let $v_{i\mbox{\tiny $T$}}$ denote the terminal estimate of the attraction score for product $i$ after $T$ votes. Using these terminal scores, we define $a^{\mbox{\tiny MNL-B}}_{\mbox{\tiny $T$}}$ to be the action (product) that maximizes the DM expected reward. The terminal regret of thr MNL-Bandit algorithm is given by $\Delta {\cal R}^{\mbox{\tiny MNL-B}}:={\cal R}^*(\Theta)-{\cal R}(a^{\mbox{\tiny MNL-B}}_{\mbox{\tiny $T$}} ,\Theta).$

\item {\sc Top-Two Probability Sampling (TTPS).} This algorithm is a variation of a recently proposed algorithm by \cite{russo2016simple} for best arm identification. Like  MV,  TTPS is a  Bayesian algorithm that updates the belief $\delta$ after each vote. The key difference is in the experiment that is used at every voting epoch. For each value of $\delta$, TTPS identifies the best and second best experiments, in terms of the reward they generate, and selects one of them at random with probabilities $\beta$ and $1-\beta$, respectively, where $\beta$ is a tuning parameter. In our simulations we use $\beta=0.5$, which is the default value used by \cite{russo2016simple}.  We let $\delta_{\mbox{\tiny $T$}}^{\mbox{\tiny TTPS}}$ denote the posterior belief produced by the TTPS algorithm after $T$ votes and let $a^{\mbox{\tiny TTPS}}_{\mbox{\tiny $T$}} \in {\cal A}^*(\delta_{\mbox{\tiny $T$}}^{\mbox{\tiny TTPS}})$ be the corresponding optimal action. The terminal regret of TTPS is equal to
$\Delta {\cal R}^{\mbox{\tiny TTPS}}:={\cal R}^*(\Theta)-{\cal R}(a^{\mbox{\tiny TTPS}}_{\mbox{\tiny $T$}} ,\Theta).$

\end{itemize}

\begin{figure}[!h]
    \begin{center}
    \includegraphics[width=16cm]{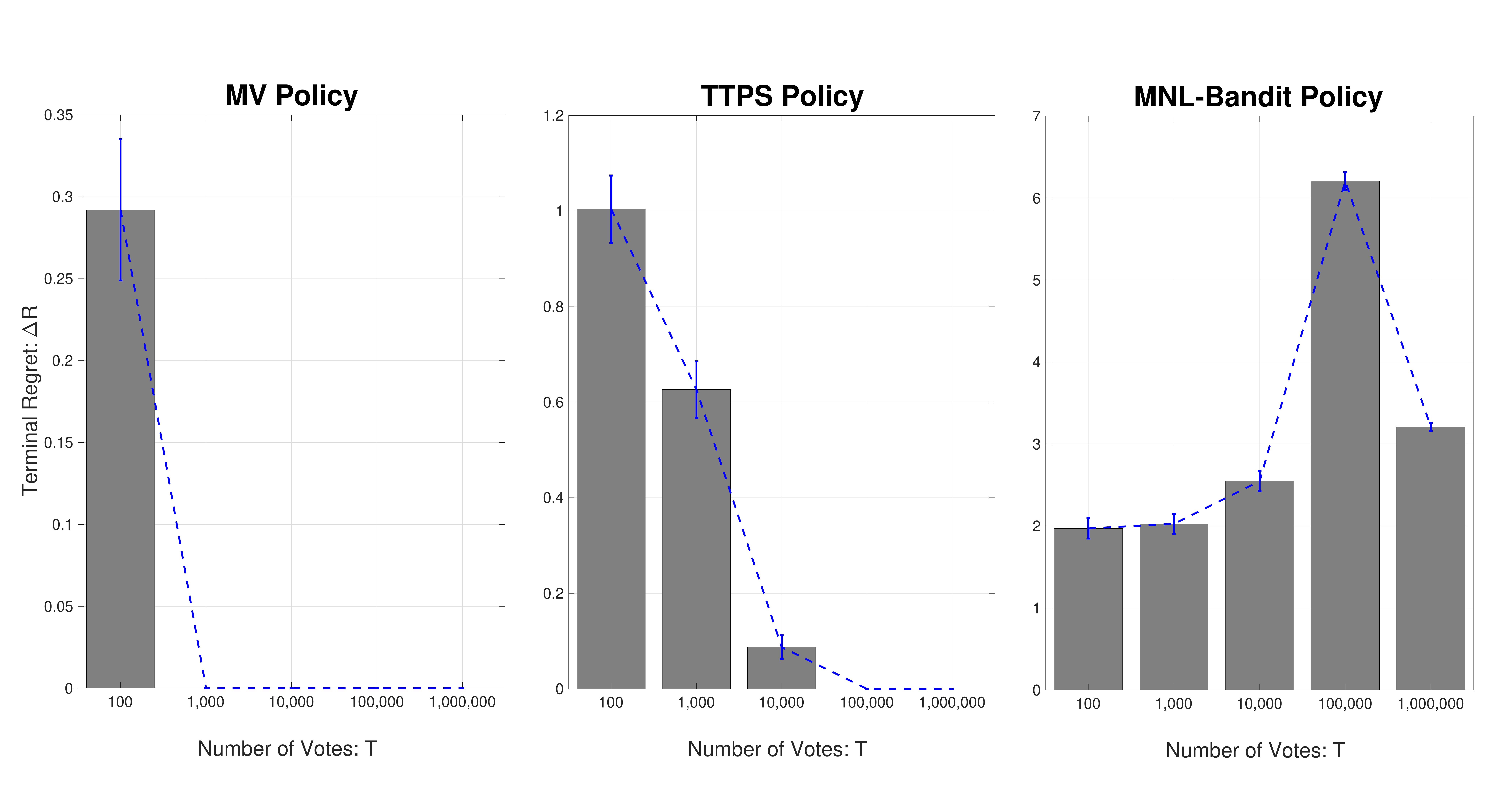}\vspace{-1cm}
    
    \end{center}\caption{\footnotesize \sf Average terminal regret for the MV, TTPS and MNL-Bandit algorithms as a function of the number of votes $T$. For each value of $T$, the average is calculated over 1,000 simulations. The error bars indicate the 95\% confidence interval for the mean.}    \label{fig:Bandits}
\end{figure}

In our numerical experiments we use simulation to evaluate the values of $\Delta {\cal R}^{\MV}, \Delta {\cal R}^{\mbox{\tiny MNL-B}}$ and $\Delta {\cal R}^{\mbox{\tiny TTPS}}$. Figure~\ref{fig:Bandits} depicts the average terminal regret of these three policies for values of $T$ ranging from 100 to one million votes for a specific instance with $n=5$ products. The attraction scores $v_i(\theta)=\exp(\mu\,u_i(\theta))$ and per unit margin $p_i-c_i$ for each of the five products is reported in Table~\ref{table:datainstance}.

\begin{table}[h]
\begin{center}
\begin{tabular}{||l||c|c|c|c|c||}  \hline
Product & 1 & 2 & 3 & 4 & 5  \\ \hline \hline
$v_i(\thetaL)$ & 0.05 & 0.08 & 0.012 & 0.05 & 0.04	  \\  \hline
$v_i(\thetaH)$ & 0.032 & 0.07 &	0.018 &	0.12 &	0.043  \\  \hline
$p_i-c_i$ & 210 & 121.5 & 506 & 42 & 208 \\  \hline
\end{tabular}
\caption{\footnotesize \sf  Vectors of attraction scores and margins for the instance used in the computational experiments reported in Figure~\ref{fig:Bandits}.}   \label{table:datainstance}
\end{center}
\end{table}

For each algorithm and value of $T$, we run 1,000 simulations to compute the average terminal regret. Figure~\ref{fig:Bandits} also depicts the 95\% confidence interval of the mean terminal regret (error bars). 

As we can see form the figure, the MV Policy outperforms the other two in terms of achieving a lower terminal regret with significantly fewer votes. Indeed, for $T \geq 1,000$, the MV policy has essentially zero terminal regret. On the other hand, the TTPS policy needs $T \geq 100,000$ to achieve a zero terminal regret. Finally, the MNL-Bandit algorithm does not produce a zero terminal regret for any value of $T$.\vspace{0.2cm} 

We note that we need to read the results in Figure~\ref{fig:Bandits} with cautious. The fact that the MNL-Bandit algorithm does not perform well in terms of minimizing terminal regret should not be surprising as this policy is not designed for this purpose but rather to minimize cumulative regret. To provide a complete picture of the performance of these policies, we have also run a set of experiments to measure their cumulative regret as a function of $T$.

\begin{figure}[!h]
    \begin{center}
    \includegraphics[width=16cm]{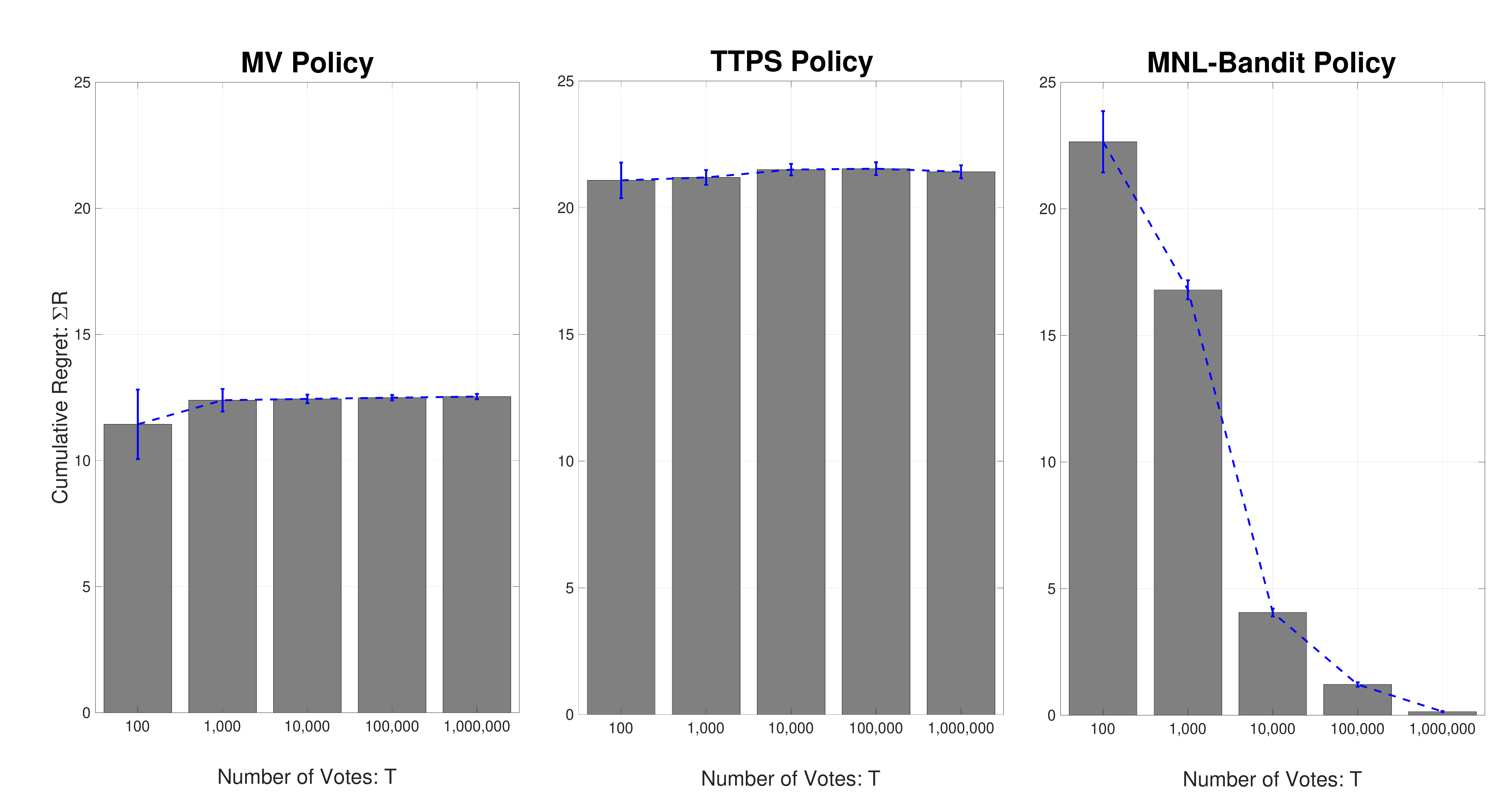}\vspace{-1cm}
    
    \end{center}\caption{\footnotesize \sf Average cumulative regret for the MV, TTPS and MNL-Bandit algorithms as a function of the number of votes $T$. For each value of $T$, the average is calculated over 1,000 simulations. The error bars indicate the 95\% confidence interval for the mean.}    \label{fig:Bandits_Cumulative}
\end{figure}

Figure~\ref{fig:Bandits_Cumulative} depicts the average (per vote) cumulative regret of the three policies. As we can see, only the MNL Bandit algorithm achieves a sublinear regret in $T$ while both MV and TTPS have linear cumulative regret. Again,  this should be expected since MV and TTPS  are pure learning policies designed to identify as quickly as possible the best assortment.

{}

\section{Conclusion}\label{sec:Conclusion}
We considered in this paper a DM that must select an action to maximize a reward function. This function is parameterized by an unknown quantity that the DM learns by experimenting. The DM has to decide dynamically which experiment to conduct and when to stop the experimentation in order to generate the discounted reward. We formulate this problem as a sequential Bayesian testing problem with dynamic experimentation. We adopt a novel diffusion-asymptotic analysis technique that relies on scaling two parameters of the problem. We do that by speeding up the \textit{frequency} at which experiments are conducted while simultaneously reducing the \textit{informativeness} of the outcome of each individual experiment. As a result, we derive a diffusion approximation for the underlying sequential testing problem. The benefit of such scaling is that it allows the limiting regime to remain comparable to the initial setting in terms of the informativeness of the experimentation process per unit of time. Therefore, one expects the corresponding asymptotic approximation to be more accurate than ones from other regimes. Interestingly, this high frequency vs. low informativeness regime has also its own merits and depicts many practical situations. It is specifically in line with online experimentation where the velocity of data (high frequency) is always contrasted with its veracity and lack of accuracy (low informativeness).  The diffusion model we obtain provides a number of important insights with respect to the nature of the problem and its solution. 
In particular, it shows that an optimal experimentation policy is one that chooses the experiment that maximizes the instantaneous volatility of the belief process. This {\em maximum volatility principle} reduces dramatically the complexity of the problem and its solution. 
On the implementation side, we suggest a universal approach to interpret the diffusion approximation solution and ``unscale" it in order to derive a heuristic for the original problem i.e., in the non-asymptotic regime.  
\vspace{0.2cm}

The model that we have studied in this paper is very general and can accommodate a number of applications. However, to test our solution, we consider in Section~\ref{sec:Crowdvoting} a specific setting on assortment selection in the context of new product introduction where the experimentation follows a crowdvoting setup. As a by-product of this analysis, we derive a diffusion limit for the belief process updated following consumers votes, themselves governed by an MNL model.  Given the popularity of the MNL model to represent consumer preferences, we believe that our  diffusion analysis and approximation have applications beyond the one discussed in this paper. 
\vspace{0.2cm}

Our work opens also some interesting and natural research avenues. 
The diffusion approximation obtained by counterbalancing large sample sizes with little informativeness has revealed to be extremely effective and suggest that such approach should be considered in other related settings where it might offer new approximations that complement those obtained by scaling only one parameter, (e.g., the sample size). The recent work of \cite{WagerXu21} on MAB and \cite{ZeniosWang21} in the context of clinical trials, represent another confirmation of this claim.  \vspace{0.2cm}

 One simplifying assumption that resonated well with the online experimentation motivation of this work, is the cost-free experimentation. However, there are other applications (such as clinical trials) where experimentation can be expensive and cannot be discarded. Adding an experimentation cost will affect the principle that governs the selection of the experiment and certainly induce an earlier stopping.  Another ingredient of our model that made some of our analysis more tractable is the discrete set of experiments available to the DM from the start. In Section~\ref{subsec:ExpDesign} we briefly discussed the design of the experimentation set in a way that leverages our model setup and analysis. We believe this is an interesting avenue to explore further by adopting probably a continuous and infinite set of possible experiments from which the DM ``designs" dynamically the ones that are more effective for learning. Finally, our assumption that the unknown parameter takes only two values is restrictive yet, this assumption is an important first step in unravelling the multiple layers the problem and the approach followed have to offer. The multi-hypothesis case requires a much more complex analysis and is left for a future work. We believe for instance that the principle of maximum volatility will be preserved, yet would require an adaptation of the definition of volatility to accommodate the multi-dimensional processes involved. As a result, the optimal experimentation would become state dependent and the diffusion limit of the belief processes would require even more advanced machinery than the one used in the two hypothesis case. We also conjecture that the optimal stopping problem would again decouple from the dynamic experimentation. However, finding the stopping regions in this multi-dimensional setting will not be easy to characterize (see \cite{Dayaniketal21}, in the case of a compound Poisson process).

{}

\vspace{1cm}
\noindent {\bf Acknowledgement:} The authors are very grateful to the Department Editor, Omar Besbes, for encouraging them to further expand the scope of the proposed asymptotic regime in which the information content of experiments is very low.   
The authors are also very grateful to the Associate Editor and three referees for their careful reading of the paper and for the many helpful and constructive comments. The second author thank the University of Chicago Booth School of Business for financial support.

\bibliographystyle{ormsv080}
    \bibliography{BiblioPricing}
\newpage


\appendix

\renewcommand{\theequation}{\thesection-\arabic{equation}}
\setcounter{equation}{0}

\section{Appendix: Proofs}

 {\sc Proof of Lemma~\ref{lem:dynamic-q_t}:} Let $t_i$ be the $i^{\mbox{\tiny th}}$ jump of $N_t$ and let $\delta_{t_i-}$ be the decision maker's belief just before observing the outcome $x_{t_i}$ of experiment ${\cal E}_{t_i}$. Then, by Bayes's rule we have that
\begin{align*}\delta_{t_i}=\Pro(\Theta=\thetaL|{\cal F}_{t_i-},x_{t_i})&= {\Pro(x_{t_i-}|{\cal F}_{t_i-},\Theta=\thetaL)\,\Pro( \Theta=\thetaL|{\cal F}_{t_i-}) \over \Pro(x_{t_i}|{\cal F}_{t_i-})}={Q(x_{t_i},{\cal E}_{t_i}, \thetaL)\,\delta_{t_i-} \over Q(x_{t_i},{\cal E}_{t_i}, \thetaL)\,\delta_{t_i-}  + Q(x_{t_i},{\cal E}_{t_i}, \thetaH)\,(1-\delta_{t_i-})}\\
&= {\delta_{t_i-} \over \delta_{t_i-}  + (1-\delta_{t_i-})\,{\cal L}(x_{t_i},{\cal E}_{t_i})}.\end{align*}
By iterating this recursion, with $\delta_{t_0}=\delta$, we get that

$$\delta_{t_i} ={\delta \over \delta +(1-\delta)\,L_{t_i}},\qquad \mbox{where } L_{t_i} =\prod_{j=1}^i {\cal L}(x_{t_j},{\cal E}_{t_j}).$$
Finally, the result follows from noticing that $\delta_t$ is a pure jump process and so $\delta_t=\delta_{t_{N_t}}$. $\Box$

 \vspace{0.5cm}

 {\sc Proof of Proposition~\ref{prop:valuefunctionconvex}:}  First, the convexity of $G(\delta)$ follows
 directly from its representation in \eqref{eq:payoff} and the fact that the `max' of convex functions is
 also a convex function.  To prove the convexity of the value function $\Pi(\delta)$ for $\delta \in (0,1)$, let us first recall that the value function $\Pi(\delta)$ satisfies the HJB equation:

\begin{equation}\label{eq:HJBpure_app}\Pi(\delta)=\max\left\{G(\delta)\;,\;
{\Lambda \over \Lambda+r}\,\max_{{\cal E} \in \mathscr{E}} \Big\{\e_\delta\Big[\Pi\big(\delta+\eta(\delta,x,{\cal E})\big)\Big]\Big\}\right\}.\end{equation}

Now consider a sequence of functions $\{\Pi_k(\delta)\}_{k \geq 0}$ defined recursively by
$$\Pi_{k+1}(\delta)=\max\left\{G(\delta)\;,\;
{\Lambda \over \Lambda+r}\,\max_{{\cal E} \in \mathscr{E}} \Big\{\e_\delta\Big[\Pi_k\big(\delta+\eta(\delta,x,{\cal E})\big)\Big]\Big\}\right\}, \qquad k=0,1,\dots$$
with $\Pi_0(\delta)=G(\delta)$. It is easy to see that the functions $\{\Pi_k(\delta)\}_{k \geq 0}$ are continuous in $\delta$ and pointwise monotonically increasing in $k$, that is, $\Pi_{k+1}(\delta) \geq \Pi_k(\delta)$ for all $\delta \in (0,1)$. Furthermore, the sequence converges uniformly to a limit $\Pi(\delta):=\lim_{k \to \infty} \Pi_k(\delta)$ that satisfies the HJB equation in \eqref{eq:HJBpure_app}. To see this, note that
\begin{align*}
0 \leq \Pi_{k+1}(\delta) -\Pi_k(\delta) & \leq {\Lambda \over \Lambda+r}\,\left[ \max_{{\cal E} \in \mathscr{E}} \Big\{\e_\delta\Big[\Pi_k\big(\delta+\eta(\delta,x,{\cal E})\big)\Big]\Big\} -\max_{{\cal E} \in \mathscr{E}} \Big\{\e_\delta\Big[\Pi_{k-1}\big(\delta+\eta(\delta,x,{\cal E})\big)\Big]\Big\}\right] \\ &\leq
{\Lambda \over \Lambda+r}\,\left[ \e_\delta\Big[\Pi_k\big(\delta+\eta(\delta,x,{\cal E}^*_k(\delta))\big)\Big] -\e_\delta\Big[\Pi_{k-1}\big(\delta+\eta(\delta,x,{\cal E}^*_k(\delta))\big)\Big]\right]\\
&\leq {\Lambda \over \Lambda+r}\, \e_\delta\Big[  \Pi_k\big(\delta+\eta(\delta,x,{\cal E}^*_k(\delta))\big)-\Pi_{k-1}\big(\delta+\eta(\delta,x,{\cal E}^*_k(\delta))\big)\Big],
\end{align*}
where
$${\cal E}^*_k(\delta):=\argmax_{{\cal E} \in \mathscr{E}} \Big\{\e_\delta\Big[\Pi_k\big(\delta+\eta(\delta,x,{\cal E})\big)\Big]\Big\}.$$
Taking the `sup' over $\delta$, it follows that
$$\rho_k \leq {\Lambda \over \Lambda+r}\,\rho_{k-1} \leq \left( {\Lambda \over \Lambda+r}\right)^k\, \sup_{\delta \in (0,1)} \Big\{G(\delta)\Big\}, \qquad \mbox{where } \rho_k:=\sup_{\delta \in (0,1)} \Big\{\Pi_{k+1}(\delta) -\Pi_k(\delta)\Big\}$$ and so $\rho_k \to 0$ as $k \to \infty$.

We now complete the proof by showing that the  HJB operator preserves convexity. That is, if $\Pi_k(\delta)$ is convex in $(0,1)$ then $\Pi_{k+1}(\delta)$ is also convex. First, since $G(\delta)$ is convex, it follows trivially that $\Pi_1(\delta)$ is convex. Now, let us suppose that $\Pi_k(\delta)$ is convex in $(0,1)$. Then, since the `max' operator preserves convexity, we just need to show that $\e_\delta\big[\Pi_k\big(\delta+\eta(\delta,x,{\cal E})\big)\big]$ is convex. We can rewrite this expectation as follows:
$$\e_\delta\Big[\Pi_k\big(\delta+\eta(\delta,x,{\cal E})\big)\Big]=\sum_{x \in {\cal X}_{\cal E}} \Pi_k\left(\delta\,Q(x,{\cal E},\thetaL) \over \delta\,Q(x,{\cal E},\thetaL)+(1-\delta)\,Q(x,{\cal E},\thetaH)\right)\, \big(\delta\,Q(x,{\cal E},\thetaL)+(1-\delta)\,Q(x,{\cal E},\thetaH)\big).$$
Since the sum of convex functions is convex, we will show that each summand on the right-hand side above is convex. To ease notation, let us define $Q_0=Q(x,{\cal E},\thetaL)$, $Q_1=Q(x,{\cal E},\thetaH)$,  $y=\delta\,Q_0+(1-\delta)\,Q_1$ and define the function
$$H(y):=y\,\Pi_k\left(ay+b \over y\right), \qquad \mbox{where }a:={Q_0 \over Q_0-Q_1}\quad \mbox{and } \; b:={Q_0\,Q_1 \over Q_1-Q_0}.$$
Since $y$ is a linear transformation of $\delta$ we can focus on proving the convexity of $H(y)$ for $y \in (Q_0,Q_1)$ (assuming, without loss of generality, that $Q_0 <Q_1$).  To this end,  we will use the following characterization of a  convex function:
\begin{quotation}
\noindent {\sf Let a continuous function $h(\delta)$ be such that for any $\delta$ in the interior of the domain of $h$   the subdifferential $\partial h(\delta)$ is not empty. Then $h$ is convex.} (see Theorem 3.2.6 in Bazaraa et al. 1993)
\end{quotation}
Thus, we would like to show that for every $y \in (Q_0,Q_1)$, there exists a subdifferential $\partial H_y$ such that
$H(z) \geq H(y)+\partial H_y\,(z-y)$, for all $z \in (Q_0,Q_1)$.
Since $\Pi_k(\delta)$ is convex then for every $\delta \in (0,1)$ there exists a subdifferential $\partial \Pi_\delta$ such that
$\Pi_k(z) \geq \Pi_k(\delta)+\partial \Pi_\delta\,(z-\delta)$, for all $z \in (0,1)$. For $y \in (Q_0,Q_1)$ define
$$\hat{y}:={a\,y+b \over y}$$
then by the convexity of $\Pi_k(\delta)$ for any $z \in (Q_0,Q_1)$ we have
$$\Pi_k\left({a\,z+b \over z} \right) \geq \Pi_k\left({a\,y+b \over y} \right)+ \partial \Pi_{\hat{y}}\,\left( {a\,z+b \over z} - {a\,y+b \over y}\right).$$Multiplying by $z\,y$ (which is nonnegative since $Q_0 >0$) and rearranging terms we get that
$$y\,H(z) \geq z \,H(y)-b\,\partial\Pi_{\hat{y}}\,(z-y) \quad \Longleftrightarrow\quad H(z) \geq H(y)+\left(H(y)-b\,\partial\Pi_{\hat{y}} \over y\right)\,(z-y)$$
and so
$$\partial H_y:=\left(H(y)-b\,\partial\Pi_{\hat{y}} \over y\right)$$
is a subdifferential for $H$ at $y$. This completes the proof. $\Box$

 {}

\vspace{0.5cm}

{\sc Proof of Lemma~\ref{lem:SDEdelta}}: Recall that the belief process can be written in terms of the likelihood function $L$ as follows:
$$\delta_t ={\delta \over \delta+(1-\delta)\, L_t}, \qquad \mbox{where}\quad L_t := \prod_{i=0}^{N_t} {\cal L}(x_{t_i},{\cal E}_{t_i}).$$

Now if we consider the log-likelihood function we can rewrite $\delta_t$ as follows:
$$\delta_t = f(Y_t)\quad\mbox{where }\quad f(Y):= {\delta \over \delta+(1-\delta)\, \exp(Y)}, \quad  Y_t:= \sum_{i=0}^{N_t} \beta_i\quad \mbox{and}\quad \beta_i:=\ln\left({\cal L}(x_{t_i},{\cal E}_{t_i})\right).$$

Using It\^o's lemma, we can express $\delta_t$ as the solution of the SDE
\begin{eqnarray*}\D \delta_t &=& f'(Y_{t-})\,\D Y_t+f(Y_t)-f(Y_{t-})-f'(Y_{t-})\,\Delta Y_t \\
&=& f(Y_t)-f(Y_{t-})\\
& =& \big(f(Y_{t-}+\beta_{N_t})-f(Y_{t-})\big)\, \D N_{t}\\
&=& (1-\delta_{t-})\,\delta_{t-}\, \left({Q(x_t,{\cal E}_t,\thetaL)-Q(x_t,{\cal E}_t,\thetaH)\over Q(x_t,{\cal E}_t,\thetaL)\,\delta_{t-} +Q(x_t,{\cal E}_t,\thetaH)\,(1-\delta_{t-})}\right)\, \D N_{t}.
\end{eqnarray*}
where the second equality follows from the fact that $Y_t$ is a pure jump process, {\em i.e.}, $\D Y_t = \Delta Y_t$. $\Box$\vspace{0.5cm}

{\sc Proof of  Proposition~\ref{prop:weakconv}:} To prove the result we invoke Theorem 4.21 in Chapter IX in  \cite{Jacod_Shiryaev} related to the convergence of Markov processes to diffusions. To this end, note that from Lemma~\ref{lem:SDEdelta} it follows that the belief process $\delta^k_t$ is a pure jump Markov process that admits a generator of the form
$${\cal A}^k f(\delta)=\int_{y \in (0,1)} [f(\delta+y)-f(\delta)]\,K^k(\delta,\D y),$$
where the kernel $K^k(\delta,y)$ satisfies
$$ \int_{y \in (0,1)}  f(y)\, K^k(\delta,\D y)=\Lambda^k \,\sum_{{\cal E} \in \mathscr{E}}\,\sum_{x \in {\cal E}} f(\eta^k(\delta,x,{\cal E}))\,Q^k_\delta(x,{\cal E})\,\pi(\delta,{\cal E})$$
where $Q^k_\delta(x,{\cal E}):= \delta\,Q^k(x,{\cal E},\thetaL) +(1-\delta)\,Q^k(x,{\cal E},\thetaH)$ and
$$\eta^k(\delta,x,{\cal E}):=(1-\delta)\,\delta\, \left({1-{\cal L}^k(x,{\cal E}) \over \delta+(1-\delta)\,{\cal L}^k(x,{\cal E}) }\right).$$
It follows that the instantaneous drift and volatility of $\delta^k_t$ are given by
\begin{align*}
b^k(\delta):=\int_{y \in (0,1)} y\, K^k(\delta,\D y) &= \Lambda^k\,\sum_{{\cal E} \in \mathscr{E}}\,\sum_{x \in {\cal E}} \eta^k(\delta,x,{\cal E})\,Q^k_\delta(x,{\cal E})\,\pi(\delta,{\cal E})=0
\end{align*}
and
\begin{align*}
c^k(\delta):=\int_{y \in (0,1)} y^2\, K^k(\delta,\D y) &= \Lambda^k \,\sum_{{\cal E} \in \mathscr{E}}\,\sum_{x \in {\cal E}} (\eta^k(\delta,x,{\cal E}))^2\,Q^k_\delta(x,{\cal E})\,\pi(\delta,{\cal E})\\
&= \Lambda^k\,\delta\,(1-\delta)\,\sum_{{\cal E} \in \mathscr{E}}\, \sum_{x \in {\cal E}}{\big(Q^k(x,{\cal E},\thetaL)-Q^k(x,{\cal E},\thetaH) \big)^2\over
Q^k_\delta(x,{\cal E})}\,\pi(\delta,{\cal E})\\
&=\Lambda\,\delta\,(1-\delta)\,\sum_{{\cal E} \in \mathscr{E}}\, \sum_{x \in {\cal E}} {\big(\alpha^k(x,{\cal E},\thetaL)-\alpha^k(x,{\cal E},\thetaL)\big)^2 \,{\cal Q}^k(x,{\cal E}) \over 1+(\delta\,\alpha^k(x,{\cal E},\thetaL)+(1-\delta)\,\alpha^k(x,{\cal E},\thetaH))/\sqrt{k}}\,\pi(\delta,{\cal E}).
\end{align*}
It follows by Assumption~\ref{assm:asymregime} that
\begin{align*}b(\delta)&:=\lim_{k \to \infty} b^k(\delta)=0 \qquad \mbox{and}\\
 c(\delta)&:=\lim_{k \to \infty} c^k(\delta)=\Lambda\,\delta\,(1-\delta)\,\sum_{{\cal E} \in \mathscr{E}}\,\sum_{x \in {\cal E}} \big(\alpha(x,{\cal E},\thetaL)-\alpha(x,{\cal E},\thetaH)\big)^2\, {\cal Q}(x,{\cal E})\,\pi(\delta,{\cal E}).\end{align*}
(Note that the convergence of $b^k(\delta)$ and $c^k(\delta)$ is trivially locally uniformly in $(0,1)$).

Since the jump size $\eta^k(\delta,x,{\cal E})$ converges to zero as $k \to \infty$ uniformly in $\delta$ for all ${\cal E}$ and all $x \in {\cal E}$, we get that for all $\epsilon >0$
\begin{align*}\sup_{\delta \in (0,1)} \int_{y \in (0,1)} y^2\, \Indi(y >\epsilon) \,K^k(\delta,\D y)&= \Lambda^k\,\sum_{{\cal E} \in \mathscr{E}}\,\sum_{x \in {\cal E}} (\eta^k(\delta,x,{\cal E}))^2\,\Indi(|\eta^k(\delta,x,{\cal E})|>\epsilon)\,  Q^k_\delta(x,{\cal E})\,\pi(\delta,{\cal E})\\
& \to 0 \qquad \mbox{as } k \to \infty.
\end{align*}
To conclude, note that $b(\delta)$ is trivially bounded and $c(\delta)$ is continuous and bounded in $(0,1)$ since are assuming that $\pi \in {\cal M}_c$.  We also have $c(\delta)>0$ for all $\delta \in (0,1)$ since we have assumed that every experiment ${\cal E}$ is informative and so we must have $\sum_{x \in {\cal E}} \big(\alpha(x,{\cal E},\thetaL)-\alpha(x,{\cal E},\thetaH)\big)^2>0$. So, by Theorem 2.34 in Chapter III in \cite{Jacod_Shiryaev} the semimartingale problem with characteristics $(b,c)$ has a unique solution for every initial condition $\delta \in (0,1)$. In sum,  all the required conditions in  Theorem 4.21 in Chapter IX in  \cite{Jacod_Shiryaev} are satisfied and so $\delta^k_t$ converges weakly to a diffusion process $\tilde{\delta}_t$ with characteristics $(b,c)$. $\Box$ \vspace{0.5cm}

{\sc Proof of  Proposition~\ref{prop:L1convergence}:} Consider an arbitrary instance of the problem, and let $(\pi, {\cal I}) \in {\cal M}(\mathscr{E}) \times {\cal B}$  be an optimal Markovian policy with corresponding value function $\Pi(\delta)$.  For future references, we recall that for any bounded function $f(\delta)$, the value-iteration recursion 
\begin{equation}\label{eq:recurPi}\Pi_{j+1}(\delta)=\Indi(\delta \in {\cal I})\,G(\delta) + \Indi(\delta \in {\cal I}^c)\, \rho\,\e_\pi\Big[\Pi_j(\delta_1)|\delta_0=\delta\Big], \quad \Pi_0(\delta)=f(\delta),\; \mbox{where } \rho :={\Lambda \over \Lambda +r}\end{equation}
produces a sequence of functions  $\{\Pi_j\}_{j \geq 0}$ that converges pointwise to $\Pi$. In \eqref{eq:recurPi}, $\delta_1$ denotes the value of the belief process after one jump (vote) and  $\e_\pi[\cdot]$ is the expectation operator induced by policy $(\pi,{\cal I})$, which satisfies 
$$\e_\pi[f(\delta_{j+1})|\delta_j=\delta]= \Indi(\delta \in {\cal I})\,f(\delta)+ \Indi(\delta \in {\cal I}^c)\, \sum_{{\cal E} \in \mathscr{E}}\, \sum_{x \in {\cal E}} f(\delta+\eta(\delta,x,{\cal E}))\,Q_\delta(x,{\cal E})\,\pi(\delta,{\cal E}).$$
Note that in \eqref{eq:recurPi} and in the definition of $\e_\pi[\cdot]$ have used the fact that the set ${\cal I}$ is absorbing under policy $(\pi,{\cal I})$. Also, the fact that 
 $\{\Pi_j\}_{j \geq 0}$  converges pointwise to $\Pi$ follows by a standard contraction mapping argument (e.g., Chapter 6 in  \citealp{Puterman05}).
\vspace{0.1cm}

Now, since the class of continuous experimentation strategies ${\cal M}_c(\Delta\mathscr{E})$ is dense in ${\cal M}(\Delta\mathscr{E})$ under the $L^1$ norm, there exists a sequence of continuous strategies $\{\hat{\pi}_n\}_{n \geq 1}$ in ${\cal M}_c(\Delta\mathscr{E})$  that converges in $L^1$ to $\pi$. Let us denote by $\widehat{\Pi}_n$ be the expected payoff function under the policy $(\hat{\pi}_n,{\cal I})$. It follows that for each $n$, the function $\widehat{\Pi}_n$ satisfies 
the fixed-point condition 
 \begin{equation}\label{eq:Pihat}\widehat{\Pi}_n(\delta)=\Indi(\delta \in {\cal I})\,G(\delta)+ \Indi(\delta \in {\cal I}^c)\,\rho\,\e_{\hat{\pi}_n}\Big[\widehat{\Pi}_n(\delta_1)|\delta_0=\delta\Big],\end{equation}
where the expectation operator $\e_{\hat{\pi}_n}[\cdot]$ under policy $(\hat{\pi}_n,{\cal I})$ is defined in a similar way to $\e_\pi[\cdot]$ above. One fact to keep in mind is that by the optimality of $(\pi,{\cal I})$ we have that $\Pi(\delta) \geq \widehat{\Pi}_n(\delta)$.
\vspace{0.2cm}

We want to show that $\widehat{\Pi}_n$ converges to $\Pi(\delta)$ in $L^1$ as $n \to \infty$. To this end, let us use the recursion in \eqref{eq:recurPi} with initial condition $\Pi_0(\delta)=\widehat{\Pi}_n(\delta)$. Combining  \eqref{eq:recurPi} and \eqref{eq:Pihat}, one can show that
\begin{align*}\Pi_1(\delta)&=\widehat{\Pi}_n(\delta) + \Indi(\delta \in {\cal I}^c)\,\rho\, \sum_{{\cal E} \in \mathscr{E}}\, \sum_{x \in {\cal E}} \widehat{\Pi}_n(\delta+\eta(\delta,x,{\cal E}))\,Q_\delta(x,{\cal E})\,[\pi(\delta,{\cal E})-\hat{\pi}_n(\delta,{\cal E})]\leq \widehat{\Pi}_n(\delta)  + F_n(\delta),\end{align*}
where
$$F_n(\delta) := \Indi(\delta \in {\cal I}^c)\,\rho\,\max_{\delta}\Big\{ \widehat{\Pi}_n(\delta)\Big\}\, \sum_{{\cal E} \in \mathscr{E}} \big|\pi(\delta,{\cal E})-\hat{\pi}_n(\delta,{\cal E})\big| .$$
We note that $\|F_n\|_1  \leq K \|\pi-\hat{\pi}_n\|_1$ for some fixed constant $K$. (The fact that we can choose $K$ independent of $n$ follows from the fact that the $\widehat{\Pi}_n$ are uniformly bounded above by $\Pi$.)\vspace{0.2cm}

Let us iterate the recursion \eqref{eq:recurPi} one more time for $\Pi_2$. Using the inequality $\Pi_1(\delta) \leq \widehat{\Pi}_n(\delta)  + F_n(\delta)$, we get that
$$\Pi_2(\delta) \leq \widehat{\Pi}_n(\delta)  + F_n(\delta) + \rho\,\e_\pi\Big[F_n(\delta_1)|\delta_0=\delta\Big].$$
If we keep iterating this inequality we get that
$$\Pi_{j}(\delta)\leq \widehat{\Pi}_n(\delta)  + \e_\pi\left[\sum_{\ell=0}^{j-1} \rho^\ell \,F_n(\delta_\ell)\big|\delta_0=\delta\right],$$
where $\delta_\ell$ denotes the state of the belief process after $\ell$ jumps (votes). 
Let us denote by $J$ the random time at which ${\delta_\ell}$ enters ${\cal I}$, that is, $J:=\inf\{\ell \geq 0 \colon \delta_\ell \in {\cal I}\}$. Since $F_n(\delta_\ell)=0$ for all $\ell \geq J$ and $F_n(\delta) \geq 0$, we have that

$$\Pi_{j}(\delta)\leq \widehat{\Pi}_n(\delta)  + \e_\pi\left[\sum_{\ell=0}^{J-1} \rho^\ell \,F_n(\delta_\ell)\big|\delta_0=\delta\right].$$
Hence, taking limit as $j  \uparrow \infty$ and using the pointwise convergence of $\{\Pi_j\}$ to $\Pi$, we get
$$\Pi(\delta)\leq \widehat{\Pi}_n(\delta)  + \e_\pi\left[\sum_{\ell=0}^{J-1}\rho^\ell\,F_n(\delta_\ell)\big|\delta_0=\delta\right].$$
Since $\Pi$ is the optimal value function, it follows that $\Pi(\delta)\geq \widehat{\Pi}_n(\delta)$ and so 
$$\|\Pi-\widehat{\Pi}_n\|_1 \leq \int_0^1  \e_\pi\left[\sum_{\ell=0}^{J-1} \rho^\ell \, F_n(\delta_\ell)\big|\delta_0=\delta\right]\, \D \delta.$$
Next, we use a localization argument. Let us define  $\bar{F}:=\sup_n \max_\delta\{F_n(\delta)\}$ and let $T$ be a fixed nonnegative integer. Then, the previous inequality implies

$$\|\Pi-\widehat{\Pi}_n\|_1 \leq \sum_{\ell=0}^{T-1} \rho^\ell \, \int_0^1 \e_\pi[F_n(\delta_\ell)\big|\delta_0=\delta]\,\D \delta + \bar{F} {\Lambda \over r}\, \rho^{T-1}\,\e_\pi\left[1- \rho^{(J-T)^+} \right]. $$

To complete the proof, we will show in Lemma~\ref{lem:equivnorms} below  that there exists a constant $\bar{\mu}>0$ such that 
$$  \int_0^1 \e_\pi[F_n(\delta_\ell)\big|\delta_0=\delta]\,\D \delta \leq \bar{\mu}^\ell\, \|F_n\|_1 \leq K\, \bar{\mu}^\ell\,\|\pi-\hat{\pi}_n\|_1.  $$
As a result, 
$$\|\Pi-\widehat{\Pi}_n\|_1 \leq K\,\left(1-(\rho\,\bar{\mu})^T \over 1-\rho\,\bar{\mu}\right)\,\|\pi-\hat{\pi}_n\|_1 + \bar{F} {\Lambda \over r}\, \rho^{T-1}\,\e_\pi\left[1- \rho^{(J-T)^+} \right]. $$
Taking limit as $n \uparrow \infty$ and using the fact that $\|\pi-\hat{\pi}_n\|_1 \downarrow 0$ as $n \uparrow \infty$ we get that 
$$\lim_{n \to \infty} \|\Pi-\widehat{\Pi}_n\|_1 \leq  \bar{F} {\Lambda \over r}\, \rho^{T-1}\,\e_\pi\left[1- \rho^{(J-T)^+} \right].$$

Finally, since $T$ is arbitrary, we can let $T \uparrow \infty$ to complete the proof.~$\Box$

\begin{lem}\label{lem:equivnorms} There exists a constant $\bar{\mu}>0$ such that for any non-negative function $f(\delta)$ with $f(\delta)=0$ in ${\cal I}$ 
$$  \int_0^1 \e_\pi [f(\delta_\ell)\big|\delta_0=\delta]\,\D \delta \leq \bar{\mu}^\ell\, \|f\|_1.$$
\end{lem}
{\sf Proof of Lemma~\ref{lem:equivnorms}:} {\sf 
We use a proof by induction. Let us consider first the case $\ell=1$. From the definition of $\e_\pi[\cdot]$ we have that 
\begin{align}\nonumber
\int_0^1 \e_\pi [f(\delta_\ell)\big|\delta_0=\delta]\,\D \delta & \leq   \sum_{{\cal E} \in \mathscr{E}}\, \sum_{x \in {\cal E}} \int_0^1 f(\delta+\eta(\delta,x,{\cal E}))\,Q_\delta(x,{\cal E})\,\pi(\delta,{\cal E})\,\D \delta \\ \nonumber
&=  \sum_{{\cal E} \in \mathscr{E}}\, \sum_{x \in {\cal E}} \int_0^1 f\left(\delta \over \delta+ (1-\delta)\,{\cal L}(x,{\cal E})\right)\,Q_\delta(x,{\cal E})\,\pi(\delta,{\cal E})\,\D \delta \\ \nonumber
&= \sum_{{\cal E} \in \mathscr{E}}\, \sum_{x \in {\cal E}} \int_0^1 {f\left(u \right)\,Q_{\delta_u}(x,{\cal E})\,\pi(\delta_u,{\cal E})\,{\cal L}(x,{\cal E})  \over (1+{\cal L}(x,{\cal E})\,u-u)^2  } \D u \quad \mbox{with } \delta_u = {{\cal L}(x,{\cal E})\, u \over 1+{\cal L}(x,{\cal E})\,u-u} \\ \nonumber
& \leq \max_{{\cal E} \in \mathscr{E}} \max_{ x \in {\cal E}}  \left\{{\cal L}(x,{\cal E}), {1 \over {\cal L}(x,{\cal E})}\right\}\,  \sum_{{\cal E} \in \mathscr{E}}\, \sum_{x \in {\cal E}} \int_0^1 f\left(u \right)\,Q_{\delta_u}(x,{\cal E})\,\pi(\delta_u,{\cal E}) \D u \\
& =  \max_{{\cal E} \in \mathscr{E}} \max_{ x \in {\cal E}}  \left\{{\cal L}(x,{\cal E}), {1 \over {\cal L}(x,{\cal E})}\right\}\, \int_0^1 f(u)\, \D u . \label{eq:lemma3}
\end{align}
In the second inequality we have used the fact that 
$$\max_{u \in [0,1]} {L \over (1+L\,u-u)^2} \leq \max\left\{L,{1 \over L}\right\}.$$
So, the result in Lemma~\ref{lem:equivnorms} holds for $\ell=1$ with
$$\bar{\mu}:= \max_{{\cal E} \in \mathscr{E}} \max_{ x \in {\cal E}}  \left\{{\cal L}(x,{\cal E}), {1 \over {\cal L}(x,{\cal E})}\right\}.$$
Suppose that the result in Lemma~\ref{lem:equivnorms} is true for $j=1, \dots, \ell-1$. From the law of iterated expectations we have that $\e_\pi[f(\delta_\ell) |\delta_0=\delta]= \e_\pi[\e_\pi[f(\delta_\ell)|\delta_{\ell-1}]|\delta_0=\delta]=\e_\pi[g(\delta_{\ell-1})| \delta_0=\delta]$ with $g(\delta):= \e_\pi[f(\delta_\ell)|\delta_{\ell-1}=\delta]$.
But, by the Markov property this is the same as  $g(\delta):= \e_\pi[f(\delta_1)|\delta_{0}=\delta]$. 
It follows from the hypothesis of induction and the inequality \eqref{eq:lemma3} that
\begin{align*}  \int_0^1 \e_\pi [f(\delta_\ell)\big|\delta_0 &=\delta]\,\D \delta = \int_0^1 \e_\pi[g(\delta_{\ell-1})| \delta_0=\delta] \, \D \delta \leq \bar{\mu}^{\ell-1} \int_0^1g(\delta) \, \D \delta = \bar{\mu}^{\ell-1}\, \int_0^1 \e_\pi[f(\delta_1)|\delta_{0}=\delta]\, \D \delta\\
& \leq \bar{\mu}^\ell \, \|f\|_1. ~~\Box\end{align*}
}\vspace{0.5cm}

{\sc Proof of  Proposition~\ref{lem:timechange}:} For a given experimentation policy $\pi$ consider the mapping
$$T_t^\pi: = \int_0^t {1 \over \tilde{\sigma}^2(\tilde{\delta}_s,\pi)}\, \D s,$$
where $\tilde{\sigma}^2(\delta,\pi)$ is defined in equation \eqref{eq: volatility}. Since, $\tilde{\sigma}^2(\delta,\pi)>0$ for all $\delta$, the mapping $T_t^\pi$ is strictly increasing $t$ with $T_0^\pi=0$. As a result, let us view $T_t^\pi$ as a random time change and let us define the process
$$\hat{\delta}_t:=\tilde{\delta}_{T_t^\pi}.$$
Let ${\cal G}_{\tilde{\delta}}$ denote the infinitesimal generator of $\tilde{\delta}_t$. Then, by Proposition~\ref{prop:weakconv} it follows that
$${\cal G}_{\tilde{\delta}} =   \tilde{\sigma}^2(\delta,\pi)\, \delta^2\, (1- \delta)^2\, {\partial^2 \over \partial  \delta^2}.$$
It follows that the  infinitesimal generator  ${\cal G}_{\hat{\delta}}$ of $\hat{\delta}_t$ is given by
$$ {\cal G}_{\hat{\delta}} = \dot{T}_t^\pi\,{\cal G}_{\tilde{\delta}} = {1 \over \tilde{\sigma}^2(\delta,\pi)} \, \tilde{\sigma}^2(\delta,\pi)\, \delta^2\, (1- \delta)^2\, {\partial^2 \over \partial  \delta^2} =  \delta^2\, (1- \delta)^2\, {\partial^2 \over \partial  \delta^2}.$$
In other words, $\hat{\delta}_t$ is a diffusion process that satisfies the SDE
\begin{equation}\label{eq:Dhat}\D \hat{\delta}_t =  \hat{\delta}_t\, (1- \hat{\delta})\,  \D W_t,\end{equation}
for some Wiener process $W_t$. Also, for a given stopping time $\tau$ for $\tilde{\delta}_t$, let us define the stopping time $\hat{\tau}$ for $\hat{\delta}_t$ such that
$\hat{\delta}_{\hat{\tau}} = \tilde{\delta}_\tau$. It follows that
\begin{equation}\label{eq:timechange}\tau = \int_0^{\hat{\tau}}  {1 \over \tilde{\sigma}^2(\tilde{\delta}_s,\pi)}\, \D s.\end{equation}
Finally, the result in Proposition~\ref{lem:timechange} follows from equations \eqref{eq:Dhat} and \eqref{eq:timechange}. $\Box$

{}

\vspace{0.5cm}

{\sc Proof of Theorem~\ref{thm:verification}}:
Let $f$ be a solution to the QVI in equation~\eqref{eq:QVI}.
Given the assumptions on $f$, we can apply integration by parts followed by It\^o's lemma (see \citealp{Protter}) to get that
\begin{equation*}
e^{-r\,\tau}\,f(\delta_\tau) = f(\delta)+\int_0^\tau e^{-r\,t}\,{\cal H} f(\delta_t)\, \D t +\int_0^\tau e^{-r\,t} \tilde{\sigma}\, \delta_t\,(1-\delta_t)\,f'(\delta_t)\, \D W_t.
\end{equation*}
Note that the process
\begin{equation*}
 f(\delta)+\int_0^t e^{-r\,s} \tilde{\sigma}\, \delta_s\,(1-\delta_s)\,f'(\delta_s)\, \D W_s,
\end{equation*}
is a local martingale, thus, by the non-negativity of $f$,  also a supermartingale.
With this, one can take expectation, canceling the stochastic integral, and use the fact that ${\cal H} f(\delta) \leq 0$ (second QVI condition) to get that
\begin{equation*}
\e[e^{-r\,\tau}\,f(\delta_{\tau})] \leq f(\delta).
\end{equation*}
This inequality together with the first QVI condition imply
\begin{equation*}
\e[e^{-r\,\tau}\,\widetilde{G}(\delta_{\tau})] \leq \e[e^{-r\,\tau}\,f(\delta_{\tau})] \leq f(\delta).
\end{equation*}
Because these inequalities hold for any stopping time $\tau$, we conclude that $f(\delta) \geq \widetilde{\cal G}(\delta)$.
Finally, we note that all the inequalities above become equalities for the QVI-control associated to $f$.
This follows from Dynkin's formula and the fact that the QVI-control is the first exit time from a bounded set (continuation region ${\cal C}$). ~~$\Box$\vspace{0.5cm}

{\sc Proof of Proposition~\ref{prop:diffusion_2}}: We need to prove both the existence and optimality of the function $\widetilde{\cal G}_{ij}(\delta)$ in equation \eqref{eq: Close_form_general}.
Let us start by proving the optimality using the QVI conditions. \vspace{0.2cm}

The first step is to show that $\widetilde{\cal G}_{ij}(\delta)$ is convex in $[0,1]$. To see this note that in the continuation region $\delta \in(\deltau_{ij},\deltab_{ij})$ the function $\widetilde\cG_{ij}(\delta)=C_{ij}^0\, (1-{\delta})^\gamma\, {\delta}^{1-\gamma}+C_{ij}^1\,(1-{\delta})^{1-\gamma}\,{\delta}^\gamma$ is convex. This follows from the fact that by construction it satisfies the ODE ${\cal H}\widetilde\cG_{ij}(\delta)=0$, and so
$${(\tilde{\sigma}\,\delta\,(1-\delta))^2 \over 2} \, \widetilde\cG_{ij}''(\delta) =r\,\widetilde\cG_{ij}(\delta) \geq 0$$
This together with the value matching and smooth pasting conditions at $\deltab_{ij}$ and $\deltab_{ij}$ ensure that $\widetilde{\cal G}_{ij}(\delta)$ is convex in $[0,1]$.

Now, by convexity and the smooth-pasting and value matching conditions, both $\widetilde{\cal R}_i(\delta)$ and $\widetilde{\cal R}_j(\delta)$ are supporting hyperplanes of $\widetilde{\cal G}_{ij}(\delta)$ in the domain $\delta \in [0,1]$. We conclude that the first QVI condition holds, that is, $\widetilde{\cal G}_{ij}(\delta) \geq
\widetilde{G}_{ij}(\delta)=\max\{\widetilde{\cal R}_i(\delta),\widetilde{\cal R}_j(\delta)\}$.

To prove the second and third QVI conditions note that in the continuation region $\delta \in(\deltau_{ij},\deltab_{ij})$, we have ${\cal H}\widetilde{\cal G}_{ij}(\delta)=0$ (by construction). On the other hand, in the
intervention region $\delta \in [0,\deltau_{ij}]\cup [\deltab_{ij},1]$, we have that (a) $\widetilde{\cal G}_{ij}(\delta)=\widetilde{G}_{ij}(\delta)$ and  (b) ${\cal H}\widetilde{\cal G}_{ij}(\delta)=-r\,\widetilde{G}_{ij}(\delta) \leq 0$.

Finally, if we define the set $N_{ij}=\{\deltau_{ij},\deltab_{ij},\hat{\delta}_{ij}\}$, it is easy to see that the function
$\widetilde{\cal G}_{ij}(\delta)$ is in ${\cal C}^1[0,1]$ and has second derivative for all $\delta \in [0,1]\setminus N_{ij}$.  We conclude that $\widetilde{\cal G}_{ij}(\delta) \in \hat{\cal C}^2$  and so by Theorem~\ref{thm:verification} it is optimal.\vspace{0.2cm}

Let us now turn to the issue of existence. For this, we need to show that there exist thresholds $\deltau_{ij}$ and $\deltab_{ij}$ so that the smooth pasting and value matching conditions are satisfied. To fix ideas, let us suppose that $\widetilde{\cal R}_i(0) \geq \widetilde{\cal R}_j(0)$  and let us consider the auxiliary function

$$V(\delta;\deltau):=\left\{\begin{array}{cl}
\widetilde{\cal R}_i(\delta) & \mbox{if}\qquad 0 \leq {\delta} \leq \deltau \\ & \\
C_{ij}^0(\deltau)\, (1-{\delta})^\gamma\, {\delta}^{1-\gamma}+C_{ij}^1(\deltau)\,(1-{\delta})^{1-\gamma}\,{\delta}^\gamma  & \mbox{if}\qquad   \deltau \leq {\delta} \leq 1\end{array} \right. $$
where the parameter $\deltau \in [0, \hat{\delta}_{ij}]$ and the constants $C_{ij}^0(\deltau)$ and $C_{ij}^1(\deltau)$
are chosen to ensure value matching and smooth pasting at $\delta=\deltau$. Recall that the payoff associated to each action is linear in $\delta$, that is, of the form $\widetilde{\cal R}_i(\delta) =\tilde{\alpha}_i+\tilde{\beta}_i\,\delta$. It follows that the constants $C_{ij}^0(\deltau)$ and $C_{ij}^1(\deltau)$ are equal to
$$C_{ij}^0(\deltau)=\left[(\gamma-1)\,\deltau\,\tilde{\beta}_i+(\gamma-\deltau)\,\tilde{\alpha}_i \over (2\,\gamma-1)\,\deltau\right]\,\left(\deltau \over 1-\deltau\right)^\gamma\quad \mbox{and}\quad
C_{ij}^1(\deltau)=\left[\gamma\,\deltau\,\tilde{\beta}_i+(\gamma-1+\deltau)\,\tilde{\alpha}_i \over (2\,\gamma-1)\,\deltau\right]\,\left(1-\deltau \over \deltau\right)^{\gamma-1}.$$

\begin{figure}[htb]
    \begin{center}
    \includegraphics[width=12cm]{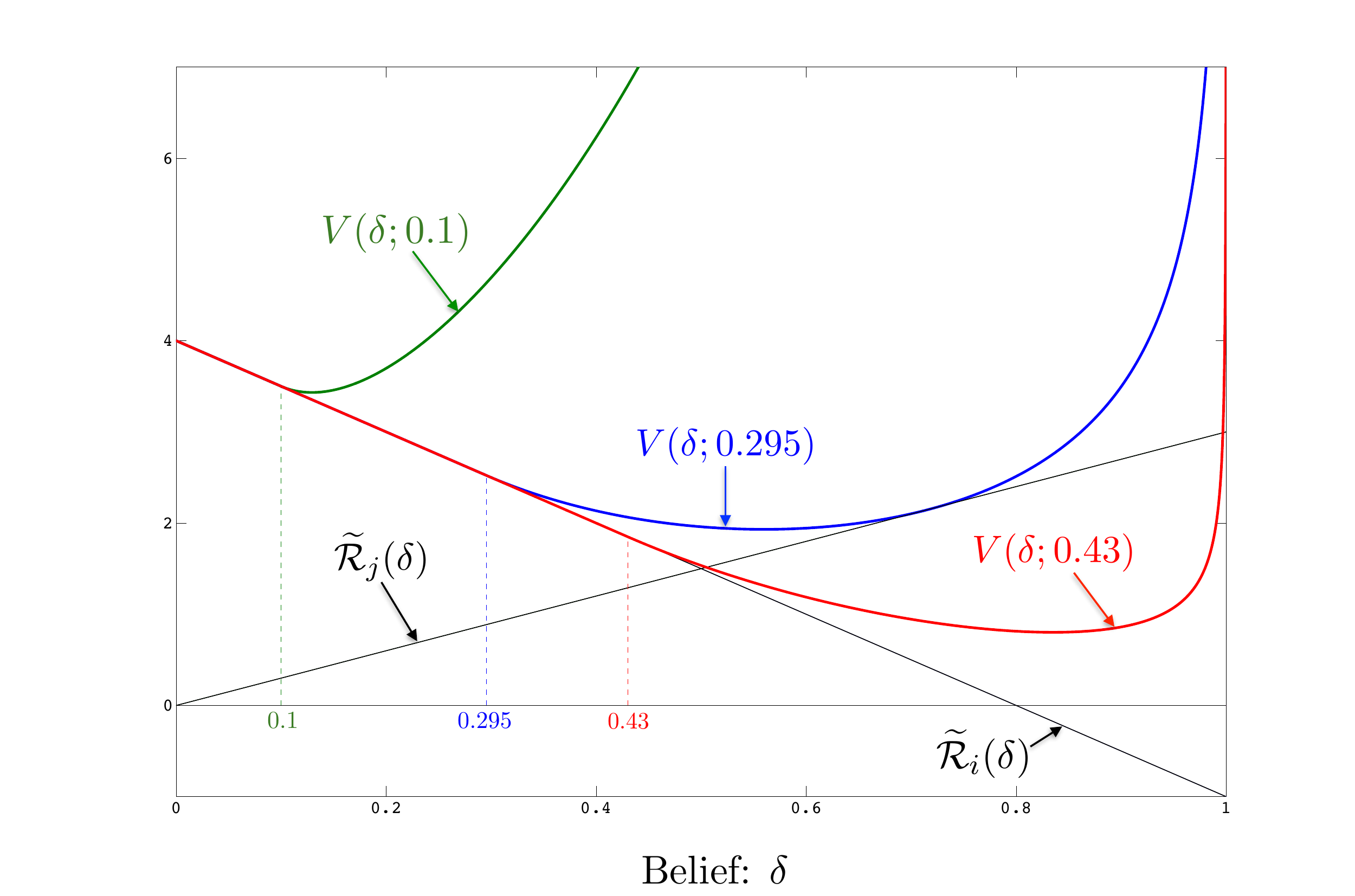}\vspace{-0.5cm}
    \end{center}\caption{\footnotesize \sf Value of $V(\delta;\deltau)$ for three values of $\deltau \in \{0.1,  0.295, 0.43\}$.
    For $\deltau=0.295$, the function $V(\delta;\deltau)$ satisfies the smooth-pasting condition at $\deltab$. {\sc Data}: $\widetilde{\cal R}_j(\delta)=3\,\delta$, $\widetilde{\cal R}_i(\delta)=4-5\,\delta$, $r=1$ and $\tilde{\sigma}=2$.}    \label{fig:ProofProp7}
\end{figure}

Of course, by construction the function $V(\delta;\deltau)$ satisfies the value matching and smooth pasting conditions at $\deltau$. Next, we show that by varying the value of $\deltau$ we can also enforce these
conditions at the upper threshold $\deltab$. To get some intuition, consider the example in Figure~\ref{fig:ProofProp7} which depicts the function $ V(\delta;\deltau)$ for three different values of $\deltau \in \{0.1, 0.295, 0.43\}$. The figure also shows the payoff functions $\widetilde{\cal R}_i(\delta)$ and $ \widetilde{\cal R}_j(\delta)$. We note that when $\deltau$ is small (in the example $\deltau=0.1$) the function $ V(\delta;\deltau)$ is greater than $\widetilde{\cal R}_j(\delta)$ for all $\delta \geq \deltau$. On the opposite case, when  $\deltau$ is large (in the example $\deltau=0.43$) the function $ V(\delta;\deltau)$ intersects $\widetilde{\cal R}_j(\delta)$ for some $\delta \geq \deltau$. By continuity, there is a value of $\deltau$ (in the example $\deltau=0.295$) so that $ V(\delta;\deltau)$ and  $\widetilde{\cal R}_j(\delta)$ meet smoothly at some $\deltab \geq \deltau$.

To formalize the previous discussion based on the example in Figure~\ref{fig:ProofProp7}, let us first note that $\lim_{\deltau \downarrow 0} C_{ij}^0(\deltau)=0$ and
$\lim_{\deltau \downarrow 0} C_{ij}^1(\deltau)=\infty$ (recall that $\gamma>1$). Hence, the $\lim_{\deltau \downarrow 0} V(\delta,\deltau)=\infty$ for all $\delta \in (0,1]$. This shows that if $\deltau$ is sufficiently small the function $V(\delta,\deltau)$ will be strictly greater than $\widetilde{\cal R}_j(\delta)$ for all $\delta \geq \deltau$.

On the flip side, we have that $\lim_{\deltau \to \hat{\delta}_{ij}} V(\deltau,\deltau)=\widetilde{\cal R}_j(\deltau)$ and $\lim_{\deltau \to \hat{\delta}_{ij}} V'(\deltau,\deltau)=\tilde{\beta}_i < \tilde{\beta}_j=\widetilde{\cal R}_j'(\deltau)$. In other words, if $\deltau$ is sufficiently close to $\hat{\delta}_{ij}$ then the function $V(\delta,\deltau)$ will intersect and go below the function $\widetilde{\cal R}_j(\delta)$.

Finally, since the function $V(\delta,\deltau)$ is continues in $\deltau$, we conclude that there exists a value $\deltau=\deltau_{ij} \in [0,\hat{\delta}_{ij}]$ such that
$$\min_{\delta \in [\hat{\delta}_{ij},1]} \big\{V(\delta,\deltau_{ij})-\widetilde{\cal R}_j(\delta)\}=0.$$
The value of $\delta$ that solves the minimization is the upper threshold $\deltab_{ij}$. ~~$\Box$

\vspace{0.5cm}

{\sc Proof of Corollary~\ref{cor:diffusion}}:  For notational convenience, let us write $\deltau=\deltau_{ij}$ and $\deltab=\deltab_{ij}$.

To compute the probability $\bar{p}(\delta)=\Pro(\tilde{\delta}_{\tau^*}=\deltab|\tilde{\delta}_0=\delta)$, we use Dynkin's formula to get
$$\e[f(\tilde{\delta}_{\tau^*})]=f(\delta)+\e\left[\int_0^{\tau^*} {\cal G}f(\tilde{\delta}_t)\, \D t\right], \qquad \mbox{where}\quad {\cal G}f(\delta):= {(\tilde{\sigma}\,\delta\,(1-\delta))^2 \over 2}\, {\D^2 f(\delta) \over \D \delta^2}, $$and ${\cal G}$ is the infinitesimal generator of the diffusion process $\tilde{\delta}_t$ in equation \eqref{eq:diffusionprobformobf2}. Consider the identity function $f(\delta)=\delta$. It follows that ${\cal G}f(\delta)=0$ and by Dynkin's formula $\e[\tilde{\delta}_{\tau^*}]=\delta$. But since $\tau^*$ is the first exit time of the process $\tilde{\delta}_t$ from the continuation region $(\deltau,\deltab)$ we have that $\e[\tilde{\delta}_{\tau^*}]=\bar{p}(\delta)\,\deltab+(1-\bar{p}(\delta))\,\deltau$ and the result part of the Corollary follows.

To compute the expectation $\e[\tau^*]$, we consider a function ${\mathcal{T}}(\delta)$ such that ${\cal G}({\mathcal{T}})(\delta)=1$. One can verify that the function ${\mathcal{T}}(\delta)={2 \over \tilde{\sigma}^2}\,(2\delta-1)\,\ln\left(\delta \over 1-\delta\right)$ satisfies this condition. It follows from Dynkin's formula that
$\e[{\mathcal{T}}(\tilde{\delta}_{\tau^*})]=\mathcal{T}(\delta)+\e[\tau^*]$ and  the result follows. $\Box$
\vspace{0.5cm}

{\sc Proof of Proposition~\ref{prop:QVI-1}}: By Proposition~\ref{prop:diffusion_2}, if follows that $\widetilde{\cal G}_{ij}(\delta) \geq \widetilde{G}_{ij}(\delta)$ for all $\delta \in [0,1]$. As a result, $\widetilde{V}(\delta)=\max_{\{i,j\} \in \widetilde{\cal O}} \{\widetilde{\cal G}_{ij}(\delta)\} \geq \max_{\{i,j\} \in \widetilde{\cal O}} \{\widetilde{ G}_{ij}(\delta)\} = \widetilde{G}(\delta)$.\vspace{0.2cm}

By Proposition~\ref{prop:diffusion_2}, we know that each function $\widetilde{\cal G}_{ij}(\delta)$ is continuously differentiable everywhere in $[0,1]$ and admits a second derivative almost everywhere in $[0,1]$ except in the set $N_{ij}:=\{\deltau_{ij},\deltab_{ij}\}$. Also, two functions $\widetilde{\cal G}_{ij}(\delta)$ and $\widetilde{\cal G}_{k\ell}(\delta)$ can cross at most a finite number of times. This follows from noticing that in the continuation regions the functions
$C_{ij}^0\, (1-{\delta})^\gamma\, {\delta}^{1-\gamma}+C_{ij}^1\,(1-{\delta})^{1-\gamma}\,{\delta}^\gamma$  and $C_{k\ell}^0\, (1-{\delta})^\gamma\, {\delta}^{1-\gamma}+C_{k\ell}^1\,(1-{\delta})^{1-\gamma}\,{\delta}^\gamma$ can only cross at most once.  Let us denote by $E_{ij,k\ell}$  the finite set of values of $\delta$ at which these two functions cross (if any) and let us define
$$N_{\mbox{\tiny \rm $\widetilde{V}$}} = \bigcup_{i,j \in \widetilde{\cal O}}N_{ij} \,\cup \bigcup_{i,j,k,\ell \in \widetilde{\cal O}}\{E_{ij,k\ell}\}.$$

Now, by our previous construction, it follows that for each $\delta \in [0,1]\setminus N_{\mbox{\tiny \rm $\widetilde{V}$}}$ there exists an open neighborhood $B(\delta)$ containing $\delta$ such that
$\widetilde{V}(x)=\widetilde{\cal G}_{ij}(x)$ for all $x \in B(\delta)$  for some pair $\{i,j\} \in \widetilde{\cal O}$. Furthermore, the function $\widetilde{\cal G}_{ij}(x)$ is twice-continuously differentiable in $B(\delta)$. Since each function $\widetilde{\cal G}_{ij}(x)$ satisfies the QVI conditions, we conclude that ${\cal H}\widetilde{V}(\delta) \leq 0$ and $\big(\widetilde{V}(\delta)-\widetilde{G}(\delta)\big)\,\mathcal{H}\widetilde{V}(\delta)=0$ for all $\delta \in [0,1]\setminus N_{\mbox{\tiny \rm $\widetilde{V}$}}$.~~$\Box$\vspace{0.5cm}

{\sc Proof of Theorem~\ref{thm:Vsmooth}}: We wish to prove that the function $\widetilde{V}(\delta)=\max_{\{i,j\} \in \widetilde{\cal O}} \big\{\widetilde\cG_{ij}({\delta}) \big\}$ is in
$\widehat{\cal C}^2$, where each function $\widetilde\cG_{ij}({\delta})$ is convex and in $\widehat{\cal C}^2$ (Proposition~\ref{prop:diffusion_2}). Furthermore, each function $\widetilde\cG_{ij}({\delta})$ solves the optimization problem
\begin{equation}\label{eq:ProofThm2}\widetilde\cG_{ij}({\delta})=\sup_{\tau \in \mathbb{T}}
\e\left[e^{-r\,\tau}\,\max\big\{ \widetilde{\cal R}_i(\delta),\widetilde{\cal R}_j(\delta)  \big\}\right] \qquad
\mbox{subject to} \qquad\D \,\tilde{\delta_t} = \tilde{\sigma}\,\tilde{\delta}_t\,(1-\tilde{\delta}_t)\,\D W_t, \qquad \tilde{\delta}_0=\delta.
\end{equation}

Let us suppose, by contradiction, that $\widetilde{V}(\delta)$ is not in $\widehat{\cal C}^2$, then it must exist a $\delta_\star$ at which $\widetilde{V}(\delta)$ is not differentiable. But since each $\widetilde\cG_{ij}({\delta})$ is smooth it follows that there are at least two functions $\widetilde\cG_{ij}({\delta})$ and $\widetilde\cG_{k\ell}({\delta})$ that intersect and that simultaneously solve the maximization in the definition of
$\widetilde{V}(\delta)$ at this value $\delta_\star$. That is,
$$\widetilde{V}(\delta_\star)= \widetilde\cG_{ij}({\delta_\star})=\widetilde\cG_{k\ell}({\delta}_\star)\qquad \mbox{and}\qquad {{\rm d} \over {\rm d}\delta} \widetilde\cG_{ij}({\delta_\star}) \not={{\rm d} \over {\rm d}\delta} \widetilde\cG_{k\ell}({\delta_\star}). $$

\begin{figure}[htb]
    \begin{center}
    \includegraphics[width=7cm]{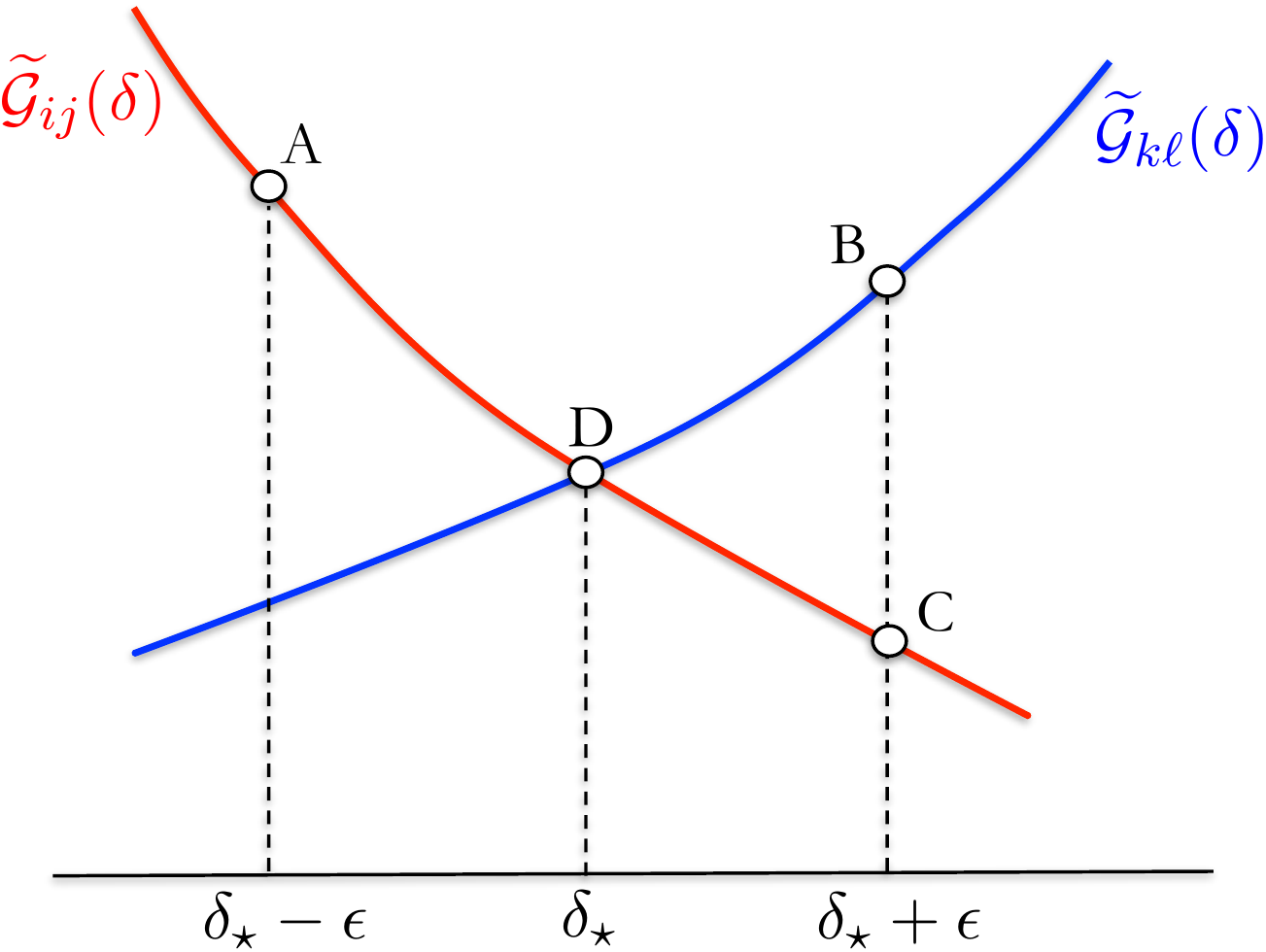}\vspace{-0.5cm}
    \end{center}\caption{\footnotesize \sf  Schematic of what needs to happen for $\widetilde{V}(\delta) = \max\big\{\widetilde\cG_{ij}({\delta}),\widetilde\cG_{k\ell}({\delta})\big\}$ to be non-smooth at some point $\delta_\star$.}    \label{fig:ProofThem2}
\end{figure}

To fix ideas, let us suppose that $\widetilde{V}(\delta)=\widetilde\cG_{ij}({\delta})$ for all $\delta \in (\delta_\star-\epsilon,\delta_\star]$ and $\widetilde{V}(\delta)=\widetilde\cG_{k\ell}({\delta})$ for all $\delta \in [\delta_\star,\delta_\star+\epsilon)$ for some small $\epsilon >0$
(as in Figure~\ref{fig:ProofThem2}). In this case, the two conditions above imply that $\widetilde\cG_{k\ell}(\delta_\star+\epsilon)>\widetilde\cG_{ij}(\delta_\star+\epsilon)$ (i.e., point B is above point C).

We will now show that point D cannot belong to $\widetilde{V}(\delta)$. For this, we will exploit the optimality
of the functions $\widetilde\cG_{ij}({\delta})$ and $\widetilde\cG_{k\ell}({\delta})$ in the sense of equation \eqref{eq:ProofThm2}. We distinguish two cases:

\begin{enumerate}
\item Suppose that $\delta_\star$ belongs to at least one of the continuation regions ${\cal C}_{ij}=(\deltau_{ij} , \deltab_{ij})$ or ${\cal C}_{k\ell}=(\deltau_{k\ell} , \deltab_{k\ell})$ associated to $\widetilde\cG_{ij}({\delta})$ and $\widetilde\cG_{k\ell}({\delta})$, respectively (see equation \eqref{eq: Close_form_general} in Proposition~\ref{prop:diffusion_2}). For concreteness let us assume that $\delta_\star \in {\cal C}_{ij}$.

By choosing $\epsilon$ small enough we can guarantee that both $\delta_\star-\epsilon$ and $\delta_\star+\epsilon$ also belong to ${\cal C}_{ij}$.
It follows then, by the principle of optimality, that
$$\widetilde\cG_{ij}({\delta_\star}) = \e\left[e^{-r\,\tau_\star}\, \widetilde\cG_{ij}(\tilde\delta_{\tau_\star})\right], \quad \mbox{where}\quad\tau_\star:=\inf\big\{t >0: \tilde\delta_t \not\in (\delta_\star-\epsilon,\delta_\star+\epsilon)\big\}.$$
In words, this identity states that the value $\widetilde\cG_{ij}({\delta_\star})$ can be obtained by letting the belief process $\tilde\delta_t$ evolve in the region $(\delta_\star-\epsilon,\delta_\star+\epsilon)$ and as soon
as one of the boundaries is hit, then the corresponding value of  $\widetilde\cG_{ij}(\delta)$ is collected (point A if the left boundary is hit first or point C if the right boundary is hit first).

Now, using the fact that $\widetilde\cG_{k\ell}(\delta_\star+\epsilon)>\widetilde\cG_{ij}(\delta_\star+\epsilon)$  we get that
$$\widetilde\cG_{ij}({\delta_\star}) <  \e\left[e^{-r\,\tau_\star}\, \max\Big\{\widetilde\cG_{ij}(\tilde\delta_{\tau_\star}),\widetilde\cG_{k\ell}(\tilde\delta_{\tau_\star})\Big\}\right].$$

But $\widetilde{V}(\delta_\star)=\widetilde\cG_{ij}({\delta_\star})$ and so the previous inequality contradicts the optimality of $\widetilde{V}(\delta_\star)$.

\item Let us now suppose  that $\delta_\star$ belong to both intervention regions ${\cal I}_{ij}:=[0,\deltau_{ij}]\cup[\deltab_{ij},1]$ and ${\cal I}_{k\ell}:=[0,\deltau_{k\ell}]\cup[\deltab_{k\ell},1]$. Without loss of generality, let us  assume that $\widetilde\cG_{ij}({\delta_\star})=\widetilde{\cal R}_i({\delta_\star})$
and $\widetilde\cG_{k\ell}({\delta_\star})=\widetilde{\cal R}_k({\delta_\star})$. That is, $\delta_\star=\hat{\delta}_{ik}$ the intersection point of $\widetilde{\cal R}_i(\delta)$ and $\widetilde{\cal R}_k(\delta)$
(see Figure~\ref{fig:ExTwoPieces} in Section~\ref{subsec:O=2}).
But, from Proposition~\ref{prop:diffusion_2} we have that
$$\widetilde{\cal R}_i(\delta_\star)<\widetilde\cG_{ik}({\delta_\star}),$$
which again contradicts the optimality of  $\widetilde{V}(\delta_\star)$. \end{enumerate}
From the previous two cases we conclude that the situation in Figure~\ref{fig:ProofThem2} cannot happen at optimality, that is, that $\widetilde{V}(\delta)$ must be smooth in $(0,1)$. ~~$\Box$\vspace{0.5cm}

{}

{\sc Proof of Proposition~\ref{prop:consistency}}: The result follows from the continuity of the operator defined by the optimization problem~\eqref{eq:Q(x,E)}. For completeness, assume that $k$ is large enough so that $|1-Q^k(x,{\cal E},\theta)/{\cal Q}(x,{\cal E})|<\varepsilon$. Observe that 
$$\left|\frac{Q^k(x,{\cal E},\theta)}{{\cal Q}'(x,{\cal E})}-1\right|\leq \max\left\{\left|\frac{(1-\varepsilon)\,{\cal Q}(x,{\cal E})}{{\cal Q'}{(x,\cal E})}-1\right|,\left|\frac{(1+\varepsilon)\,{\cal Q}(x,{\cal E})}{{\cal Q'}{(x,\cal E})}-1\right|\right\}.$$
Therefore, subject to $\sum_{x \in {\cal E}} {\cal Q}(x,{\cal E})=1,$ we have that
$$\min_{{{\cal Q}'} \geq 0}\; \max_{\theta \in \{\thetaL,\thetaH\}}\; \max_{x \in {\cal E}}\; \left|\frac{Q^k(x,{\cal E},\theta)}{{\cal Q}'(x,{\cal E})}-1\right|\leq \min_{{{\cal Q}'} \geq 0}\; \max_{x \in {\cal E}}\max\left\{\left|\frac{(1-\varepsilon)\,{\cal Q}(x,{\cal E})}{{\cal Q'}{(x,\cal E})}-1\right|,\left|\frac{(1+\varepsilon)\,{\cal Q}(x,{\cal E})}{{\cal Q'}{(x,\cal E})}-1\right|\right\}\leq \varepsilon.$$

The second inequality is obtained by taking ${\cal Q}'=\cal Q$. This shows that the optimization operator is continuous at $\cal Q$
and that ${\cal Q}^k\rightarrow {\cal Q}$ as $k\rightarrow\infty.$ ~~$\Box$




\vspace{0.5cm}
{\sc Proof of Proposition~\ref{prop:E*MNL}}: Let $\widetilde{\cal E}^*$ be a solution to \eqref{eq:optdisplayMNLasympt}. We will prove the first part of the proposition by showing that $\widetilde{\cal E}^*$ satisfies the following properties:
\begin{enumerate}
    \item If $i \in \widetilde{\cal E}^*$ and $\Delta u_i \geq \Delta \bar{u}(\widetilde{\cal E}^*)$ then $j \in \widetilde{\cal E}^*$ for all $j \geq i$ such that $\Delta u_i < \Delta u_j$.
    \item If $i \in \widetilde{\cal E}^*$ and $\Delta u_i \leq \Delta \bar{u}(\widetilde{\cal E}^*)$ then $j \in \widetilde{\cal E}^*$ for all $j \leq i$ such that $\Delta u_j < \Delta u_i$.
\end{enumerate}

These two conditions imply that there exist two integers $n_1$ and $n_2$ such that $\widetilde{\cal E}^*={\cal E}[n_1,n_2]$.

We will only show the first point since the second follows the same line of arguments. Suppose by contradiction that there exist $i \in \widetilde{\cal E}^*$ and $j \not\in \widetilde{\cal E}^*$ such that $\Delta \bar{u}(\widetilde{\cal E}^*) \leq \Delta u_i < \Delta u_j$. Let us consider another display set $\widehat{\cal E}=\widetilde{\cal E}^*\cup\{j\} \setminus\{i\}$. We will show that $\tilde{\sigma}^2(\widehat{\cal E}) > \tilde{\sigma}^2(\widetilde{\cal E}^*)$ which contradicts the optimality of $\widetilde{\cal E}^*$. Let $m = m_{\widetilde{\cal E}^*}=m_{\widehat{\cal E}}$ denote the cardinality of the sets $\widetilde{\cal E}^*$ and $\widehat{\cal E}$, we have that

\begin{eqnarray*}
\tilde{\sigma}^2(\widetilde{\cal E}^*) &=& {1 \over m} \sum_{ k \in \widetilde{\cal E}^*} (\Delta u_k)^2 - {1 \over m^2}\Big(\sum_{ k \in \widetilde{\cal E}^*} \Delta u_k \Big)^2 \\ &=& {1 \over m} \Big(\sum_{ k \in \widehat{\cal E}} (\Delta u_k)^2  + (\Delta u_i)^2 - (\Delta u_j)^2\Big)- {1 \over m^2}\Big(\sum_{ k \in \widehat{\cal E}} \Delta u_k + \Delta u_i -\Delta u_j \Big)^2 \\
& = & \tilde{\sigma}^2(\widehat{\cal E}) + {\Delta u_i - \Delta u_j \over  m} \, \left ( \Delta u_i + \Delta u_j  -{2 \over m} \sum_{ k \in \widehat{\cal E}} \Delta u_k - {\Delta u_i - \Delta u_j \over  m}  \right) \\
& = & \tilde{\sigma}^2(\widehat{\cal E}) + {\Delta u_i - \Delta u_j \over  m} \, \left ( { (m-1)\,\Delta u_i + (m+1)\, \Delta u_j \over m}  - {2 \over m} \Big(\sum_{ k \in \widetilde{\cal E}^*} \Delta u_k + \Delta u_j - \Delta u_i\Big) \right) \\
& = & \tilde{\sigma}^2(\widehat{\cal E}) + 2\, {\Delta u_i - \Delta u_j \over  m} \, \left ( { (m+1)\,\Delta u_i + (m-1)\, \Delta u_j \over 2\, m}  - {1 \over m} \sum_{ k \in \widetilde{\cal E}^*} \Delta u_k  \right) \\
& = & \tilde{\sigma}^2(\widehat{\cal E}) + 2\, {\Delta u_i - \Delta u_j \over  m} \, \left ( { (m+1)\,\Delta u_i + (m-1)\, \Delta u_j \over 2\, m}  - \Delta \bar{u}(\widetilde{\cal E}^*) \right) < \tilde{\sigma}^2(\widehat{\cal E}),
\end{eqnarray*}
where the last inequality follows from noticing that the argument inside the large parentheses in the last line is positive since
$\Delta \bar{u}(\widetilde{\cal E}^*) \leq \Delta u_i < \Delta u_j$.

Let us now turn to the proof of second part of the proposition. To this end,  let us suppose that all $\{\Delta u_i\}$ are non-negative. (The proof of the case where  all $\{\Delta u_i\}$ non-positive uses the same argument.)  We will prove the result by invoking the following lemma.

\begin{lem}\label{lem:var}Let $X$ a bounded random variable on $[0,A]$. Then $\var [X]\leq A^2/4.$
\end{lem}
{\sc Proof of Lemma~\ref{lem:var}}: {\sf We prove first the lemma. For this notice that $$\var[X]=\e[ X^2]-(\e[ X])^2 \leq A\,\e[X] -(\e[ X])^2=\e[X](A-\e[X]).$$ The inequality is due to the fact that $X\in [0,A]$. Finally, we observe that $g(x)=x(A-x)$  is maximized on $[0,A]$ at $x=A/2$ with $g(1/2)=A^2/4$. Hence, $\var[ X]\leq A^2/4$.}~$\Box$\vspace{0.2cm}

To use this result, note maximizing the value of  $\tilde{\sigma}^2({\cal E})$ over ${\cal E}$ is equivalent to maximize the variance of a non-negative random variable $X_{\cal E}$ taking values in the set $(\Delta u_i\colon i \in {\cal E})$ with equal probability. It follows   from Lemma~\ref{lem:var} that
$$\var[X_{\cal E}] \leq {1 \over 4} \max_{i \in {\cal E}} \{\Delta u_i^2\} = {(\Delta u_n)^2 \over 4}.$$ For the second inequality, recall that we have indexed the products so that $ \Delta u_1 \leq \Delta u_2 \leq \cdots \leq \Delta u_n$ and we are assuming that they are all non-negative (i.e., $\Delta u_1 \geq 0$). At the same time, if we set $\widetilde{\cal E}^* = \{0,n\}$, then it is easy to see that
$$\var[X_{\widetilde{\cal E}^*}] = {(\Delta u_n)^2 \over 4}.$$
We conclude that $\widetilde{\cal E}^*$ maximizes $\tilde{\sigma}({\cal E})$ over  ${\cal E} \in \mathscr{E}$.~$\Box$\vspace{0.5cm}

{}

{}

\newpage


{}

{}

\end{document}